\documentclass{sn-jnl} % Use the SN journal article format
\usepackage[T1]{fontenc}
\usepackage{lmodern}
\usepackage{amsmath, amsfonts, amssymb}
\usepackage{mathtools} % for additional math functionality
\usepackage{graphicx,xcolor}
\usepackage[utf8]{inputenc}
\usepackage{textcomp}
\usepackage[version=4]{mhchem}
\usepackage{siunitx}
\usepackage{longtable,tabularx}
\usepackage{algorithm}
\usepackage{algpseudocode}
\usepackage{hyperref}
\usepackage{cleveref}
\usepackage{appendix}
\usepackage[sort&compress,numbers]{natbib}
\usepackage{lipsum}
\usepackage{amsmath}
\usepackage{bm}
\usepackage{amsfonts}
\usepackage{amssymb}
\usepackage{listings}
\usepackage{matlab-prettifier}
\usepackage{extramarks}
\usepackage[version=4]{mhchem}
\usepackage{siunitx}
\usepackage{longtable,tabularx}
\usepackage{fancyhdr}
\usepackage[normalem]{ulem}
\usepackage{marginnote}
\usepackage{float}
\newcommand{\norm}[1]{\left\lVert#1\right\rVert}

\newcommand{\R}{\mathbb{R}}

\newcommand{\ocmt}[1]{\normalmarginpar\marginnote{\ob{#1}}}

\newcommand{\oldr}{\textcolor{black}}
\newcommand{\ob}{\textcolor{white}}
\newcommand{\tr}{\textcolor{black}}

\crefname{align}{Eq.}{Eqs.}
\crefname{algorithm}{Algorithm}{Algorithms}
\crefname{equation}{Eq.}{Eqs.}
\crefname{figure}{Fig.}{Figs.}
\crefname{table}{Table}{Tables}
\crefname{section}{Section}{Sections}
\crefname{secinapp}{Appendix}{Appendices}

\title[Autonomous Observability-constrained Lyapunov Control]{\tr{Autonomous Observability-constrained Lyapunov Control for Horizon-based \tr{Small Body} Navigation}}

\author*[1]{\fnm{Aditya Arjun} \sur{Anibha}}\email{aanibha@purdue.edu}
\author[2]{\fnm{Kenshiro} \sur{Oguri}}\email{koguri@purdue.edu}

\affil[1]{\orgdiv{School of Aeronautics and Astronautics}, \orgname{Purdue University}, \orgaddress{\city{West Lafayette}, \state{Indiana}, \postcode{47906}, \country{USA}}}
\affil[2]{\orgdiv{School of Aeronautics and Astronautics}, \orgname{Purdue University}, \orgaddress{\city{West Lafayette}, \state{Indiana}, \postcode{47906}, \country{USA}}}

\abstract{\tr{Small body} exploration is a pertinent challenge due to \tr{low gravity environments and strong sensitivity to perturbations like Solar Radiation Pressure (SRP)}. Thus, autonomous methods are being developed to enable safe \oldr{navigation \tr{and control} around \tr{small bodies}}. These methods often involve using Optical Navigation (OpNav) to determine the spacecraft's location. Ensuring OpNav reliability would allow the spacecraft to maintain an accurate state estimate throughout its mission. \tr{This research presents an observability-constrained Lyapunov controller that steers a spacecraft to a desired target orbit while guaranteeing continuous \tr{OpNav} observability. We design observability path constraints to avoid regions where horizon-based OpNav methods exhibit poor performance, ensuring control input that maintains good observability.} \oldr{This controller is implemented with} a framework that simulates \tr{small body} dynamics, synthetic image generation, edge detection, horizon-based OpNav, and filtering. \tr{We evaluate the approach in two representative scenarios—orbit maintenance and approach with circularization—around spherical and ellipsoidal target bodies. In Monte Carlo simulations, the proposed approach improves the rate of attaining target orbits without observability violations by up to 94\% compared to an unconstrained Lyapunov baseline, demonstrating improved robustness over conventional methods.} \footnote{A preliminary version of this work was previously presented during the 4th Space Imaging Workshop at Georgia Institute of Technology, Atlanta, GA.}}


\begin{document}

\maketitle

% Rewrite from scratch
\section{Introduction}
\begin{figure}[htbp!]
	\centering
	\includegraphics[width=0.9\linewidth]{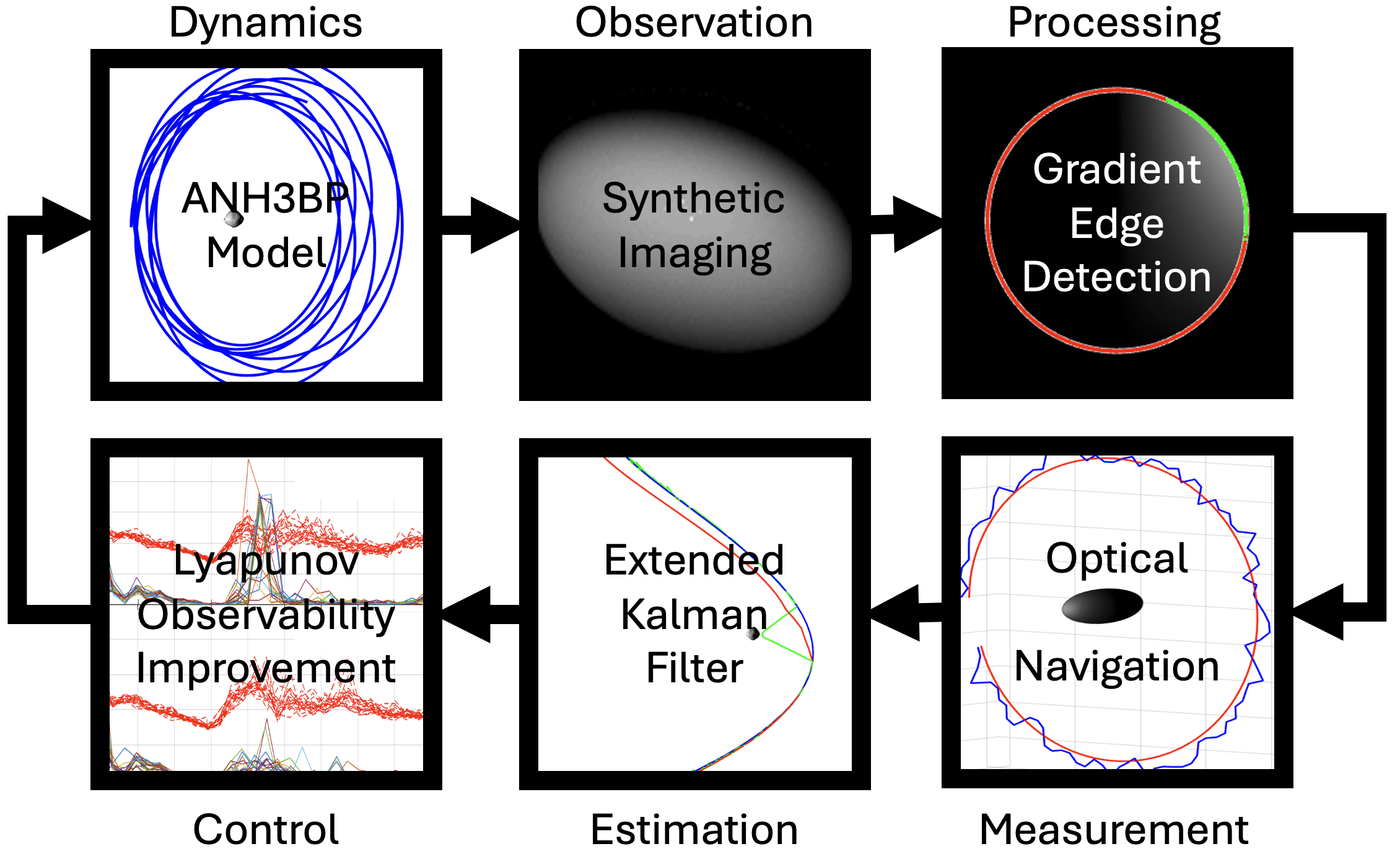}
	\caption{Overall Algorithm Schematic}
        \label{schem}
\end{figure}

\tr{Small body} exploration has gained significant interest, driven by scientific research, resource utilization, and planetary defense purposes. Successful missions, such as \textit{Hayabusa2} to 162173 Ryugu and \textit{OSIRIS-REx} to 101955 Bennu, have provided valuable insights into the composition and structure of these \tr{small bodies} \cite{HayabusaRyugu, bennudia}. However, \oldr{both missions used human-in-the-loop OpNav throughout the mission and} required extensive observation periods to safely navigate their respective targets' uncertain and weak gravitational environments\tr{, which pose a significant navigational challenge due to trajectories being highly sensitive to perturbations like Solar Radiation Pressure (SRP) that may cause spacecraft deviation into poor observability regions or unsafe paths}\cite{Ogawa2020, McCarthy2022}. For example, \textit{OSIRIS-REx} spent nearly a year mapping Bennu before attempting proximity operations \cite{Adam2022}, while \textit{Hayabusa2} similarly mapped Ryugu for \oldr{3} months \cite{HayabusaOVR}. In addition to these missions, the \textit{DART} and \textit{Lucy} missions also emphasize the need for detailed pre-mission observations to ensure safe proximity operations \cite{micheldart,levison2021lucy}. These missions highlight the need for improvements in autonomous Optical Navigation (OpNav) to reduce the reliance on long observation periods or human-in-the-loop processes to improve mission efficiency and reliability. 
\\\\
OpNav has become a preferred method for space missions due to its reliability, accuracy, and accessibility, particularly when navigating larger spherical bodies like the Moon or Mars \cite{qi2023autonomous}. OpNav relies on visual images of celestial bodies, allowing spacecraft to determine their position by analyzing surface features or horizons. Some types of OpNav include surface feature tracking, Line Of Sight (LOS) \cite{ning2017doppler}, Central and Apparent Diameter (CAD) \cite{christian2015optical}, Lidar-based \cite{mcmahon2014asteroid,woods2016lidar}, Limb-based \cite{liounis_operational_2022} and Pole-from-Silhouette, Shape-from-Silhouette or Localization-from-Silhouette \cite{mcmahon2022light} methods. The type of OpNav used in this research is the Christian-Robinson Algorithm (CRA). The CRA is a type of horizon-based OpNav method that calculates the position of an observer with respect to the center of mass of an ellipsoid target body directly from the edge-detected apparent horizon points in an image rather than using curve fitting, and it has been used extensively in spherical-body missions and shown to be computationally efficient, accurate and feasible for real-time autonomous applications \cite{OpNav}. \tr{However, like other horizon-based methods, CRA’s performance strongly depends on the viewing geometry and lighting conditions. When the spacecraft is positioned near the dark side or in configurations with limited visible limb curvature, the quality of the measurements can degrade sharply due to poor observability, reducing the ability of the navigation filter to maintain accurate state estimates. Ensuring persistent observability throughout a mission is therefore a critical challenge for autonomous navigation around small bodies.}
\\\\
There are existing techniques that \oldr{maintain} OpNav \oldr{observability, however, they are computationally expensive and infeasible for autonomous application onboard a single spacecraft.} \cite{qiao2022asteroid} introduces an approach to optimize the spacecraft's orbit during the approach phase by enhancing OpNav observability using the Fisher Information Matrix (FIM). While this method significantly improves measurement reliability, it is still challenged by uncertainties due to the asteroid's irregular surface and dynamic environment, which could introduce errors. Additionally, the optimization process is too time-consuming and computationally costly, thus posing a challenge for its onboard application. In another work \cite{Pugliatti2022}, the authors develop data-driven image processing techniques for OpNav to enhance navigation accuracy around binary asteroids. However, the computational demand for processing large datasets in real-time presents a challenge, potentially limiting onboard resources. \cite{jia2020observability} presents an observability-based navigation strategy that combines optical and radiometric measurements to improve navigation during the asteroid approach phase. The spacecraft configuration is optimized to enhance measurement accuracy and state estimation. However, this is only applicable to multi-spacecraft formations.
\\\\
\tr{Small body proximity operations demand controllers that can satisfy safety and mission constraints such as keep-out zones, actuator limits, field-of-view/lighting requirements for navigation sensors while remaining computationally feasible for onboard usage. Optimization-based methods have been developed extensively towards this demand. Convex programming approaches can include thrust bounds, approach corridors, and state-triggered constraints with formal feasibility guarantees. Some examples include lossless convexification and successive convexification for powered descent and proximity guidance, which deliver real-time, fuel-efficient solutions under tight constraints \cite{AcikmesePloen2007,SzmukAcikmese2020,HayesPei2024}. Another type is Model Predictive Control (MPC), which has been demonstrated for autonomous rendezvous and docking with explicit state/control constraints and keep-out regions, achieving robust performance against target tumbling and environmental uncertainties \cite{Li2017,Bashnick2023,DiCairano2012}. Complementary safety-critical control with control barrier functions (CBFs) provides a Lyapunov-like invariance mechanism enforced through online quadratic programs, now a matured tool to guarantee set invariance (e.g., collision avoidance, line-of-sight cones) while permitting separate performance objectives \cite{AmesTAC2017,AmesECC2019}. On the estimation side, observability has also been treated as a constraint or objective since angles-only navigation is known to lose range observability along certain trajectories thereby motivating information-aware trajectory design and “observability maneuvers” that enforce good view-geometry and lighting constraints \cite{WoffindenGeller2009,DAmico2013,Mok2020,Chu2020}. This paper builds upon similar principles by treating navigation observability and sensing geometry as operational constraints to be considered by the controller through path constraints.}
\\\\
\tr{Lyapunov methods offer nonlinear, feedback designs with provable stability and light onboard computational cost, which is ideal for this application. These have seen sustained use across space applications in the form of hovering and precision descent near small bodies via sliding-mode controllers with Lyapunov proofs \cite{Furfaro2013,Furfaro2014}, formation flying and relative motion regulation using orbital-element feedback and artificial potentials \cite{Kristiansen2009,Renevey2019}, and attitude/pose regulation under uncertainties \cite{Lee2020}. Classical Lyapunov designs target asymptotic convergence without explicitly encoding path constraints. However, constraint handling can be integrated by augmenting the Lyapunov construction with artificial potentials, state penalties, or by coupling with barrier-type invariance conditions, which is an approach that bridges to the constrained Lyapunov and CBF literature \cite{AmesTAC2017,AmesECC2019}. Recent advances even address constrained Lyapunov stabilization for orbital transfer problems using Gauss variational equations, illustrating how Lyapunov functions and constraint satisfaction can coexist within a principled design \cite{Garone2024}. Taken together, this literature shows Lyapunov control is both established for nonlinear spacecraft dynamics and extensible to include constraint-aware control.}
\\\\
\tr{Due to the limitations of current OpNav observability maintaining techniques and the controller demands of small body proximity operation applications, there is a need to develop a computationally efficient, robust, and accurate \oldr{observability-maintaining control} algorithm to facilitate safe, robust and autonomous small body navigation.} Ideally, this algorithm should function at a wide range of distances and angles while accounting for regions with poor observability to ensure continuous and safe operation. Various studies discuss robust orbit control around asteroids. \oldr{\cite{oguri_risk-aware_2021, oguri_robust_2021} apply stochastic optimal control to optimize trajectories under gravitational and operational uncertainties, with more recent work employing chance-constrained convex programming for safe autonomy \cite{oguri_chance_constrained_control_2024}, chance-constrained sensing-optimal path planning for angles-only navigation \cite{ra_oguri_sensing_optimal_path_2024}, and sequential covariance steering for robust cislunar low-thrust trajectory optimization under uncertainty \cite{ kumagai_oguri_robust_cislunar_2025}. \cite{batista_negri_autonomous_2022} uses state transition tensors for rapid autonomous orbit transfers after parameter estimation, \cite{ishizuka_asteroid_2022, ishizuka_robust_2022} develop uncertainty-aware rendezvous optimization, \cite{boone_stochastic_2024, liu_convex_2024} apply chance-constrained convex optimization for stochastic maneuver design, and \cite{wang_spacecraft_2024} ensures stability through Lyapunov redesign under mass distribution uncertainties. However,} they do not explicitly consider the navigation performance in their formulations.
\\\\
This research proposes a novel solution: developing a Lyapunov controller that uses path constraints to avoid poor optical observability during asteroid missions using the \oldr{CRA}. Lyapunov control has been explored for application in the asteroid mission context, however the focus has been on orbit-attitude control and hovering operations \cite{oguri_solar_2020, furfaro2015hovering, lee2019noncertainty}.  This newly proposed controller is dedicated to maintaining the usability of OpNav by ensuring that the spacecraft avoids regions with poor observability through real-time trajectory adjustments.  This approach allows for continuous and reliable optical measurements, even near the asteroid's dark side, thereby enhancing the overall safety and robustness of autonomous asteroid navigation.
\\\\
\tr{In the problem context of this paper, the controller must keep the spacecraft in observability-favorable configurations for OpNav, effectively a path constraint tied to lighting and viewing geometry, while remaining simple and robust enough for onboard execution. Optimization-centric schemes (convex/MPC) offer powerful guarantees but incur nontrivial online solve times, tuning/initialization overhead, and algorithmic complexity \cite{SzmukAcikmese2020,Li2017,Bashnick2023}. By contrast, a Lyapunov-based design yields a closed-form feedback law with stability guarantees and negligible per-step compute, and it can incorporate observability-related path constraints through smooth artificial potentials or barrier-like terms, which is conceptually aligned with safety-critical control \cite{AmesTAC2017,AmesECC2019}. This makes Lyapunov control an appropriate choice to solve this problem as it preserves stability and respects observability-based constraints as documented in observability-aware guidance \cite{WoffindenGeller2009,Mok2020,Chu2020}, and remains lightweight enough for continuous onboard operation during extended small-body proximity phases.}
\\\\
The core technical approach of this paper involves deriving a Lyapunov controller that utilizes artificial potential functions to maintain a path with optimal observability. Additionally, we develop a synthetic image generation tool to simulate varying lighting conditions and test the performance of horizon-based OpNav in different scenarios. These simulations allow us to identify the limitations of existing OpNav methods, which then serve as inputs to our controller design. The proposed controller is tested in simplified mission scenarios, including orbit maintenance and approach for \tr{small body} capture, with state estimation handled using an Extended Kalman Filter (EKF). The controller's performance is evaluated using Monte Carlo simulations to ensure robustness across various mission profiles. A schematic displaying the flow of the algorithm is shown in \cref{schem}.
\\\\
The remainder of this paper is structured as follows. First, we discuss the dynamics model, horizon-based OpNav, and EKF, which form the foundation of our navigation system. Following this, we present the novel contribution of the observability-constrained Lyapunov controller. Finally, we demonstrate the controller's effectiveness by analyzing OpNav and EKF performance, validated using Monte Carlo simulations under simplified mission conditions. \tr{The synthetic image generation pipeline and edge detection techniques used are described in \cref{app:sigp}.}
\section{Background and Preliminaries}
\subsection{Dynamics}\label{dyn}
The motion of a spacecraft near \tr{small bodies} is influenced by several forces, including the body's gravitational pull and the Sun's Solar Radiation Pressure (SRP). The equations governing this motion are encapsulated in the Augmented Normalized Hill Three-Body Problem (ANH3BP), which provides a normalized framework accounting for these forces in a Cartesian rotating reference frame. \\\\
The Hill frame is defined with the $x$-axis pointing from the Sun to the \tr{small body}, the $z$-axis pointing toward the angular velocity of the \tr{small body}, and the $y$-axis completing the frame.\\\\
The spacecraft's normalized position and velocity are denoted by \( ^H\bm{r} = [x, y, z]^\top \) and \( ^H\bm{v} = [\dot{x}, \dot{y}, \dot{z}]^\top \) respectively within the Hill frame, where \( ^H\dot{\bm{r}} \) and \( ^H\dot{\bm{v}} \) are their time derivatives and the left superscript H represents that the vector is in the Hill frame. The dynamical system is non-dimensionalized with the unit length \((\mu/\mu_{\mathrm{Sun}})^{1/3}R\) and unit time \(1/N\) using the gravitational parameter of the primitive body \( \mu \) and the Sun \( \mu_{\mathrm{Sun}} \), the distance between the Sun and the primitive body $R$ and the mean motion of the primary orbits, \(N = \sqrt{\mu_\mathrm{Sun}/R^3}\). The equations of motion for the ANH3BP are \cite{anh3bp}:
\begin{equation}
\begin{aligned}
& \ddot{x}=2 \dot{y}+3 x-x /\|^H\mathbf{r}\|^3+\beta+u_x \\ & \ddot{y}=-2 \dot{x}-y /\|^H\mathbf{r}\|^3+u_y \\ & \ddot{z}=-z-z /\|^H\mathbf{r}\|^3+u_z
\end{aligned}
\end{equation}
The non-dimensional acceleration due to SRP is represented by \( \beta \) and calculated as follows:
\begin{align}\label{srpeq}
\beta=\left(\frac{G_1}{(m / A) R^2}\right)\left(\frac{1}{N}\right)^2\left(\left(\frac{\mu}{\mu_{\text {Sun }}}\right)^{1 / 3} R\right)^{-1}=\frac{G_1}{(m / A) \mu_\mathrm{S u n}^{2 / 3} \mu^{1 / 3}}
\end{align}
where \( G_1 \) is the solar flux constant and \( m/A \) is the spacecraft mass-to-area ratio. It is important to note that this model involves approximation due to the assumptions that the \tr{small body} is in a circular orbit and SRP acceleration is constant.
\\\\
An important defining condition for an orbit to be bounded around the primitive body within this model is the maximum semi-major axis, defined as follows \cite{anh3bpstable}:
\begin{align}
a_{\mathrm{max}}=\frac{\sqrt{3}}{4} \sqrt{\frac{\mu (m/A)}{G_1}} R
\end{align}
where \(a_\mathrm{max}\) is the maximum semi-major axis of the orbit beyond which escape occurs. \\\\
An alternate representation of these dynamics can be formulated using Milankovitch elements. \oldr{Milankovitch elements are used for the controller because they have no singularity and can target a certain orbit state over a specific point.} First, let $^H\bm{x}_\mathrm{Milan}$ be the state of our system in Milankovitch elements and express the equation of motion as follows:
\begin{equation}
^H\dot{\bm{x}}_\mathrm{Milan}=^H\bm{f}_{0}\left(^H\bm{x}, t\right)+B ^H\bm{u}
\end{equation}
where $^H\bm{f}_{0}\left(^H\bm{x}, t\right)$ represents the natural orbital dynamics and $B$ is the control matrix that maps the control input $^H\bm{u}$ to the state.\\\\
To model the ANH3BP using Milankovitch elements, the Gauss planetary equations are used \cite{rosengren_milankovitch_2014}. However, since Gauss planetary equations are conventionally represented in the inertial frame, $^H\bm{f}_{0}\left(^H\bm{x}_\mathrm{Milan}, t\right)$ must be modified accordingly. The rotation of the Hill frame relative to the inertial frame is considered using the transport theorem as follows:
\begin{equation}
\begin{aligned}
    ^H\dot{\bm{h}}&=\dot{\bm{h}}-\tilde{\bm{\Omega}}\bm{h}=\tilde{\bm{r}}\bm{a}_d-\tilde{\bm{\Omega}}\bm{h}\\
    ^H\dot{\bm{e}}&=\dot{\bm{e}}-\tilde{\bm{\Omega}}\bm{e}=\frac{1}{\mu}(\tilde{\bm{v}}\tilde{\bm{r}}-\tilde{\bm{h}})\bm{a}_d-\tilde{\bm{\Omega}}\bm{e}
\end{aligned}
\end{equation}
where the inertial rate of change of the angular momentum vector and eccentricity vector are $\dot{\bm{h}}$ and $\dot{\bm{e}}$ respectively. $\bm{a}_d$ includes the perturbing accelerations of SRP, solar gravity and the control input. The rotation terms resolve to $-\tilde{\bm{\Omega}}\bm{h}=[-\Omega h_1,\Omega h_2,0]^\top$ and $-\tilde{\bm{\Omega}}\bm{e}=[-\Omega e_1,\Omega e_2,0]^\top$. $\bm{h}=\tilde{\bm{r}}\bm{v}$ is the angular momentum vector,  $\bm{e}$ is the eccentricity vector and the \tr{small body}'s angular momentum vector is $\bm{\Omega}=[0,0,\Omega]$. The tilde operator notation is used on 3D vectors to create the cross product matrix in the following form:
\begin{align}  
\tilde{\bm{r}}=\left[\begin{array}{ccc}0 & -r_3 & r_2 \\ r_3 & 0 & -r_1 \\ -r_2 & r_1 & 0\end{array}\right]
\end{align}
Additionally, the SRP and solar gravity perturbations are added by using the control influence matrix to map them from their Cartesian form to the Milankovitch form. Therefore, the equations of motion in Milankovitch elements are as follows:
\begin{equation}
\begin{aligned}
^H\bm{f}_{0}(^H\bm{x}_\mathrm{Milan}, t)&=[-\Omega h_1,\Omega h_2,0,-\Omega e_1,\Omega e_2,0,\frac{h}{r^2}]^\top+ B(^H\bm{a}_\mathrm{SRP}+^H\bm{a}_\mathrm{Solar})\\
^H\bm{x}_\mathrm{Milan}&=[^H\bm{h}^\top,^H\bm{e}^\top,L]^\top\\
B&=\left[\begin{array}{c}
^H\tilde{\bm{r}}\\
\frac{1}{\mu}(^H\tilde{\bm{v}}^H\tilde{\bm{r}}-^H\tilde{\bm{h}})\\
\frac{^H\hat{\bm{z}}\cdot^H\bm{r}}{h(h+^H\hat{\bm{z}}\cdot^H\bm{h})}^H\bm{h}
\end{array}\right]\\
^H\bm{a}_{SRP}&=[\beta,0,0]^\top\\
^H\bm{a}_\mathrm{Solar}&=-\mu\left(\frac{^H\bm{r}}{\left\|^H\bm{r}\right\|_2^3}+\frac{^H\bm{r}_\mathrm{sb}}{\left\|^H\bm{r}_\mathrm{sb}\right\|_2^3}\right)
\end{aligned}
\end{equation}
where $^H\hat{\bm{z}}$ is the unit vector in the z-axis direction in the Hill frame, $h$ is the angular momentum magnitude, $h_1$ and $h_2$ are the angular momentum components in the x and y directions, $e_1$ and $e_2$ are the eccentricity components in the x and y directions, and $^H\bm{r}_\mathrm{sb}=\left[R,0,0\right]^\top$ in the Hill frame.
\subsubsection{Frozen Terminator Orbit}
The concept of a Frozen Terminator Orbit (FTO) is used as a preliminary test case throughout this work because it offers natural dynamical stability and good horizon-based observability conditions. The condition for an FTO is defined as follows \cite{anh3bpstable}:
\begin{align}
\frac{^H\bm{h}}{h}&=\pm^H\hat{\bm{x}}\\
\frac{^H\bm{e}}{e}&=^H\tilde{\hat{\bm{y}}}\frac{^H\bm{h}}{h}\\
e&=\cos{\Lambda}
\end{align}
where $^H\hat{\bm{x}}$ and $^H\hat{\bm{y}}$ are the unit vectors in the x and y directions in the Hill frame, while $^H\bm{e}$ represents the eccentricity vector and $e$ its magnitude. $\Lambda$ is an angle that indicates the relative strength of the SRP to parametrize the secular SRP dynamics. The result is a near-circular polar orbit with $^H\bm{h}$ aligned with the x-axis in the Hill frame and $^H\bm{e}$ is accordingly positioned above or below the ecliptic. This is a unique frozen orbit solution when considering SRP as the primary perturbation with a point mass \tr{small body} gravity assumption. Any references to an FTO within the study will refer to this definition.
\subsubsection{\oldr{Point Mass Feasibility}}
\oldr{Modeling the \tr{small body} as a point mass is sufficient as the simulations conducted in this research are at relatively high distances from small masses where irregular gravity field effects become negligible. To justify this assumption, we compare the magnitudes of the two most major potential sources of perturbation, J2 and SRP, to determine which one is more significant. Assuming \tr{the considered smooth small body has the properties of} Bennu \tr{and} is a uniform–density triaxial ellipsoid of mass $M = 7.329\times10^{10}\,\mathrm{kg}$, radius $r=0.241\,\mathrm{km}$ and shape ratios $[a,b,c]=[1.1051, 1.0769, 1]$, the principal semi-axes are $[a,b,c]_\mathrm{dim} = \frac{r}{abc^{1/3}}[a,b,c]= [0.25158,\;0.24520,\;0.22752]\,\mathrm{km}$ \cite{bennudia,bennugm}. Thus, the J2 (second zonal harmonic) coefficient follows
\begin{equation}
J_2 \;=\;
\frac{I_z-\tfrac12(I_x+I_y)}{M\,R_{\rm ref}^2}
\;\approx\;3.41\times10^{-2}.
\end{equation}
where$
I_x = \frac{1}{5}M(b^2+c^2),\quad
I_y = \frac{1}{5}M(a^2+c^2),\quad
I_z = \frac{1}{5}M(a^2+b^2)$.}

\oldr{The maximum magnitude of the \(J_2\) perturbation acceleration at a spacecraft
distance \(d_\mathrm{sat}\) occurs over the poles and for the closest distance encountered in this paper's simulations, $1\,\mathrm{km}$, is
\begin{equation}
a_{J_2}^{\max}(d_\mathrm{sat})
\;=\;
\frac{3\,J_2\,\mu\,R_{\rm ref}^2}{d_\mathrm{sat}^4}
\quad\Longrightarrow\quad
a_{J_2}^{\max}(1\,\mathrm{km})
\approx 2.8906\times10^{-11}\;\mathrm{km/s^2},
\end{equation}
where \(\mu=4.8904\times10^{-9}\,\mathrm{km^3/s^2}\) is \tr{the small body}’s gravitational
parameter.  }

\oldr{By contrast, the SRP acceleration in this case is \(\beta\approx1.0243\times10^{-10}\,\mathrm{km/s^2}\) according to \cref{srpeq} and the dynamical parameters outlined in \cref{dynparam}. The gravitational acceleration at $r=1\,\mathrm{km}$ from \tr{the small body} is $a_g=4.8904\times10^{-9}\,\mathrm{km/s^2}$. Comparing, we find that at its maximum, the $J_2$ acceleration is $0.5911\%$ of the gravitational acceleration and $28.2202\%$ of the SRP. Note that for the majority of the trajectories simulated, the spacecraft is not perfectly aligned with the poles and is at further distances, thereby experiencing even lower $J_2$ effects while SRP remains constant. Therefore, we can conclude that the effect of J2 is secondary to SRP, which remains the most significant perturbation. Accordingly, higher order gravity-field irregularities are of even lower significance and would be adjusted for by a filtering algorithm considering an appropriate process noise level. However, for \tr{small bodies} with larger mass, highly elliptic shape, or at farther distances from the Sun, the J2 perturbation may become more significant than SRP. Therefore, if required, extension to mascon or polyhedron models may be performed by adding these small perturbation terms to the ANH3BP dynamics similarly as the SRP by using the control‐influence matrix \(B\), with no change to the core EKF or Lyapunov‐controller structure.}

\subsection{Horizon-based OpNav}\label{horop} 
The CRA \cite{OpNav} and its analytical measurement covariance formulation \cite{OpNavcov} are utilized for the measurement implementation. The observations for the EKF and the simulation are derived from this OpNav process. \oldr{The measurements are weighed by the analytical measurement covariance which propagates through the standard deviation of the detected horizon points $\sigma_\mathrm{pix}$ and is directly used as the measurement noise covariance $R_k$ in the EKF's update step. Since we use synthetic image generation for this research, $\sigma_\mathrm{pix}$ is approximated by representing the detected limb points in polar form relative to the image center then computing the standard deviation of the radii scaled for ellipticity as outlined in previous studies that apply OpNav to synthetic images \cite{Shimane1,Shimane2}.}
The inputs include the pixel coordinates of all the detected points in the camera frame, the inverse camera calibration matrix, the attitude transformation matrix between the parent body and the camera, and the triaxial parameters of the shape of the target body. The inverse camera matrix is defined as follows:
\begin{align}  
\bm{C}^{-1}=\left[\begin{array}{ccc}\frac{1}{d_x} & \frac{-\alpha}{d_x d_y} & \frac{\alpha v_p-d_y u_p}{d_x d_y} \\ 0 & \frac{1}{d_y} & \frac{-v_p}{d_y} \\ 0 & 0 & 1\end{array}\right]
\end{align}
\\
where $d_x$ and $d_y$ are the unit pixel density in the $x$ and $y$ directions in the Hill frame, respectively, $l$ is the focal length, $u_p = S/2$ and $v_p = S/2$ are the principal point coordinates, and \(\alpha\) is the skew of the pixels. $S$ is the image size measured in pixels. The focal length can be calculated as follows:
\begin{align}
l=\frac{(S/2)}{\tan(\theta/2)}
\end{align}
\\
where \(\theta\) is the camera field of view angle.
\\\\
The OpNav measurement returns values in the format 
\begin{align}
^H\bm{z}=\begin{bmatrix}
r_{C,x},
r_{C,y},
r_{C,z}
\end{bmatrix}^\top 
\end{align}
where $^H\bm{r}_C$ is the relative position of the spacecraft to the center of mass of the target body in the spacecraft camera frame.\\\\
Since this is a case of known attitude, the magnitude of the position measurement returned by the OpNav algorithm is transformed into the ANH3BP coordinate frame using the current attitude.\\\\
\oldr{The known attitude assumption has two types of consequences. First, in terms of relative \tr{small body}-camera attitude \textit{determination}, the target \tr{small body}'s inertial attitude can be determined accurately by determining its pole direction and rotation rate through a variety of methods \cite{att1,att2} such as the recently developed PoleStack method that can determine the target's rotational motion onboard the spacecraft as it approaches the \tr{small body} which allows definition of a rotating, target-fixed reference frame and in turn the \tr{small body} inertial attitude \cite{pstack}. The spacecraft's inertial attitude is known from star trackers \cite{star1,star2} and the camera's orientation on the spacecraft is known. These combined usually give an attitude error less than a degree \cite{OpNav,att3}. On the other hand, attitude \textit{control} error is effectively thruster directional error, which are also generally on the scale of a degree or lower based on previous \tr{small body} missions \cite{actr1,actr2}. Therefore, known attitude is a reasonable assumption in the context of this research.}

\subsection{Extended Kalman Filter}
An Extended Kalman Filter (EKF) is used to perform the state estimation using the OpNav measurements. Testing the EKF's performance across various operational scenarios helps establish the algorithm's robustness and identify areas for enhancement. First, some key assumptions include the satellite's ability to continuously observe the \tr{small body} with its optical camera via nadir pointing (i.e., the camera is always oriented to point towards the center of the target body), the attitude of the spacecraft relative to the Hill frame is always known, the \tr{small body} shape is known, and the availability of initial state estimates with inherent uncertainties. These assumptions are made to simplify the problem and focus on estimating the position. Constraints involve the limitations of the camera's resolution and the accuracy of onboard sensors, the maximum semi-major axis for a stable orbit in this system, and the computational complexity of generating and processing synthetic observation images in real time. The EKF algorithm used is as follows:
\begin{equation}
\begin{aligned}
& \hat{\bm{x}}_{k \mid k-1}=f\left(\hat{\bm{x}}_{k-1 \mid k-1}, \bm{u}_{k-1}\right) \\
& \bm{P}_{k \mid k-1}=\bm{F}_k \bm{P}_{k-1 \mid k-1} \bm{F}_k^\top +\bm{Q}_{k-1} \\
& \tilde{\bm{y}}_k=\bm{z}_k-h\left(\hat{\bm{x}}_{k \mid k-1}\right) \\
& \bm{S}_k=\bm{H}_k \bm{P}_{k \mid k-1} \bm{H}_k^\top +\bm{R}_k \\
& \bm{G}_k=\bm{P}_{k \mid k-1} \bm{H}_k^\top  \bm{S}_k^{-1} \\
& \hat{\bm{x}}_{k \mid k}=\hat{\bm{x}}_{k \mid k-1}+\bm{G}_k \tilde{\bm{y}}_k \\
& \bm{P}_{k \mid k}=\left(\bm{I}-\bm{G}_k \bm{H}_k\right) \bm{P}_{k \mid k-1}
\end{aligned}
\end{equation}
where $f$ is the non-linear state transition function, the state transition matrix $\bm{F}$ is calculated as the Jacobian of the equations of motion of the ANH3BP from Equation 1 and the observation matrix $\bm{H}$ is $\begin{bmatrix} I_{3 \times 3},0_{3 \times 3} \end{bmatrix}^\top$.\\\\
\oldr{Summarizing the EKF setup used in this paper, the states estimated are the spacecraft's normalized position and velocity in the Hill frame in the form $^H\boldsymbol{x}=[^H\boldsymbol{r},^H\boldsymbol{v}]^T$. The ANH3BP dynamical model used for the truth is also used by the EKF. Maneuvers are considered by using the control accelerations computed by the Lyapunov controller and executed perfectly (Described in \cref{obscondes}), \(\mathbf{u}\), as known time-varying inputs to the prediction step and adding $B\boldsymbol{u}$ to the ANH3BP state propagation. \tr{ This ‘perfect-dynamics, perfect-control’ assumption isolates the negative effects of poor observability and erroneous measurements on state estimation performance, which is the primary challenge we demonstrate the proposed control solution overcomes.} However, we still include a small diagonal process-noise matrix Q to prevent filter over-confidence. The process noise used is $Q=\mathrm{diag}\bigl([10^{-3} \tr{\text{ km}^2},\,10^{-3}\tr{\text{ km}^2},\,10^{-3}\tr{\text{ km}^2},\;10^{-6}\tr{\text{ (km/s)}^2},\,10^{-6}\tr{\text{ (km/s)}^2},\,10^{-6}\tr{\text{ (km/s)}^2}]\bigr)$, \tr{which is a tuning parameter, and was tuned by using $10^{-3}$ times the order of magnitude of the initial conditions (1 km for position and 10 mm/s for velocity) considered as a starting point, then scaling the order of position and velocity process noise values until desirable performance is attained}. The EKF is iterated at each measurement update. }\\\\
\tr{We note that for all figures depicting distance norm error from either the OpNav measurement or EKF estimate in this paper, we use the square root of the maximum eigenvalue of the position covariance ($3\sqrt{\lambda_{max}(P_{\boldsymbol{r}})}$) to define the error bounds.}\\\\
\section{Observability-constrained Controller Design}\label{obscondes}
In spacecraft control, Lyapunov functions can be designed to guide the spacecraft's state toward a desired configuration, such as a stable orbit, attitude, or particular rendezvous point \cite{schaub2003analytical}. By defining an appropriate Lyapunov function for a spacecraft’s position and velocity, the system can be controlled to remain stable under perturbations, such as SRP or irregular gravitation in the case of an \tr{small body}. Control inputs are applied to approach a desired target state. These inputs are derived from the gradient of the Lyapunov function, ensuring that the spacecraft moves in a direction that reduces the overall system energy. The approach is to design a controller $\bm{u}=\bm{c}(\bm{x})$ such that \( V(\bm{x}) \) is a Lyapunov function.
\\\\
\tr{While Lyapunov-based controllers have been extensively studied for spacecraft orbit and attitude stabilization around small bodies \cite{furfaro2015hovering, lee2019noncertainty, oguri_solar_2020}, prior works generally focus on stability with respect to gravitational and dynamical perturbations, without explicitly incorporating real-time navigation performance metrics into the control design. This section develops an observability-constrained Lyapunov controller in which the penalty functions are derived directly from experimentally measured poor performance regions of a horizon-based OpNav algorithm (CRA). These poor observability regions have been characterized experimentally in Section \tr{4}. Unlike conventional constraint handling in Lyapunov control which often uses generic keep-out zones or safety envelopes, our formulation parameterizes the path constraints in closed form using OpNav performance and observability limitations then augments the Lyapunov function with smooth, continuously differentiable artificial potentials that are compatible with stability proofs while being computationally lightweight.  This combined integration of navigation observability consideration within small body dynamical environments with artificial potential constraints results in a feedback law that both maintains stability and actively steers trajectories away from poor-visibility regions while being real-time onboard feasible. By targeting slow variables and isolating them from fast orbital phase effects, the derivation also reduces onboard computational load without sacrificing robustness. Evidence of constraint satisfaction by the Lyapunov controller derived in this paper is included with the numerical results in \cref{results}.}
\\\\
\subsection{Path Constraint Formulation}\label{pcf}
The path constraints have been experimentally determined by testing the OpNav algorithm’s performance across varying distances and angles.
\\\\
The constraints are listed as follows:
\begin{enumerate}
    \item Minimum radius constraint: $R_{\min}-\norm{^H\bm{r}}_2 \leq 0$ with $R_{\min}=2R_\mathrm{sb}$
    \item Maximum radius constraint: $\norm{^H\bm{r}}_2-R_{\max} \leq 0$ with $R_{\max}=25R_\mathrm{sb}$
    \item Keep-out Cone: $\frac{y^2+z^2}{x^2}-\tan^2\alpha \leq 0$ with $\alpha=30^{\circ}$
\end{enumerate}
where $R_\textrm{sb}$ is the radius of the \tr{small body}.\\\\
Now, we define our three penalty constraints in mathematical terms applicable to either controller. Start by deriving the minimum radius constraint using the parameter $r_\mathrm{min}$:
\begin{align}
g_1(^H\bm{x})&=r_\mathrm{min}^2-\norm{^H\bm{r}}_2^2=r_\mathrm{min}^2-\frac{h^2/\mu}{1+e}
\end{align}
Repeat the process for the maximum radius constraint $r_\mathrm{max}$.
\begin{align}
g_2(^H\bm{x})&=\norm{^H\bm{r}}_2^2-r_\mathrm{max}^2=\frac{h^2/\mu}{1+e}-r_\mathrm{max}^2
\end{align}
Repeat the process for the cone constraint along $x$-axis with half angle $\alpha$.
\begin{align}
g_3(^H\bm{x})&=\cos(\frac{\pi}{2}+\alpha)-(\frac{h_1}{h})
\end{align}
$h_1$ is the component of angular momentum in the x-axis direction in the Hill frame. The partial derivatives of the path constraints are computed in \cref{app:pcpd}
\subsection{Controller Derivation}
We start \oldr{the controller derivation} by only targeting the slow variables and freeing the fast variable, which in this case would be the true longitude. Although the rotation between the Hill and inertial frames was found to affect the angular momentum and eccentricity vector components of $\bm{f}_0$ in \cref{dyn}, it is possible to assume $\bm{f}_{0,slow}\approx0$. Therefore, $^H\dot{\bm{x}}_\textrm{slow}=B_\textrm{slow}{^H}\bm{u}$. This assumption is reasonable because the orbital angular velocity of \tr{the small body (assuming Bennu's properties)} is approximately $(360/436.649)=0.824\mathrm{[degrees/day]}$ or $1.665\times10^{-7} \mathrm{[rad/s]}$ which only leads to a total rotation of about 5 degrees over the duration of the simulations in this research, which is at most 6 days long. For reference, 436.649 days is the orbital period of \tr{the small body considered} \cite{lauretta_osiris-rex_2015}. However, this assumption is only applied for the controller derivation and only affects the theoretical stability guarantees in the context of Lyapunov control. Numerical dynamics propagations are based on the true dynamics. Thus, the error state can be defined as follows:
\begin{align}
\delta ^H\bm{x}_\textrm{slow}=^H\bm{x}_\textrm{slow}-^H\bm{x}_\textrm{slow}^{*}
\end{align}
where the slow state can be written as follows:
\begin{align}
^H\bm{x}_\textrm{slow}=\left[\begin{array}{ll}
^H\bm{h}^{\top},^H\bm{e}^{\top}
\end{array}\right]^{\top}
\end{align}
 Next, considering a candidate Lyapunov function as follows:
\begin{align}
V=\delta ^H\bm{x}_\textrm{slow}^{\top} K \delta ^H\bm{x}_\textrm{slow}
\end{align}
where $K\in \R^{6 \times 6}$ and is positive definite. Therefore, the Lyapunov rate is:
\begin{equation}
\begin{aligned}
\dot{V}&=2\delta ^H\bm{x}_\textrm{slow}^{\top} K \delta ^H\dot{\bm{x}}_\textrm{slow}
\end{aligned}
\end{equation}
The next step is to incorporate path constraints using artificial potential functions. To do this, the Lyapunov function and rate results need to be augmented by considering some additive potential function $V_{P_i}$ from each penalty function, which we can define in the following form:
\begin{align}
V_P=wV(^H\bm{x})P(g(^H\bm{x}))
\end{align}
where $w$ is the weight of the penalty, and $P$ is the penalty function in terms of the path constraint $g(^H\bm{x})\leq0$, which must be negative at the target state and smooth everywhere. The penalty function should also monotonically increase in $g$ for $g>0$. For this, we define the penalty functions as set to $0$ when the constraint condition is not violated and greater than $0$ when it is. This can be represented as follows:
\begin{align}
P(g) \begin{cases}>0 & g>-\varepsilon \\ =0 & g \leq-\varepsilon\end{cases}
\end{align}
where $\epsilon$ may be defined to represent the point at which the constraint becomes applicable. We use the exponential form for all of our constraints since it is a smooth and continuous function and, therefore, compatible with the Lyapunov context.
\begin{align}
P_i(g_i)=e^{k_ig_i}
\end{align}
where $i=1,2,3$ represents the corresponding path constraint and $k$ is the sharpness parameter, one of the variables that may be tuned. Now, we may add the three desired artificial potentials for each path constraint penalty as follows:
\begin{align}
\hat{V}=V+V_{P_1}+V_{P_2}+V_{P_3}
\end{align}
Substituting in for the weights and penalty functions and then taking the derivative, we find the augmented Lyapunov rate expression:
\begin{align}
\dot{\hat{V}}=\dot{V}(1+\sum{\omega_iP_i})+V\sum{\omega_i\dot{P}_i}
\end{align}
Use the following chain rule properties to compute the derivatives of $V$ and $P_i$:
\begin{align}
\dot{V}=\frac{\mathrm{d}V}{\mathrm{d}^H\bm{x_\textrm{slow}}}\frac{\mathrm{d}^H\bm{x_\textrm{slow}}}{\mathrm{d}t}=(\delta^H\bm{x_\textrm{slow}}^\top K\bm) B_\textrm{slow} ^H\bm{u}
\end{align}
\begin{align}
\dot{P_i}=\frac{\mathrm{d}P_i}{\mathrm{d}g_i}\frac{\mathrm{d}g_i}{\mathrm{d}^H\bm{x}}\frac{\mathrm{d}^H\bm{x}}{\mathrm{d}t}=\frac{\mathrm{d}P_i}{\mathrm{d}g_i}\frac{\mathrm{d}g_i}{\mathrm{d}^H\bm{x}}+B_\textrm{slow} ^H\bm{u}
\end{align}
Additionally, to get the desired stabilizing controller, we then set the Lyapunov rate equal to a negative definite quadratic form and algebraically solve for the control expression.
\begin{align}
\delta^H\bm{x}^\top_{\mathrm{slow}} (2(1+\sum{\omega_i\dot{P}_i})K+(K\delta^H\bm{x}_{\mathrm{slow}})\sum{\omega_i\frac{\mathrm{d}P_i}{\mathrm{d}g_i}\frac{\mathrm{d}g_i}{\mathrm{d}^H\bm{x}_{\mathrm{slow}}}})B_{\mathrm{slow}} \bm{u})=-\delta^H\bm{x}^\top_{\mathrm{slow}} I_{6\times6}\delta^H\bm{x}_{\mathrm{slow}}
\end{align}
For compact expression, define the following:
\begin{align}
L=(2(1+\sum{\omega_i\dot{P}_i})K+(K\delta^H\bm{x}_{\mathrm{slow}})\sum{\omega_i\frac{\mathrm{d}P_i}{\mathrm{d}g_i}\frac{\mathrm{d}g_i}{\mathrm{d}^H\bm{x}_{\mathrm{slow}}}})B_{\mathrm{slow}}
\end{align}
where $L \in \R^{6 \times 3}$ and assumed to be full rank. Lastly, Solve for the control $^H\bm{u}$ by using a pseudo-inverse since $L$ is not square:
\begin{align}
^H\bm{u}=-\left(L^{\top} L\right)^{-1} L^{\top} \delta ^H\bm{x}_{\textrm {slow }}
\end{align}

\section{\tr{Determining Poor Observability Regions}}
\oldr{This section demonstrates the method of determining observability-constraining path constraints for the controller, so that the derivation and approach can be generalized and applied to other types of optical navigation methods. All simulations in this paper are conducted using only spheres and triaxial ellipsoids using the radius and shape ratio parameters of} the asteroid Bennu for which abundant information is available \cite{bennudia,bennugm}. We note that this is a simplified base case ideal for testing and tuning the filter and controller as simulated 'truth' and filter predictions can be validated with realistic \tr{small body} physical properties and dynamical scenario, which aligns with the primary control focus of this research. The CRA cannot be directly applied to an irregular body such as Bennu without further extension, which is a problem not considered or solved in this paper and instead suggested as a step for future research \cite{Anibha2025AAS}. reiterate that this paper does not extend image generation or processing, or the CRA for applications to Bennu or other irregular bodies. This is  The relevant initial conditions and properties used are shown in \cref{dynparam}, where $M$ is the \tr{small body}'s mass, $e$ is the eccentricity of the \tr{small body}'s orbit, and $r$ is the mean radius of the \tr{small body}.
\begin{table}[htbp!]
    \centering
    \caption{Dynamical Parameters}
    \begin{tabular}{|c|c|c|c|}
        \hline
        Parameter & Value & \tr{Unit} & \tr{Description} \\
        \hline
        $M$ & $7.329 \times 10^{11}$ & \tr{kg} & \tr{Small body mass} \\
        $\mu$ & $4.890 \times 10^{-9}$ & \tr{km$^{3}$/s$^{2}$} & \tr{Small body gravitational parameter} \\
        $\mu_{\text{Sun}}$ & $1.327 \times 10^{11}$ & \tr{km$^{3}$/s$^{2}$} & \tr{Sun gravitational parameter} \\
        $G_1$ & $1 \times 10^{8}$ & \tr{kg$\cdot$km$^{3}$/(s$^{2}\cdot$m$^{2}$)} & \tr{Solar radiation pressure constant} \\
        $m/A$ & $33$ & \tr{kg/m$^{2}$} & \tr{Spacecraft mass-to-area ratio} \\
        $a$ & $1.685 \times 10^{8}$ & \tr{km} & \tr{Small body orbital semi-major axis} \\
        $e$ & $0.204$ & \tr{--} & \tr{Small body orbital eccentricity} \\
        $d$ & $1.720 \times 10^{8}$ & \tr{km} & \tr{Heliocentric distance at epoch} \\
        $r$ & $0.241$ & \tr{km} & \tr{Small body mean radius} \\
        \hline
    \end{tabular}
    \label{dynparam}
\end{table}
\\\\
Additionally, the OpNav parameters used for this research are shown in \cref{oparam}.
\begin{table}[htbp!]
    \centering
    \caption{OpNav Parameters}
    \begin{tabular}{|c|c|c|c|c|c|}
        \hline
        Parameter & $\theta_{\text{FOV}}$ & $S$ & $\alpha$ & $dx$ & $dy$ \\
        \hline
        Value & $30^\circ$ & 1000 pixels & 0 & 1 & 1\\
        \hline
    \end{tabular}
    \label{oparam}
\end{table}
\\
\oldr{These settings are chosen based on camera settings used on real space missions \cite{Ogawa2020,McCarthy2022,Adam2022,HayabusaOVR,liounis_operational_2022,actr1,actr2,att3,Antreasian2019,Antreasian2022}. An intermediary FOV value is chosen to allow the largest range of capabilities. A narrow FOV camera allows resolution at farther distances but leads to the object exceeding the image frame in proximity, while a wide FOV camera only leads to resolved images at closer distances which may be too small from a safety perspective. Thus, a range of 10-30 is generally ideal for most applications. The image size matches realistic cameras and the other parameters are chosen to emulate an ideal camera since modeling and testing non-ideal camera parameters is outside the focus of this research.}\\\\
The OpNav algorithm's accuracy at varying distances and lighting conditions is examined by running multiple simulations with noise at varied fixed points. \oldr{A perfectly spherical target body is used for this analysis as we intend to determine an exact observability cutoff in the ideal case.}
\subsection{Distance Accuracy Analysis}\label{distacc}
In the case of distance, 1000 test cases are run starting from a distance of 1 km up to 30 km. A side-on view where exactly half of the body is lit is used for consistency to fix the angle at a value where the OpNav algorithm is known to be accurate.\\\\
\tr{As previously mentioned, for all figures depicting distance norm error from either the OpNav measurement or EKF estimate in this paper, we use the square root of the maximum eigenvalue of the position covariance ($3\sqrt{\lambda_{max}(P_{\boldsymbol{r}})}$) to define the error bounds.}

\begin{figure}[htbp!]
    \centering
    \includegraphics[width=0.75\linewidth]{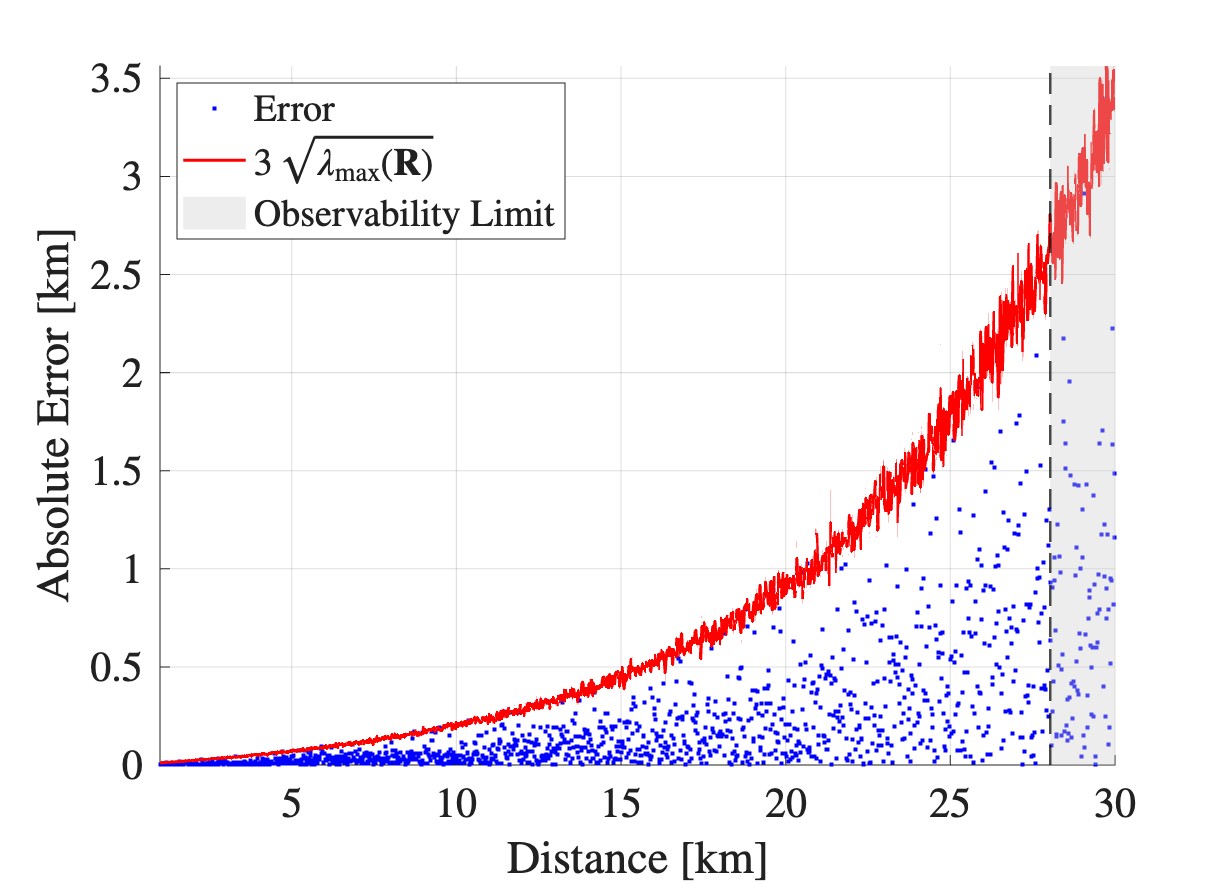}
    \caption{Plot of the OpNav Measurement Error Variation over Distance with Error Magnitude (blue points) and Analytical Covariance (red line)}
    \label{fig:distrror}
\end{figure}
As can be seen in \cref{fig:distrror}, the OpNav algorithm is highly effective at low range but starts to involve major error at distances greater than 25 \tr{km, in this case more than 100} times the radius of the \tr{small body} \oldr{for the combination of $\theta_{FOV}$ and $S$ in \cref{oparam}}. \tr{This is because as range increases, the small body’s apparent radius in pixels shrinks and the measurement becomes progressively less informative.} However, the analytical covariance still provides a good envelope for the errors. We can define one range observability constraint based on this result. \tr{We may define a practical usable "observability limit" when the error bound reaches $10\%$ of the range, which is close to the aforementioned 25 km threshold as seen in \cref{fig:distrror}.} While OpNav still provides results beyond this, eventually, the magnitude of the error exceeds the absolute value of the distance itself, making it unusable after a certain point. Therefore, the algorithm must be used within this range, which has now been successfully determined.\\\\
\tr{Importantly, this threshold is not a fixed distance. it depends on camera intrinsics and the target’s physical radius. For a given camera, a larger body remains observable to greater distances, while a smaller body reaches the same ill-conditioning sooner. Moreover, when the object’s image size varies very slowly with distance (far-field), multiple nearby ranges can yield effectively indistinguishable silhouettes, leading to similar solutions over a band of distances. This explains the flattening/ambiguity observed at long range and motivates a distance (or apparent-size) cutoff for reliable autonomous operation.}
\\\\
\subsection{\tr{Angle Accuracy Analysis}}
In the case of lighting conditions, 1000 test cases are run starting from -90 degrees (straight-on or full moon) up to +90 degrees (dark-side or new moon), where 0 degrees is aligned with the side-on view. These are done with a fixed distance of 5 km for consistency and at a range where the OpNav algorithm is accurate.
\\
\begin{figure}[htbp!]
    \centering
    \includegraphics[width=0.75\linewidth]{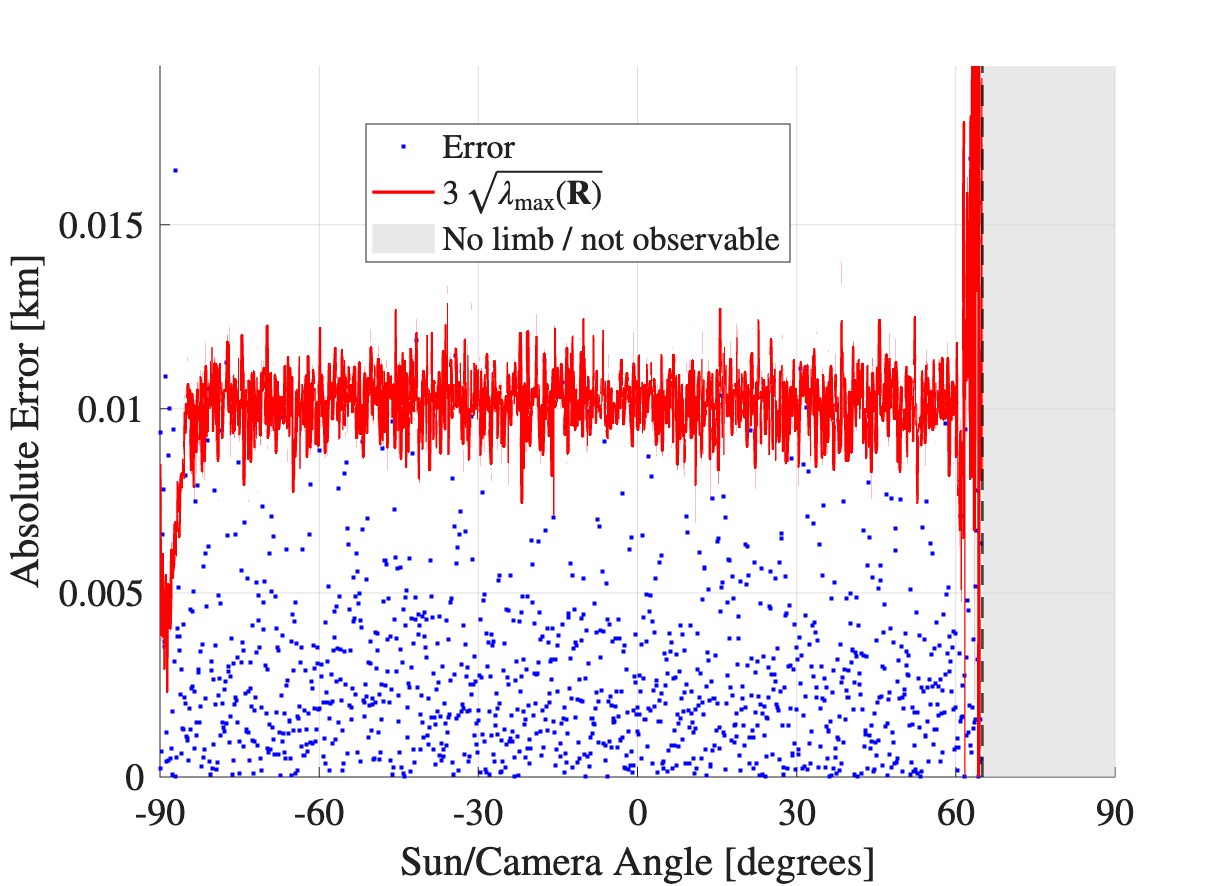}
    \caption{Plot of the OpNav Measurement Error Variation over Angle with Error Magnitude (blue points) and Analytical Covariance (red dashed line)}
    \label{fig:beta}
\end{figure}
\\
We can infer from \cref{fig:beta} that the algorithm is accurate for most angles except straight-on due to the terminator effect and angles close to the dark side as the size of the lit horizon shrinks. The performance deteriorates if the spacecraft views the \tr{small body} from within 30 degrees of the $+x$ axis near the dark side. The covariance is undefined when viewing the dark side, as no horizon points are detected. The error is accordingly equivalent to the true distance of the spacecraft in the implementation used for this research, as the OpNav algorithm returns a zero measurement due to an absence of input horizon points. In other words, the measurement is unusable in these lighting conditions. From this experiment, we define a keep-out cone to ensure observability, which is given as a cone with a 30-degree half-angle along the positive $x$-axis \tr{in the Hill frame}.

\section{Numerical Results}\label{results}
To simulate the application of the Lyapunov controller in a simplified scenario, we implemented it in a loop such that the Lyapunov controller computes each successive timestep based on the EKF's prediction rather than using the true position. The true state is then updated using the control calculated by the controller, from which the OpNav measurement is taken at a realistic frequency of every 1.5 hours and provided to the EKF for the next prediction. This means that any significant error in the EKF's position estimate for the spacecraft could lead to a complete divergence of the true and estimated states. \\\\
\tr{In the context of this paper, we define divergence as any case where the spacecraft violates the observability cone constraint or minimum and maximum distance safety constraints, which we consider as failure cases. We use this definition of divergence as the primary performance metric because the goal of our controller contribution is to ensure a spacecraft can safely and robustly attain a desired orbit around a primary \tr{small body} without colliding into or leaving orbit around the \tr{small body}, which it accomplishes and improves upon conventional control by adding observability constraints that ensure position measurements are persistently reliable along its controlled trajectory.\\\\ Divergence happens because of a combination of factors, starting with an extended period of erroneous measurements in a poor observability region due to violation of the cone constraint. These erroneous measurements cause the EKF's state estimate to drift from the truth trajectory, which in turn leads to incorrect control input that exacerbates this deviation between truth and EKF. Thus, this leads to failure cases where the spacecraft violates the safety constraints, thereby colliding with the primary \tr{small body} or exiting orbit around it. This is because the EKF estimate converges to an incorrect state knowledge matching the desired target orbit while the truth diverges with no corrective control, or erroneous control leads to the spacecraft persistently stuck in the poor observability region. Particularly in cases where the spacecraft leaves orbit around the \tr{small body}, the OPNAV measurement can no longer compensate even after exiting the poor observability region, as its covariance expands at higher distances.}
\\\\
\cref{penwet} summarizes all the relevant penalty weights for both scenarios:
\begin{table}[htbp!]
    \centering
    \caption{Penalty Weight Parameters}
    \begin{tabular}{|c|c|c|}
        \hline
        Parameter & Value & \tr{Description} \\
        \hline
        $\omega_1$ & $1$ & \tr{Weight for min-radius constraint} \\
        $k_1$ & $1$ & \tr{Sharpness for min-radius constraint} \\
        $\omega_2$ & $1$ & \tr{Weight for max-radius constraint} \\
        $k_2$ & $1$ & \tr{Sharpness for max-radius constraint} \\
        $\omega_{3.1}$ & $10$ & \tr{Weight for cone constraint (Controller 1)} \\
        $\omega_{3.2}$ & $100$ & \tr{Weight for cone constraint (Controller 2)} \\
        $k_3$ & $1$ & \tr{Sharpness for cone constraint} \\
        \hline
    \end{tabular}
    \label{penwet}
\end{table}
The penalty weight parameter subscript indicates the corresponding path constraint the parameter is relevant to according to the definitions in \cref{pcf}. All parameters have identical values except the cone weight, for which $\omega_{3.1}$ is used for the \tr{Small Body} Approach Targeting controller and $\omega_{3.2}$ is used for the Orbit Transfer controller.\\\\
\tr{We reiterate that for all figures depicting distance norm error from either the OpNav measurement or EKF estimate in this paper, we use the square root of the maximum eigenvalue of the position covariance ($3\sqrt{\lambda_{max}(P_{\boldsymbol{r}})}$) to define the error bounds.}
\subsection{Orbit Maintenance Scenario}\label{oms}
\begin{figure}[htbp!]
\centering

% --- First row ---
\includegraphics[width=0.45\linewidth,trim={1cm 0.1cm 0 0.1cm},clip]{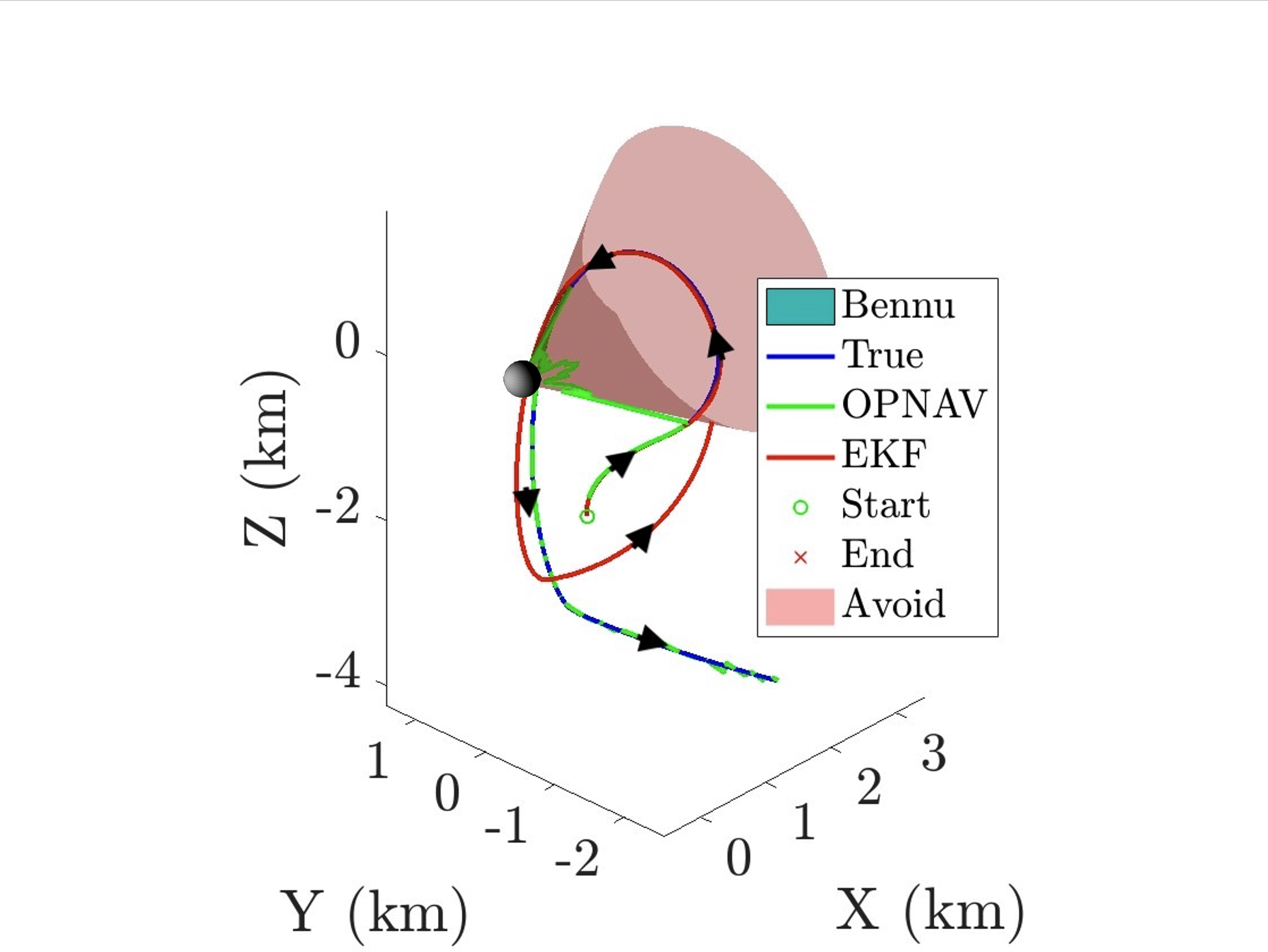}
\hfill
\includegraphics[width=0.45\linewidth,trim={1cm 0.1cm 0 0.1cm},clip]{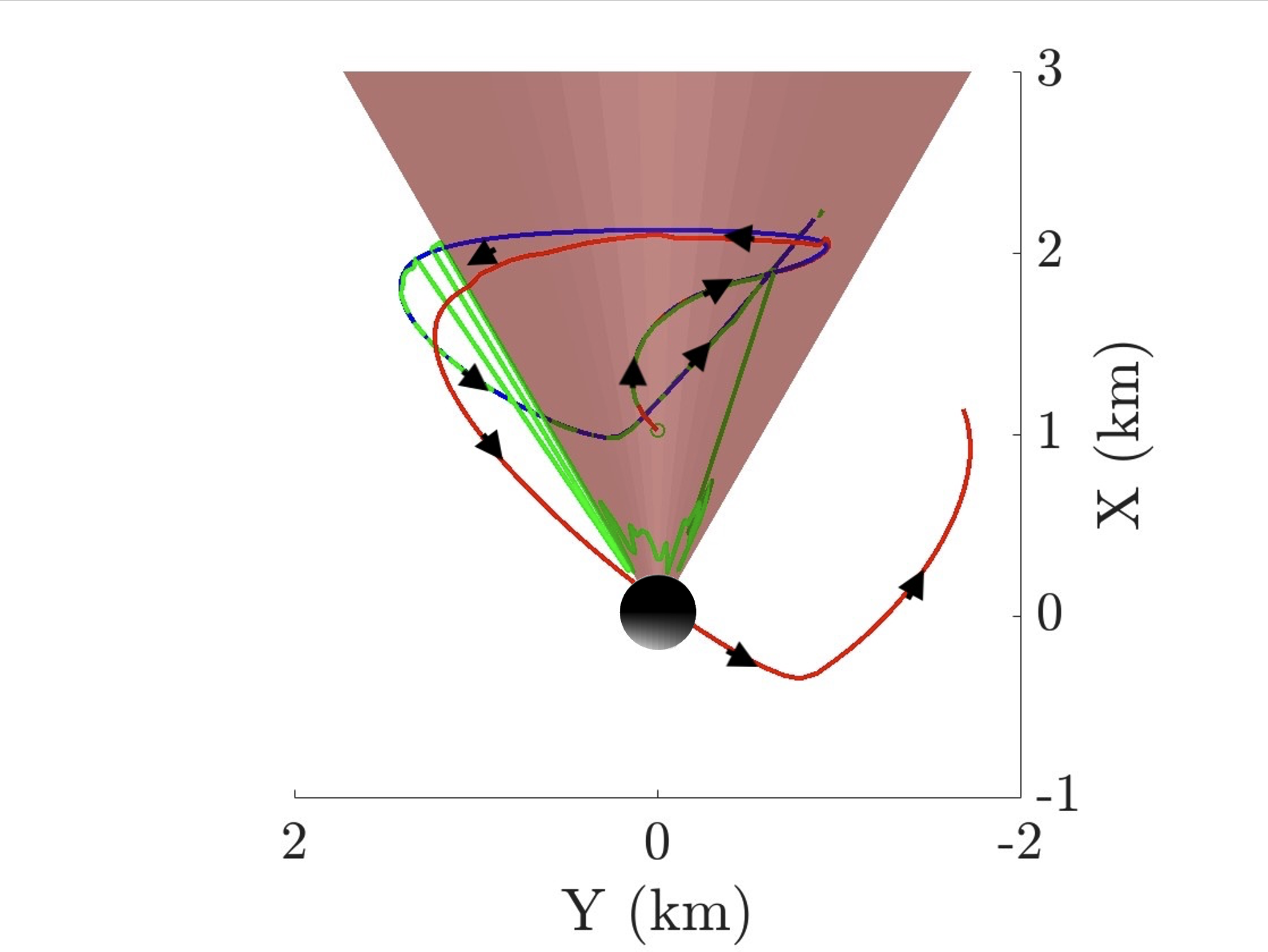}

\vspace{0.3em} % Space between plot and label
\makebox[0.45\linewidth][c]{\small(a) Trajectory (3D view)}%
\hfill
\makebox[0.45\linewidth][c]{\small(b) Trajectory (XY view)}%

\vspace{1em} % Space before next row

% --- Second row ---
\includegraphics[width=0.45\linewidth,trim={1cm 0.1cm 0 0.1cm},clip]{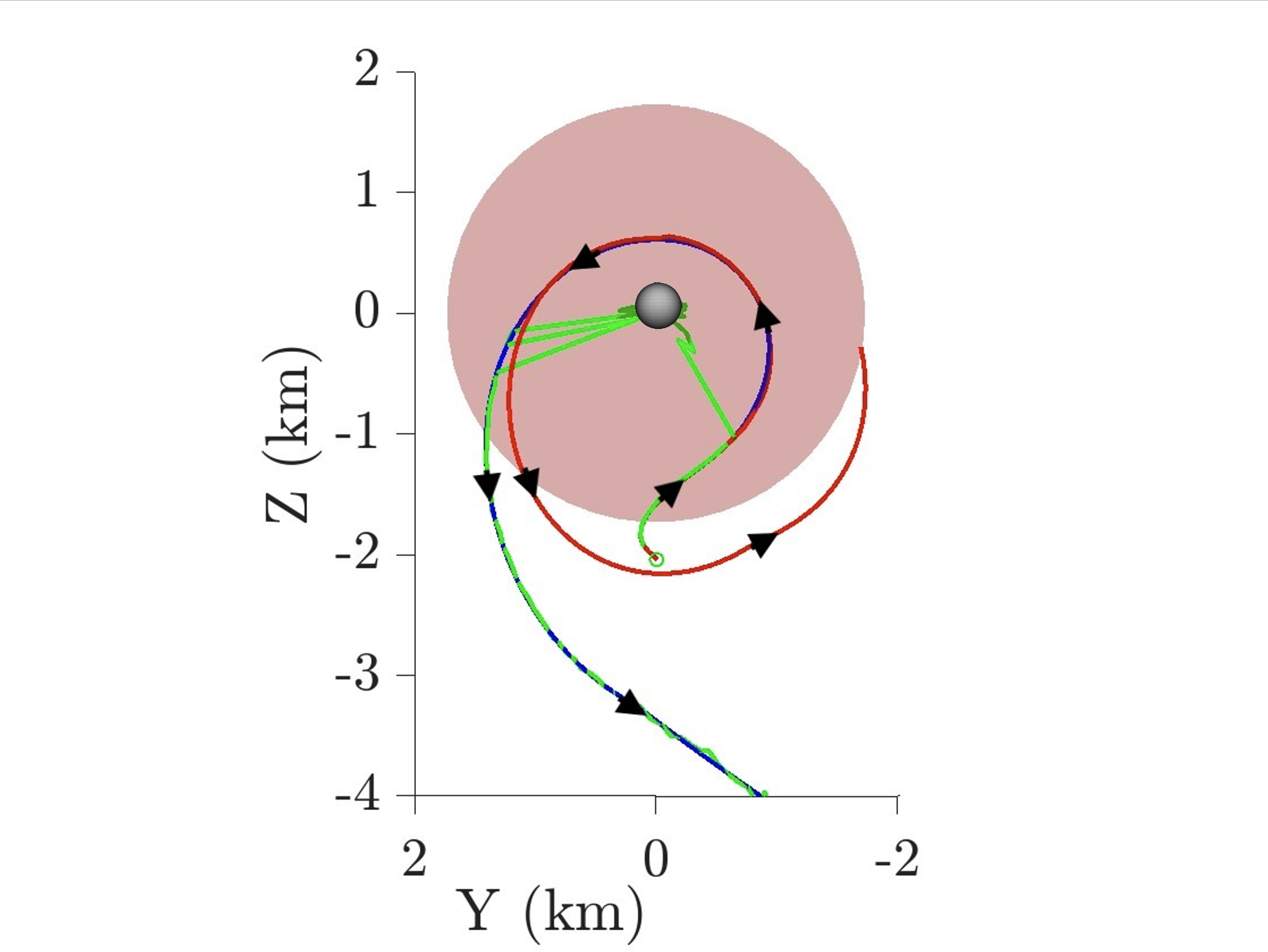}
\hfill
\includegraphics[width=0.45\linewidth,trim={1cm 0.1cm 0 0.1cm},clip]{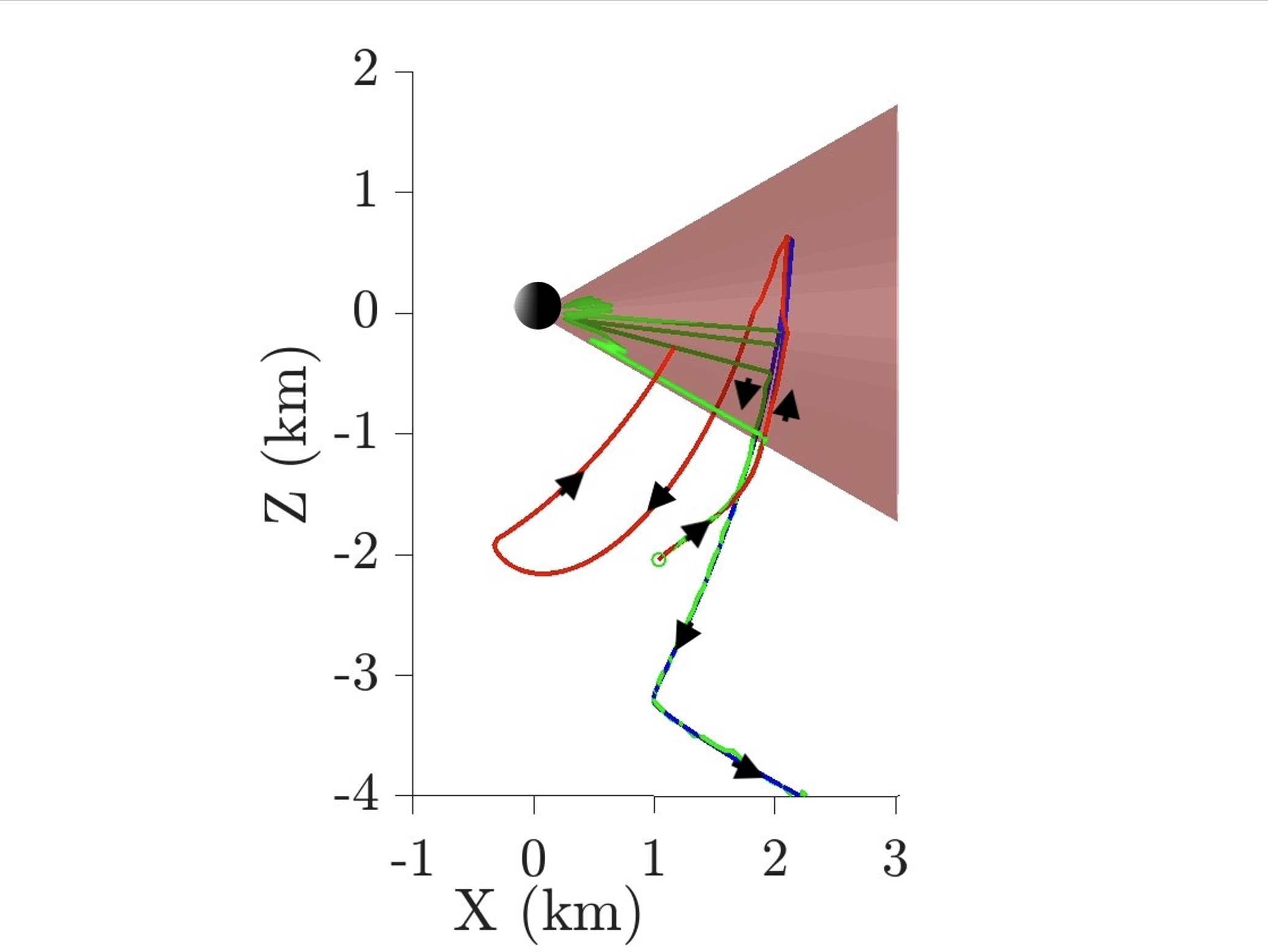}

\vspace{0.3em}
\makebox[0.45\linewidth][c]{\small(c) Trajectory (YZ view)}%
\hfill
\makebox[0.45\linewidth][c]{\small(d) Trajectory (XZ view)}%

\vspace{1em}

% --- Third row ---
\includegraphics[width=0.45\linewidth]{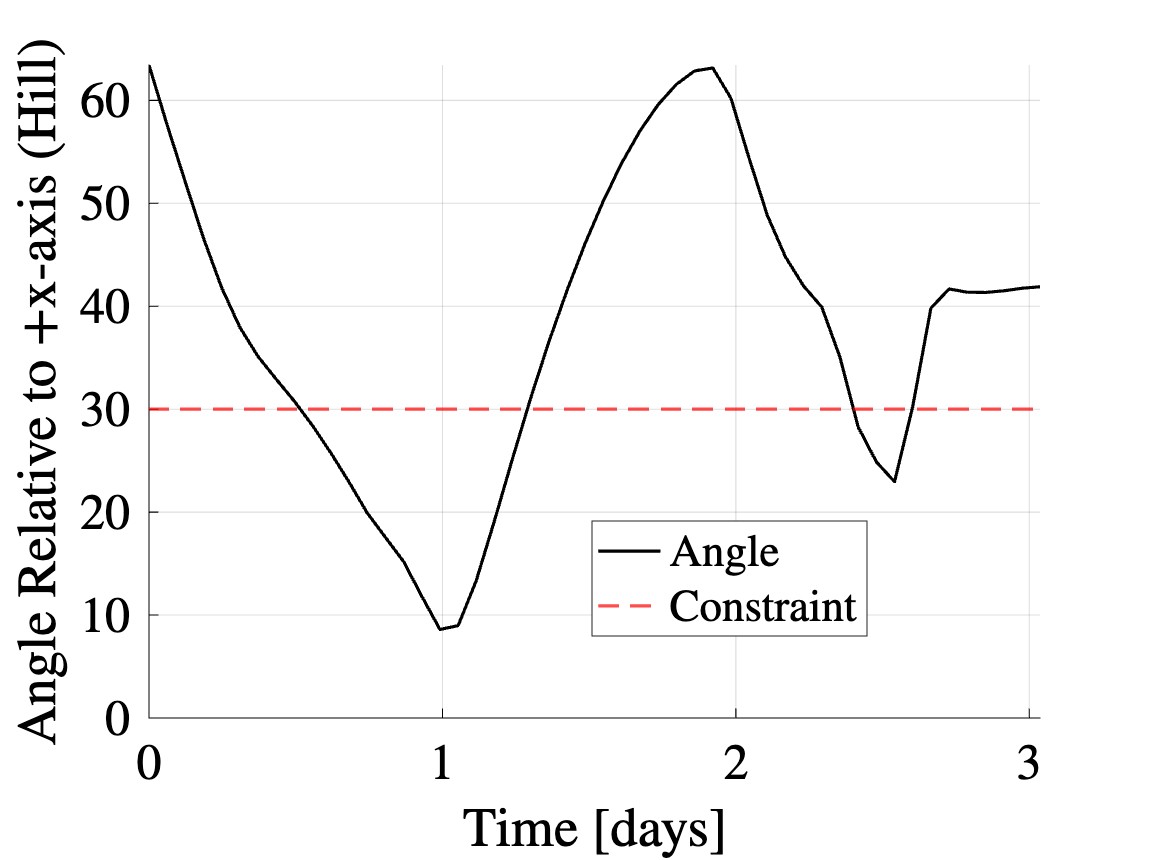}
\hfill
\includegraphics[width=0.45\linewidth]{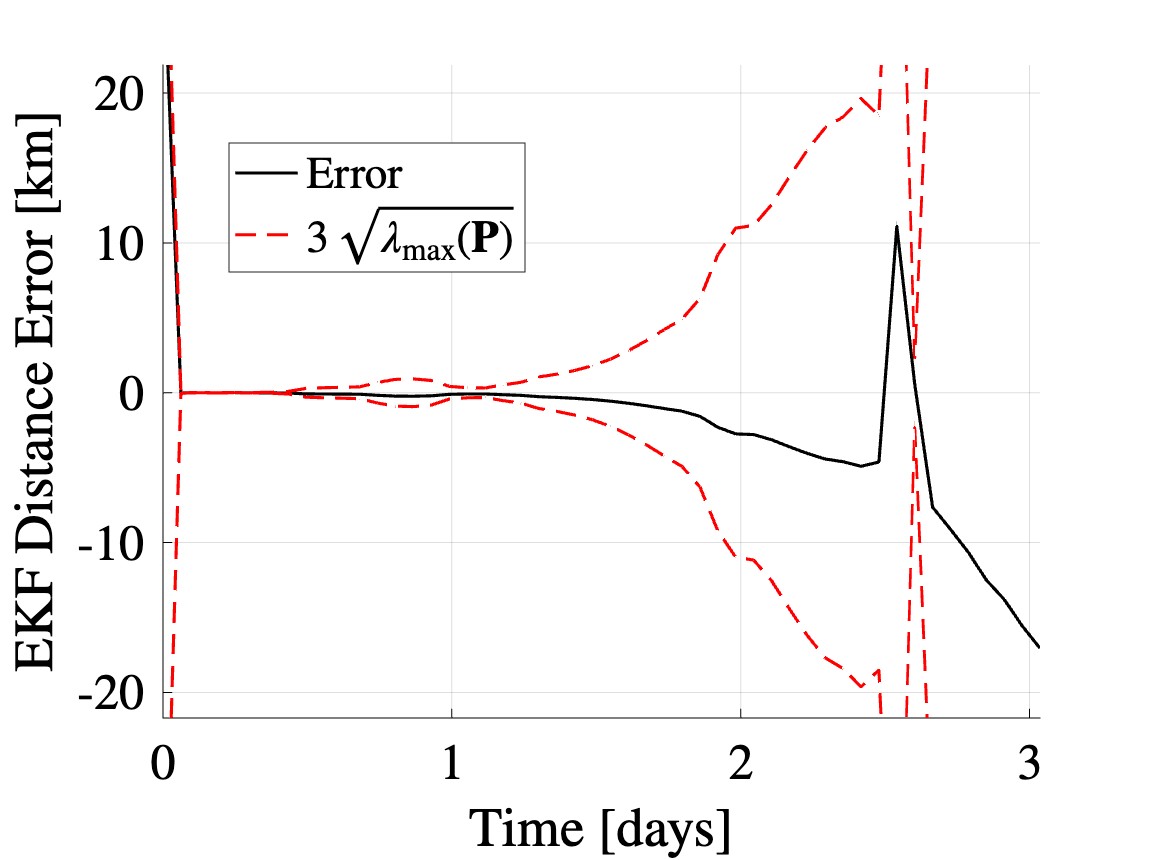}

\vspace{0.3em}
\makebox[0.45\linewidth][c]{\small(e) Angle to +x-axis (Hill) over time}%
\hfill
\makebox[0.45\linewidth][c]{\small(f) EKF distance error over time}%

\caption{\textbf{Fig.\ X} Stationkeeping scenario without Observability-constrained
Penalty Lyapunov control. True (blue), OpNav measurement (green), EKF-estimated (red)
trajectories and keep-out cone (shaded red) are shown in four views. Initial position
(green circle), final position (red cross) and orbital direction (black arrows) indicated.
+x-axis (Hill) relative angle and EKF distance error plots are included to correlate
constraint violation and divergence.}
\label{fig:np}
\end{figure}

\begin{figure}[htbp!]
\centering

% --- First row ---
\includegraphics[width=0.45\linewidth,trim={1cm 0.1cm 0 0.1cm},clip]{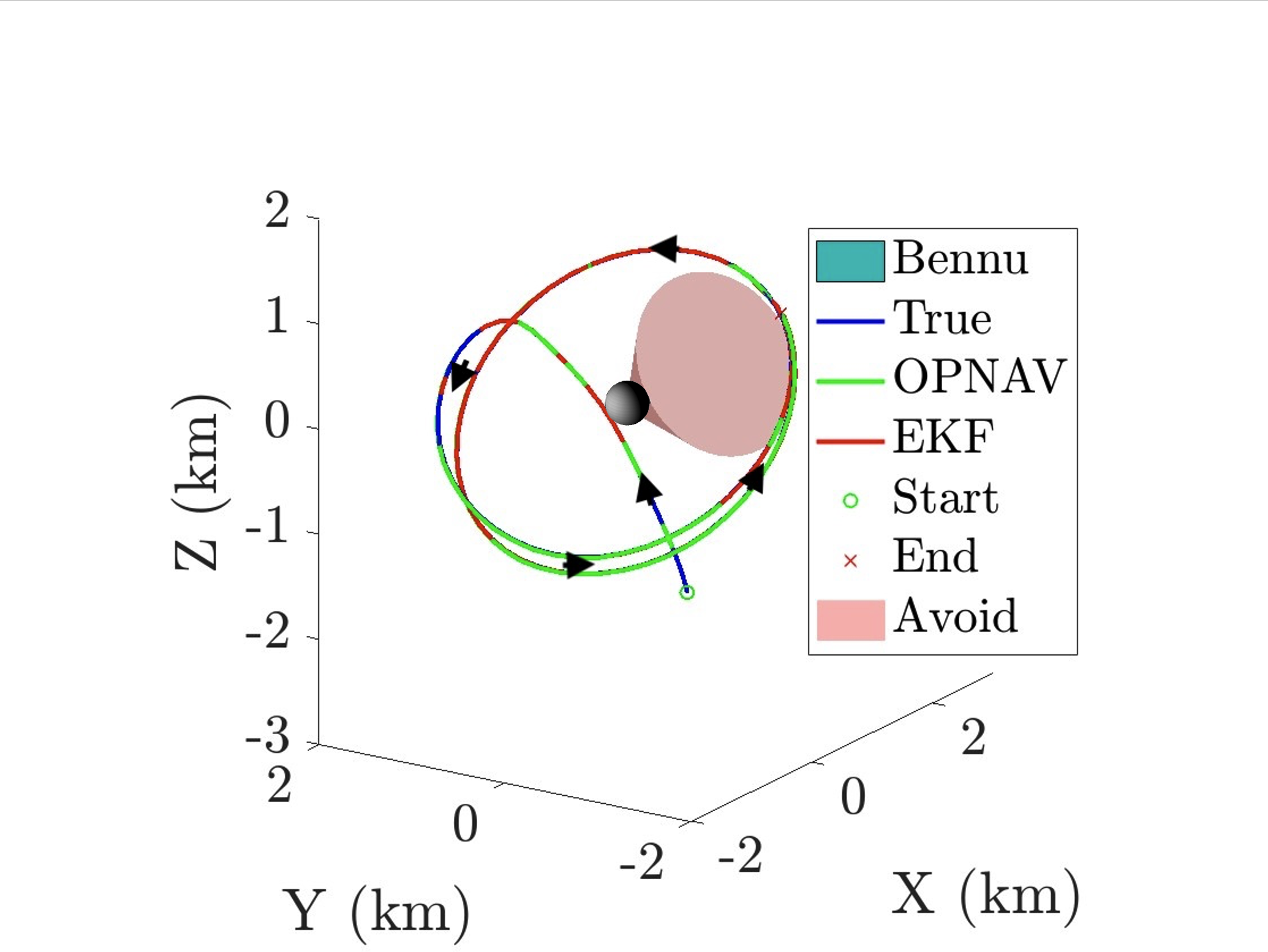}
\hfill
\includegraphics[width=0.45\linewidth,trim={1cm 0.1cm 0 0.1cm},clip]{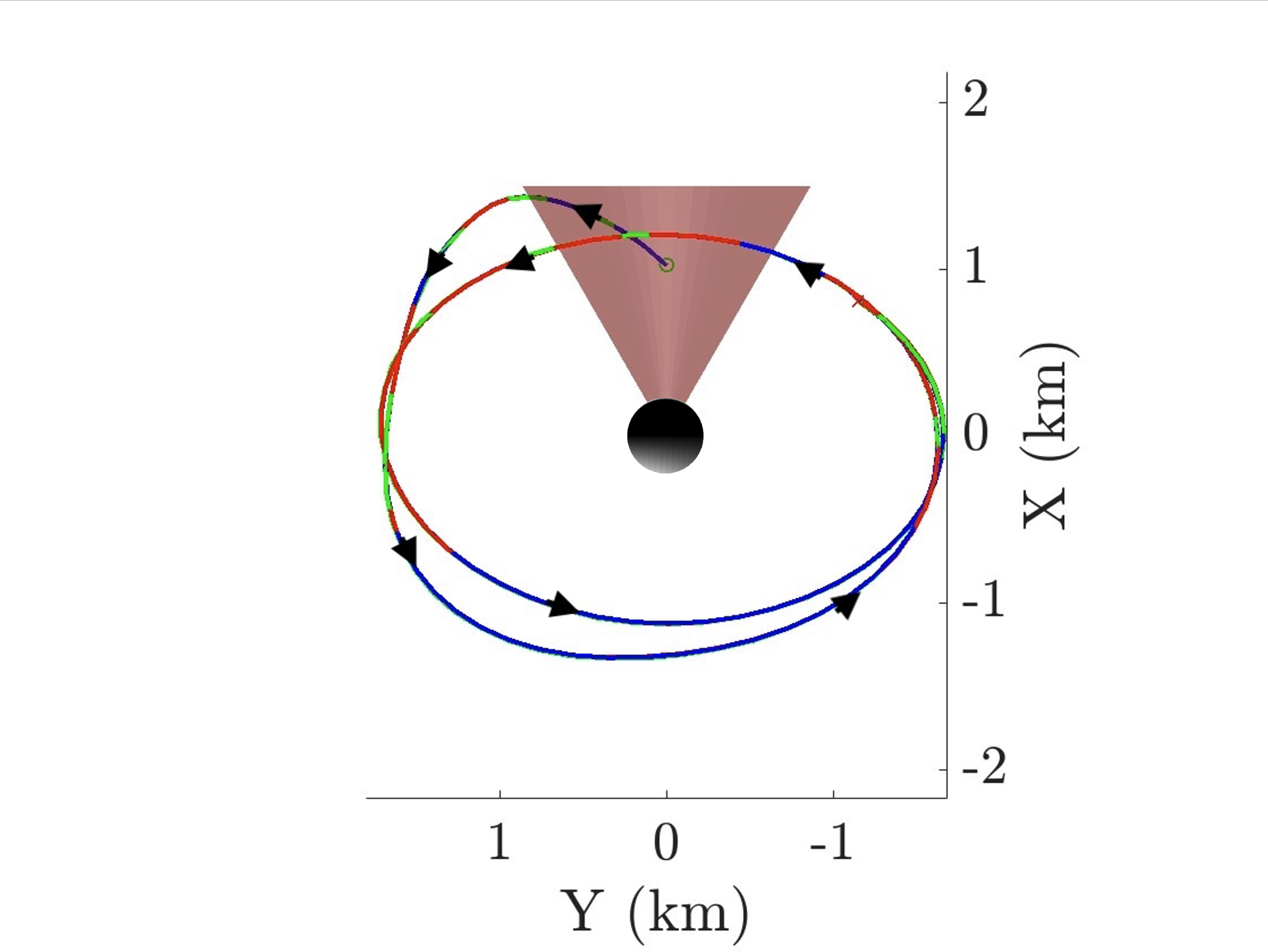}

\vspace{0.3em} % Space between plot and label
\makebox[0.45\linewidth][c]{\small(a) Trajectory (3D view)}%
\hfill
\makebox[0.45\linewidth][c]{\small(b) Trajectory (XY view)}%

\vspace{1em} % Space before next row

% --- Second row ---
\includegraphics[width=0.45\linewidth,trim={1cm 0.1cm 0 0.1cm},clip]{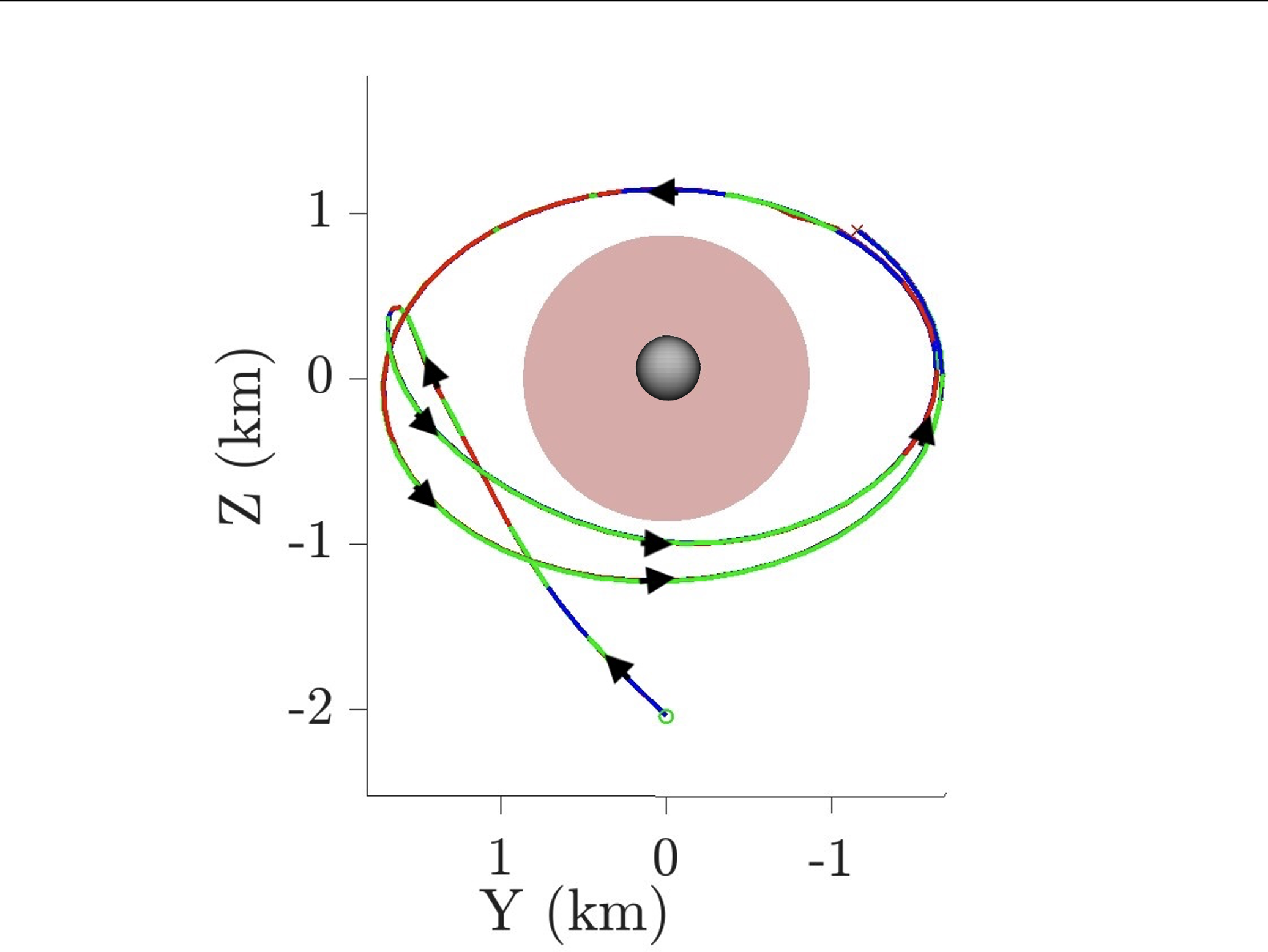}
\hfill
\includegraphics[width=0.45\linewidth,trim={1cm 0.1cm 0 0.1cm},clip]{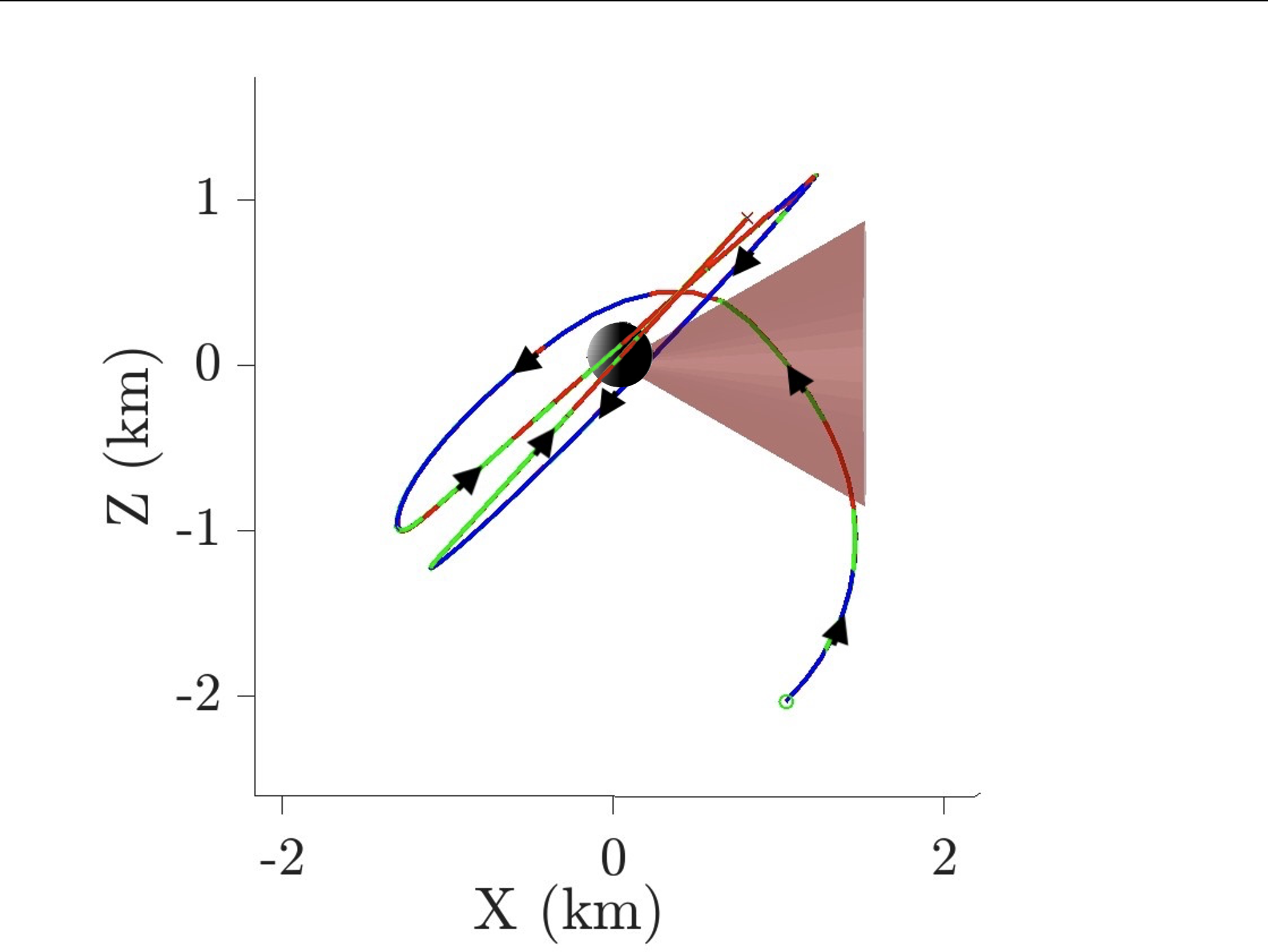}

\vspace{0.3em}
\makebox[0.45\linewidth][c]{\small(c) Trajectory (YZ view)}%
\hfill
\makebox[0.45\linewidth][c]{\small(d) Trajectory (XZ view)}%

\vspace{1em}

% --- Third row ---
\includegraphics[width=0.45\linewidth]{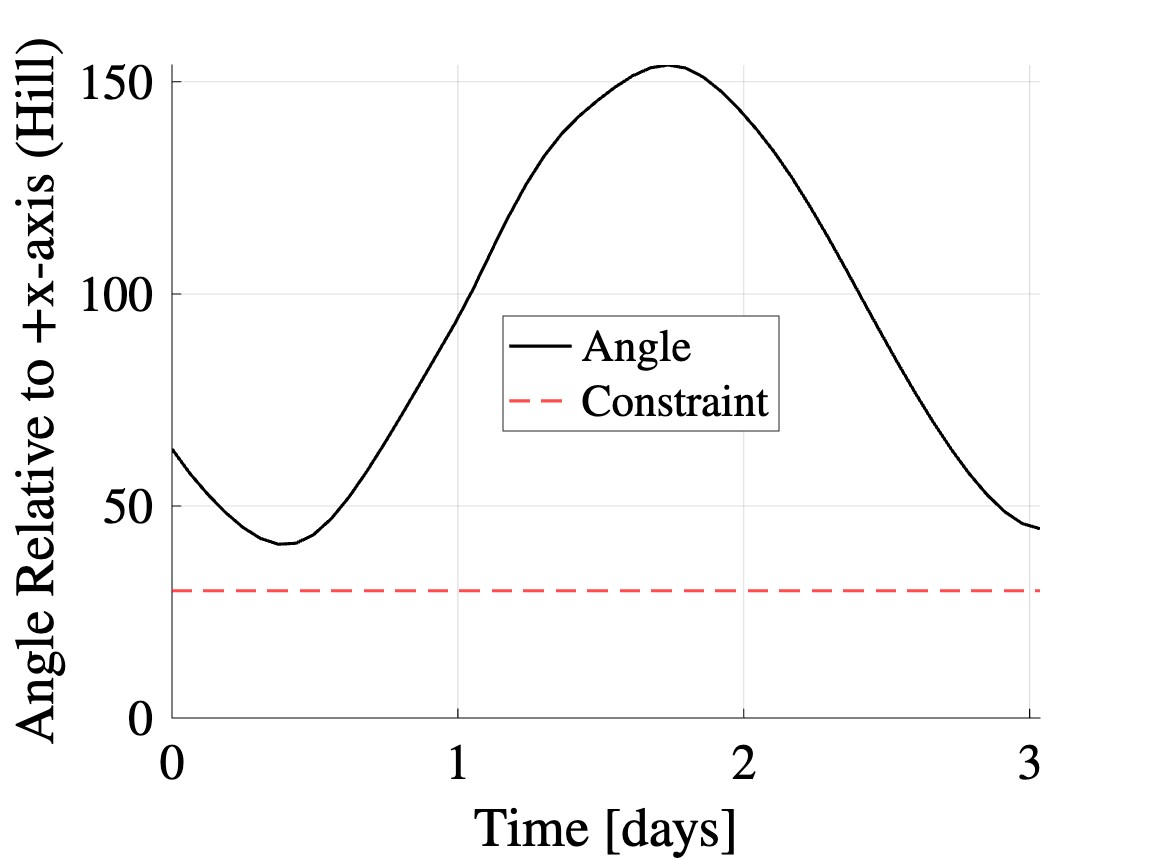}
\hfill
\includegraphics[width=0.45\linewidth]{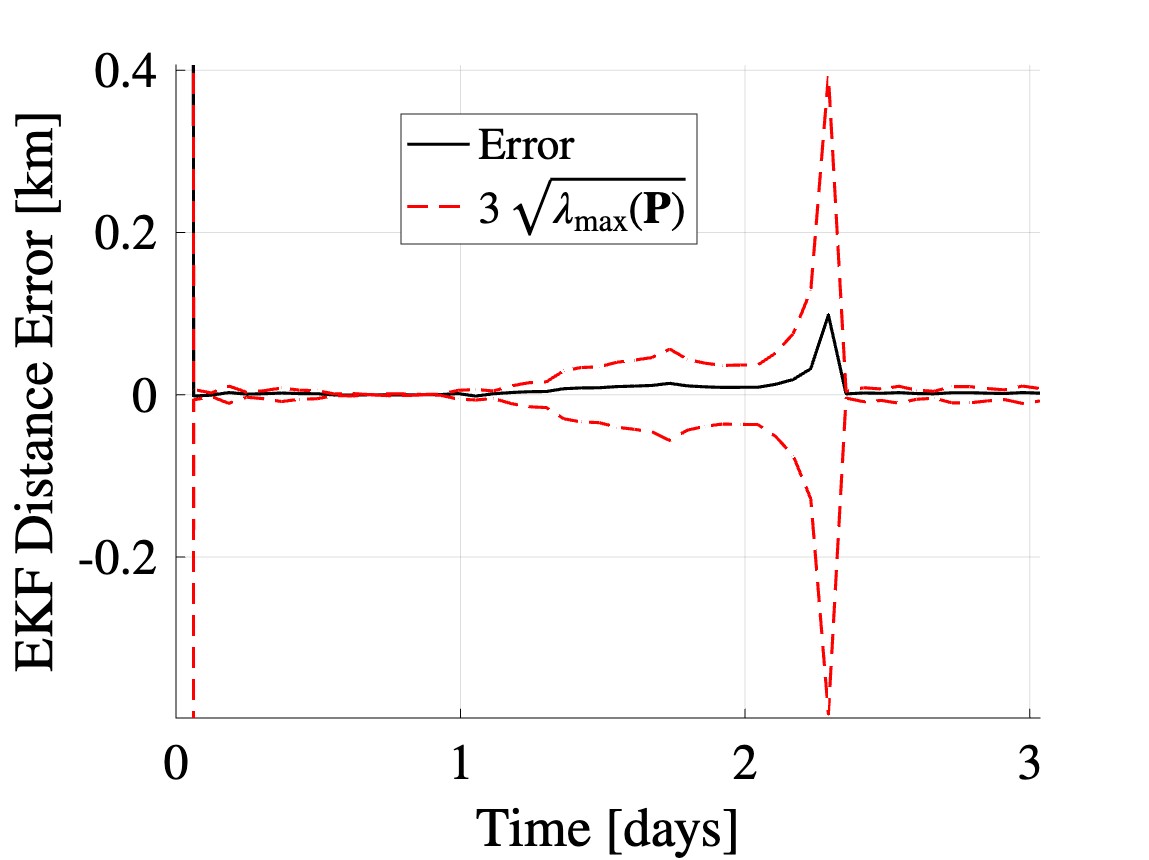}

\vspace{0.3em}
\makebox[0.45\linewidth][c]{\small(e) Angle to +x-axis (Hill) over time}%
\hfill
\makebox[0.45\linewidth][c]{\small(f) EKF distance error over time}%

\caption{\textbf{Fig.\ X} Stationkeeping scenario with Observability-constrained
Penalty Lyapunov control. True (blue), OpNav measurement (green), EKF-estimated (red)
trajectories and keep-out cone (shaded red) are shown in four views. Initial position
(green circle), final position (red cross) and orbital direction (black arrows) indicated.
+x-axis (Hill) relative angle and EKF distance error plots are included to correlate
constraint violation and divergence.}
\label{fig:wp}
\end{figure}

\begin{figure}[htbp!]
\centering

% --- First image ---
\begin{minipage}[b]{0.3\textwidth}
    \centering
    \includegraphics[width=\linewidth]{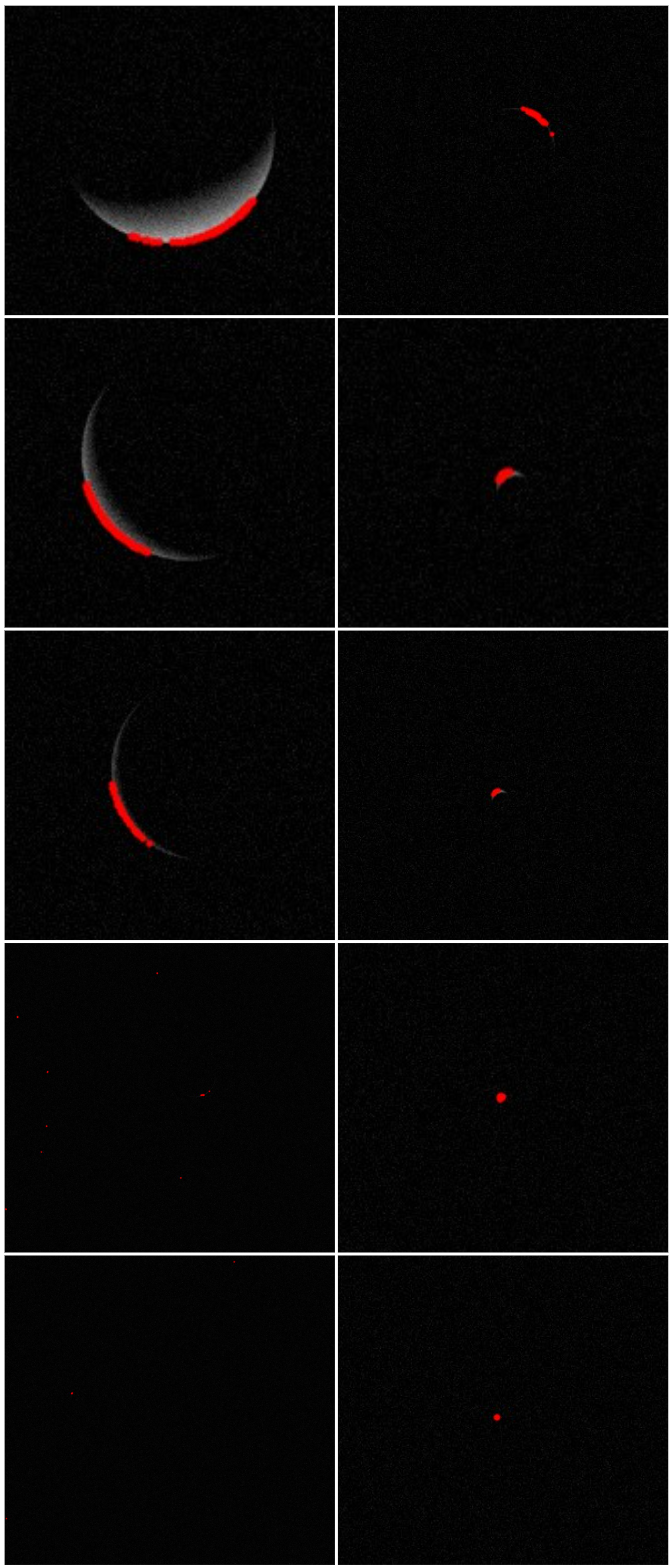}
    \vspace{0.3em}
    {\small (a) Without observability-constrained control}
\end{minipage}
\hfill
% --- Second image ---
\begin{minipage}[b]{0.3\textwidth}
    \centering
    \includegraphics[width=\linewidth]{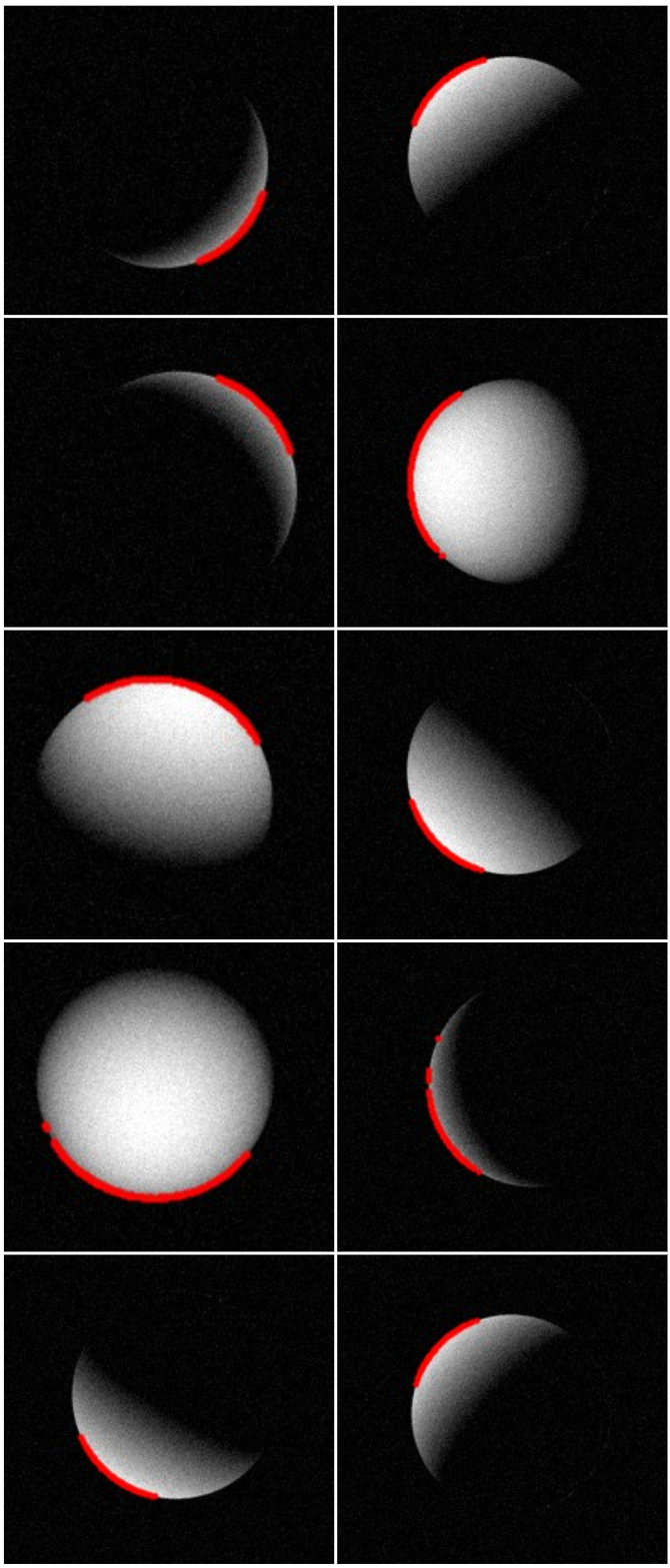}
    \vspace{0.3em}
    {\small (b) Constrained (Spherical primary)}
\end{minipage}
\hfill
% --- Third image ---
\begin{minipage}[b]{0.295\textwidth}
    \centering
    \includegraphics[width=\linewidth]{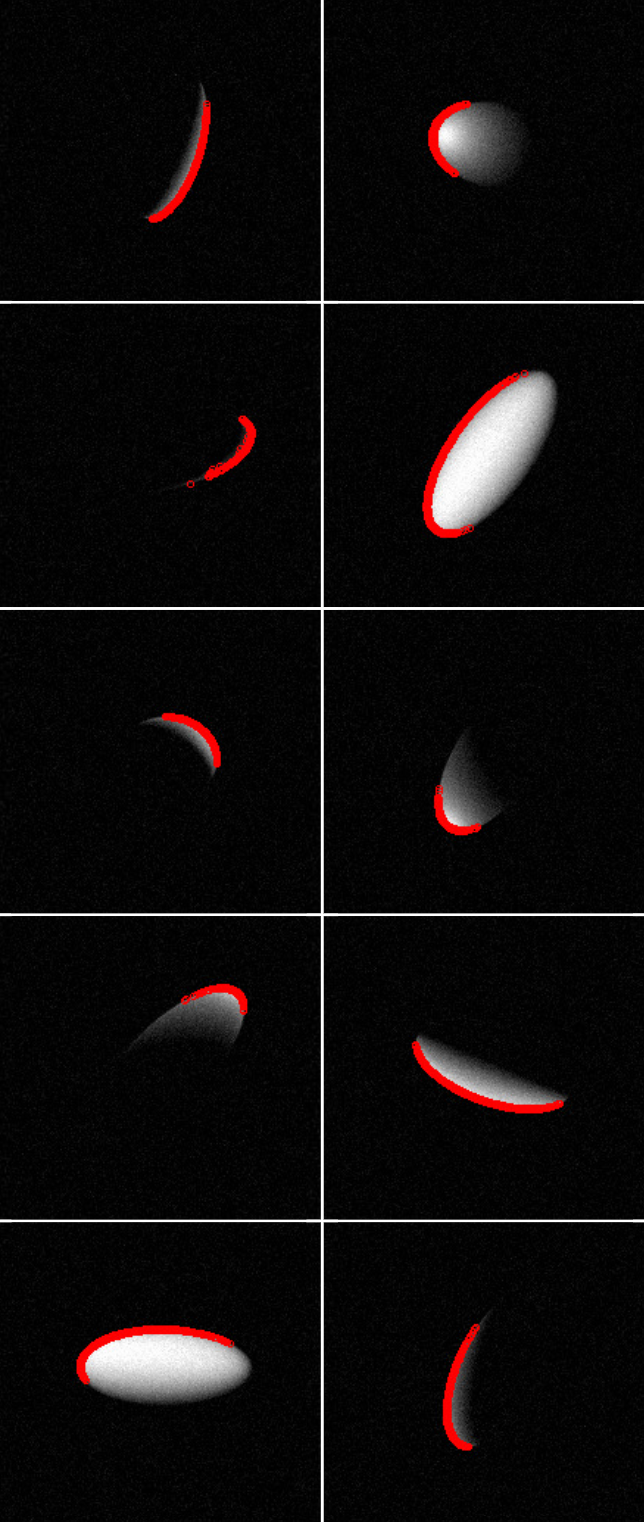}
    \vspace{0.3em}
    {\small (c) Constrained ([2.5 1 1] Ellipsoid primary)}
\end{minipage}

\caption{\oldr{Stationkeeping scenario.} Sample image history with detected horizon points
from \cref{fig:wp}. Shown are: (a) without observability-constrained control, (b) with
observability-constrained control using a spherical primary, and (c) with control using a
[2.5\,1\,1] ellipsoid primary. Images are ordered top–bottom, then left–right.}
\label{fig:gcol}
\end{figure}

To demonstrate the effectiveness of the observability-constrained controller, we develop a test scenario where a spacecraft starts with an arbitrary initial condition on a trajectory to flyby passing behind the dark side of \tr{the small body}. The target is a circular orbit with an inclination just above 30 degrees. This avoids the dark-side cone but provides a useful edge case to test where the spacecraft may often travel through the poor observability region in its controlled trajectory without an observability penalty. The initial conditions in Cartesian terms are $^H\bm{r}=[1.0214,0,-2.0429]^\top$ km and $^H\bm{v}=[40.493,40.493,40.493]^\top$ mm/s. \oldr{Using the tuning guidelines outlined in \cref{tuning},} the gain found to provide the best performance in terms of stability and rate of convergence is as follows: \begin{equation}K_\textrm{2}=\mathrm{diag}(10^{-2},10^{-3},10^{-3},10^{-4},10^{-3},10^{-4})\end{equation}
For this case, the following initial covariance is used (covariances for the position in $\left[\textrm{km}^2\right]$ and velocity in $\left[\frac{\textrm{km}^2}{\textrm{s}^2}\right]$):
\begin{equation}
\begin{aligned}
P_0 = \mathrm{diag}(3.2761,3.2761,3.2761,8.544\times10^{-17},8.544\times10^{-17},8.544\times10^{-17})
\end{aligned}
\end{equation}
To compare results, we test the Lyapunov controller without the observability improving penalty functions as shown in \cref{fig:np}. The figure depicts the spacecraft's true trajectory, OpNav measurement, EKF position estimate and a visualization of the keep-out cone. As can be seen in the results for the controller without an observability penalty, the spacecraft travels through the poor observability region near the \tr{small body}'s dark side. This leads to a faulty measurement, which causes the true and EKF-estimated trajectories to diverge.
\\\\
We now compare this performance to that of the observability-constrained controller in \cref{fig:wp}. In the observability-constrained case, the controller successfully avoids the poor observability region, as can be seen by consistently matching the OpNav measurement and EKF prediction to the true trajectory at all times. An additional test was executed with a [2.5 1 1] ellipsoid using the rotation parameters of Bennu and an approximation from its known mean radius.
\\\\
Camera snapshots of each case with the detected horizon points highlighted in red can be seen in \cref{fig:gcol}. It displays the image history from the spacecraft camera perspective and allows us to make physical inferences about the controller's performance. For the leftmost case without the observability-constrained control, the spacecraft does not avoid the keep-out cone which causes the lit limb to go out of view in the fourth and fifth images. Thus, the number of detected horizon points significantly decreases, causing the spacecraft to diverge. The limb comes back into view later, but is at a larger distance due to the divergence, making the measurement less effective. The middle case displays the same example but with the observability-constrained control, with which we see that the lit limb is kept within view throughout the trajectory. The rightmost case displays the view of the [2.5 1 1] ellipsoid using observability-constrained control to demonstrate the controller's consistency in performance across varying shapes of celestial bodies. Although the size of the lit limb varies considerably due to the body's rotation and viewing the ellipsoidal geometry from different angles, it maintains the lit limb within view by avoiding the poor observability zone and therefore produces more reliable measurements. Both the middle and rightmost cases avoid the divergence and disappearance of lit limb encountered in the leftmost case, supporting the benefit of the observability-constrained controller.
\\\\
\ocmt{OSIRIS-REx used impulses, inappropriate for this case. Replaced with Hayabusa2's ion engines.} \oldr{For this trajectory, the average control acceleration was $4.2865\times10^{-7}\,\textrm{m/s}^{2}$, with a maximum instantaneous acceleration of $4.4163\times10^{-6}\,\textrm{m/s}^{2}$ and total $\Delta v=0.1687\,\textrm{m/s}$. Using the Hayabusa2 spacecraft's ion engine system as a reference,} the maximum fuel consumption to execute the control used during this trajectory was calculated to be approximately \oldr{0.0034} kg over 3 days, which is a reasonable quantity within its fuel reserves. The properties of the \tr{Hayabusa2} spacecraft used were an ISP of \oldr{3000} seconds and a spacecraft mass of \oldr{600} kg \cite{hayion1,hayion2}. The equation used was $\Delta m = m_{\text{sc}} \cdot \sum \left( u_{\text{vec}} \right) \cdot \Delta t \cdot \frac{1}{\text{Isp} \cdot g_0}$ where \( m_{\text{sc}} = \oldr{600} \, \text{kg} \) is the spacecraft mass, \( u_{\text{vec}} \) is the total acceleration imparted (in \(\text{m/s}^2\)), \( \Delta t \) is the time step, \( \text{Isp} = \oldr{3000} \, \text{s} \) is the specific impulse, and \( g_0 = 9.80665 \, \text{m/s}^2 \) is the standard acceleration due to gravity.

\newpage
\subsubsection{Monte Carlo Analysis}\label{mc1}
\begin{figure}[htbp!]
\centering

% --- Left plot ---
\begin{minipage}[b]{0.3\textwidth}
    \centering
    % trim = {left bottom right top}
    \includegraphics[width=\linewidth,trim={0cm 0cm 26cm 0cm},clip]{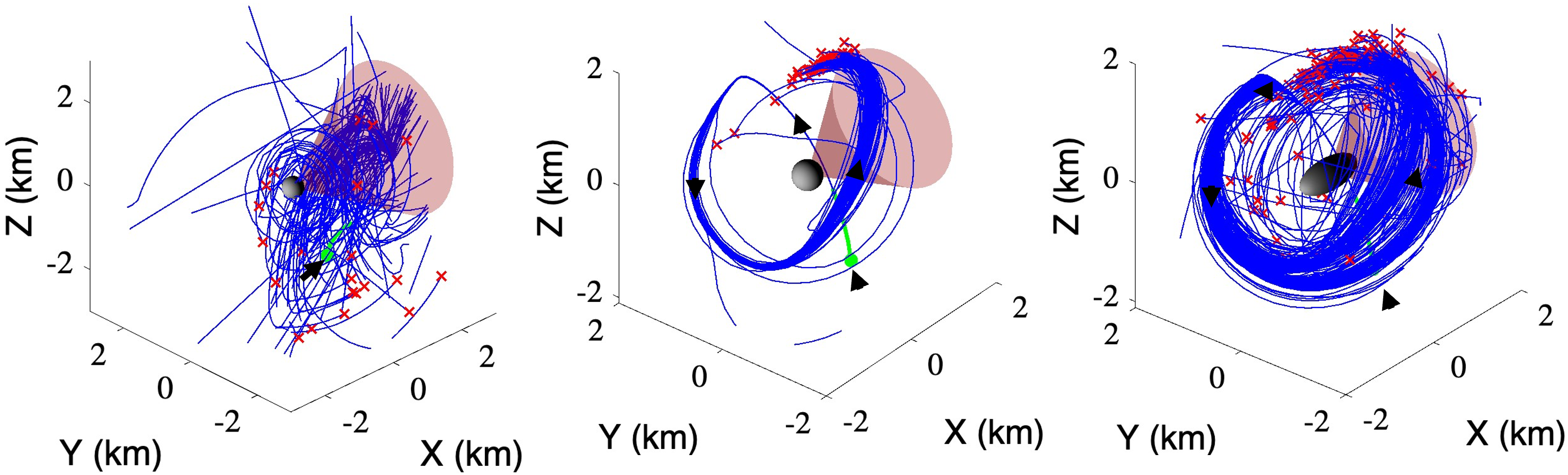}
    \vspace{0.3em}
    {\small (a) Without observability-constrained control}
\end{minipage}
\hfill
% --- Middle plot ---
\begin{minipage}[b]{0.3\textwidth}
    \centering
    \includegraphics[width=\linewidth,trim={13cm 0cm 12cm 0cm},clip]{mccollage.png}
    \vspace{0.3em}
    {\small (b) Constrained (Spherical primary)}
\end{minipage}
\hfill
% --- Right plot ---
\begin{minipage}[b]{0.3\textwidth}
    \centering
    \includegraphics[width=\linewidth,trim={25cm 0cm 0cm 0cm},clip]{mccollage.png}
    \vspace{0.3em}
    {\small (c) Constrained ([2.5 1 1] Ellipsoid primary)}
\end{minipage}

\caption{\oldr{Stationkeeping scenario.} Monte Carlo simulation of 200 true trajectories
using (a) basic Lyapunov controller, (b) observability-constrained Lyapunov controller
with spherical primary, and (c) with [2.5\,1\,1] ellipsoid primary. Initial position
(green circle), final position (red cross), and orbital direction (black arrows) are
indicated.}
\label{fig:mccollage}
\end{figure}

\begin{figure}[htbp!]
    \centering
    \includegraphics[width=0.99\linewidth]{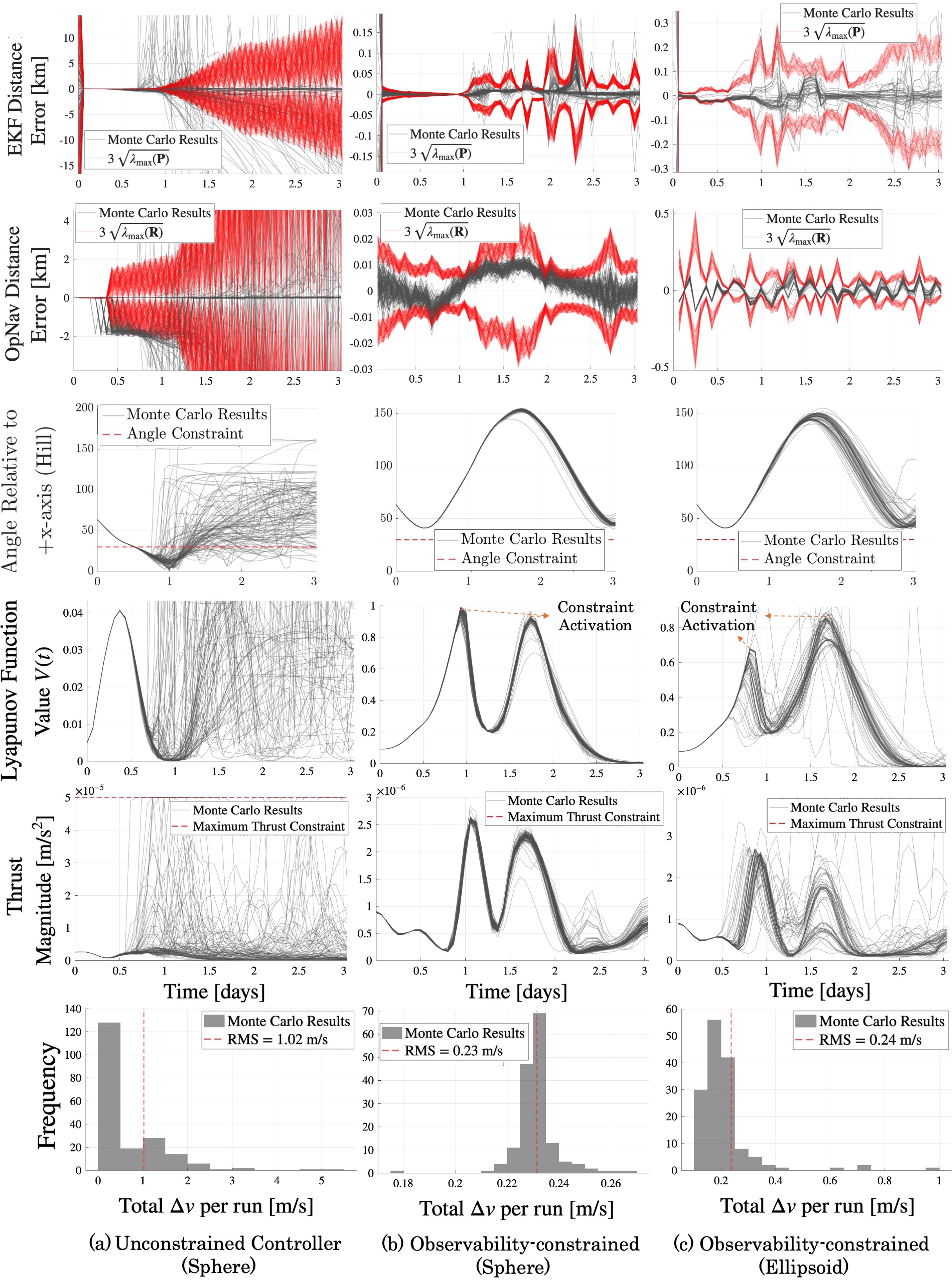}
    \caption{\oldr{Stationkeeping scenario. Monte Carlo Simulation of EKF Position \tr{Norm Error} (Row 1), OpNav Measurement \tr{Norm Error} (Row 2) and Angle Constraint Satisfaction (Row 3) with 200 Trials using Basic Lyapunov Controller (Column 1), Observability-constrained Lyapunov Controller with Spherical (Column 2) and [2.5 1 1] Ellipsoid Primary (Column 3).}}
    \label{fig:set1}
\end{figure}

To demonstrate the algorithm's robustness, a Monte Carlo analysis is conducted to analyze its success rate statistically. A \oldr{pessimistic} initial standard deviation is defined for the position error across each direction ($x,y\text{ and }z$ in the Hill frame) of $\sigma=30$ meters, approximately 10\% of Bennu's radius\oldr{, using the error experienced during the OSIRIS-REx mission as a standard \cite{Antreasian2019,Antreasian2022}}. This is used to apply an initial error to the true initial state to generate the EKF's initial known state. The Monte Carlo is run for 200 iterations. To start, we test the version without observability constraints to understand the severity of failures in this scenario. As per the results, the Lyapunov controller fails in almost every case without the observability constraint, as it passes through the poor observability region in every test. It succeeds in 8 out of 200 iterations for a success rate of 4\%. The EKF and true dynamics diverge completely and lead to the spacecraft either crashing into the \tr{small body} or traveling out of the system. Now, we compare this with the observability-constrained case. The observability-constrained Lyapunov controller shows better performance and success rate of the spacecraft reaching its desired orbit. It succeeds in 196 out of 200 iterations for a success rate of 98\%. This is a major performance improvement and supports the utility of the observability-constrained controller. The ellipsoid case succeeds 161 out of 200 iterations for a success rate of 80.5\%. The performance degrades with nonspherical objects but maintains an improvement over the case without an observability constraint. All \oldr{trajectory} Monte Carlo results are shown in \cref{fig:mccollage}.
\\\\
\oldr{Additionally, the evolution of EKF position \tr{Norm Error}, OpNav measurement \tr{Norm Error}, angle constraint satisfaction, \tr{Lyapunov function value, thrust magnitude over time and a $\Delta v$ consumption histogram} across the unconstrained, constrained with sphere and constrained with ellipse test cases are displayed in \cref{fig:set1}. \tr{Note that these plots depict error by subtracting truth distance norm from the EKF distance norm or OPNAV measurement norm respectively, leading to many diverging cases for the basic Lyapunov controller that have a negative norm error value due to the spacecraft leaving the system in truth. Likewise, the error bounds are constructed using the EKF position covariance estimate or OpNav analytical covariance.} The error magnitude in the unconstrained case is significantly higher than the constrained cases, and has considerable divergence due to poor measurements as the trajectory violates the angle constraint. Meanwhile, the ellipsoid with constraint performs significantly better than the unconstrained case, but experiences slightly more deviation and error than the sphere case with periodic variations in the measurement from the \tr{small body}'s rotation. In \tr{the observability-maintaining} cases, the EKF is mostly bounded by the \tr{$3\sqrt{\lambda_{max}(P_{\boldsymbol{r}})}$} bounds with few outliers, supporting its proper performance. \tr{However, the EKF covariance poorly bounds the errors in the basic controller case, corresponding to a deviation between the truth and EKF position estimate, which is concordant with our earlier definition of divergence.} Furthermore, the constraint\tr{s} \tr{are} correctly satisfied by the \tr{observability-maintaining} Lyapunov controller as the unconstrained scenario enters the dark side cone with an angle lower than $30^\circ$ while the constrained cases remain consistently above. }\tr{For the unconstrained case, The Lyapunov function value starts at a relatively lower value before increasing due to the natural progression of the trajectory, after which the controller attempts to reach the target orbit, in the process of which it encounters divergence since it does not consider the angle constraint. Meanwhile, the constrained cases show a similar pattern with two peaks corresponding to the angle constraint activation, causing the controller to maintain trajectories that maintain observability and thereby minimizing divergence. This is concordant with the continuous thrust profiles depicting behaviour with two peaks corresponding to the angle constraint inflations in the Lyapunov function value. The unconstrained trajectory encounters the maximum thrust constraint due to divergence and therefore has a significantly higher $\Delta v$ usage, while the constrained trajectories have a reasonable thrust distribution.}

\subsection{Approach and Circularization Scenario}
\begin{figure}[htbp!]
\centering

% --- First row ---
\includegraphics[width=0.45\linewidth,trim={1cm 0.1cm 0 0.1cm},clip]{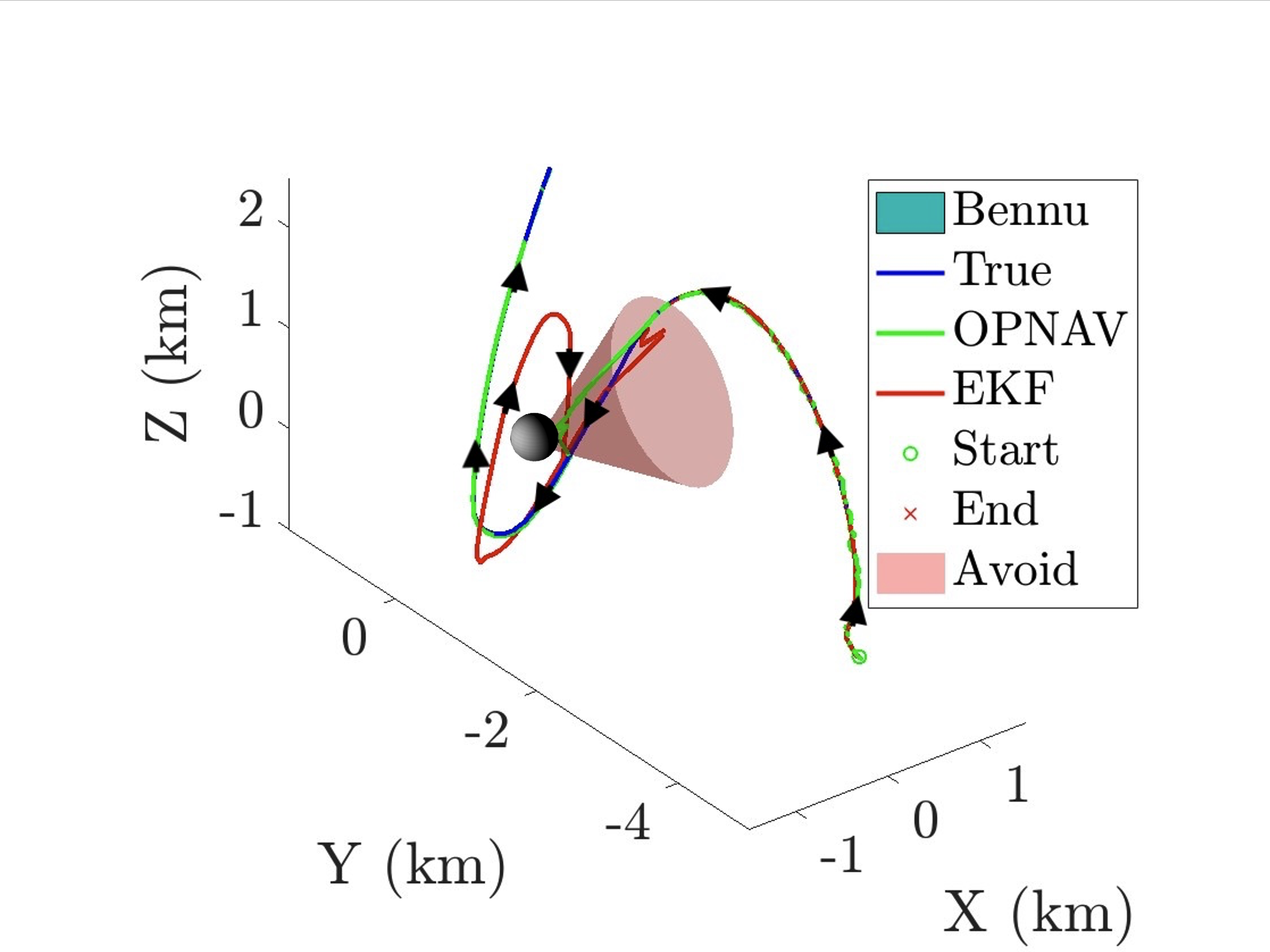}
\hfill
\includegraphics[width=0.45\linewidth,trim={1cm 0.1cm 0 0.1cm},clip]{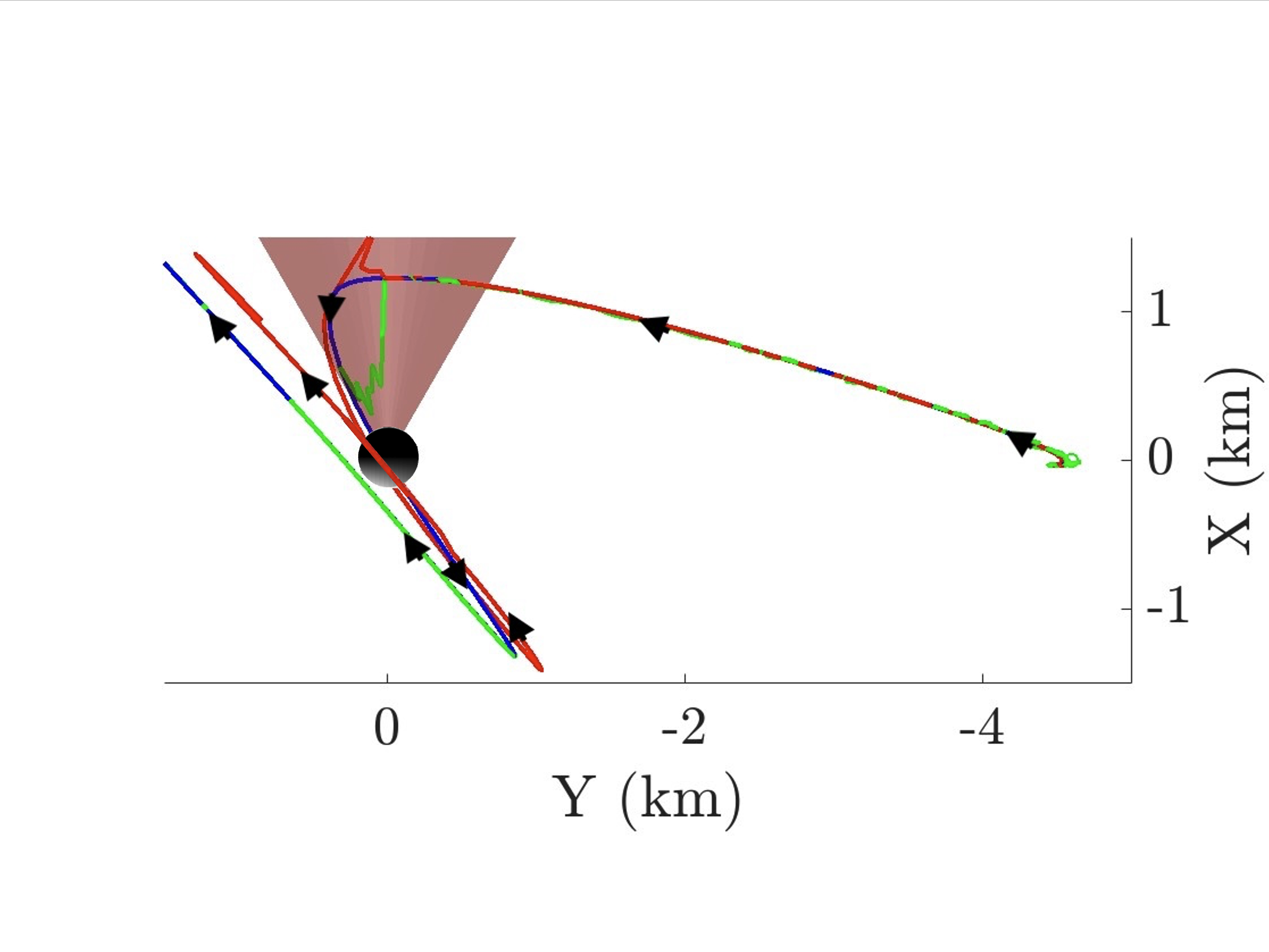}

\vspace{0.3em}
\makebox[0.45\linewidth][c]{\small(a) Trajectory (3D view)}%
\hfill
\makebox[0.45\linewidth][c]{\small(b) Trajectory (XY view)}%

\vspace{1em}

% --- Second row ---
\includegraphics[width=0.45\linewidth,trim={0cm 0.1cm 0 0.1cm},clip]{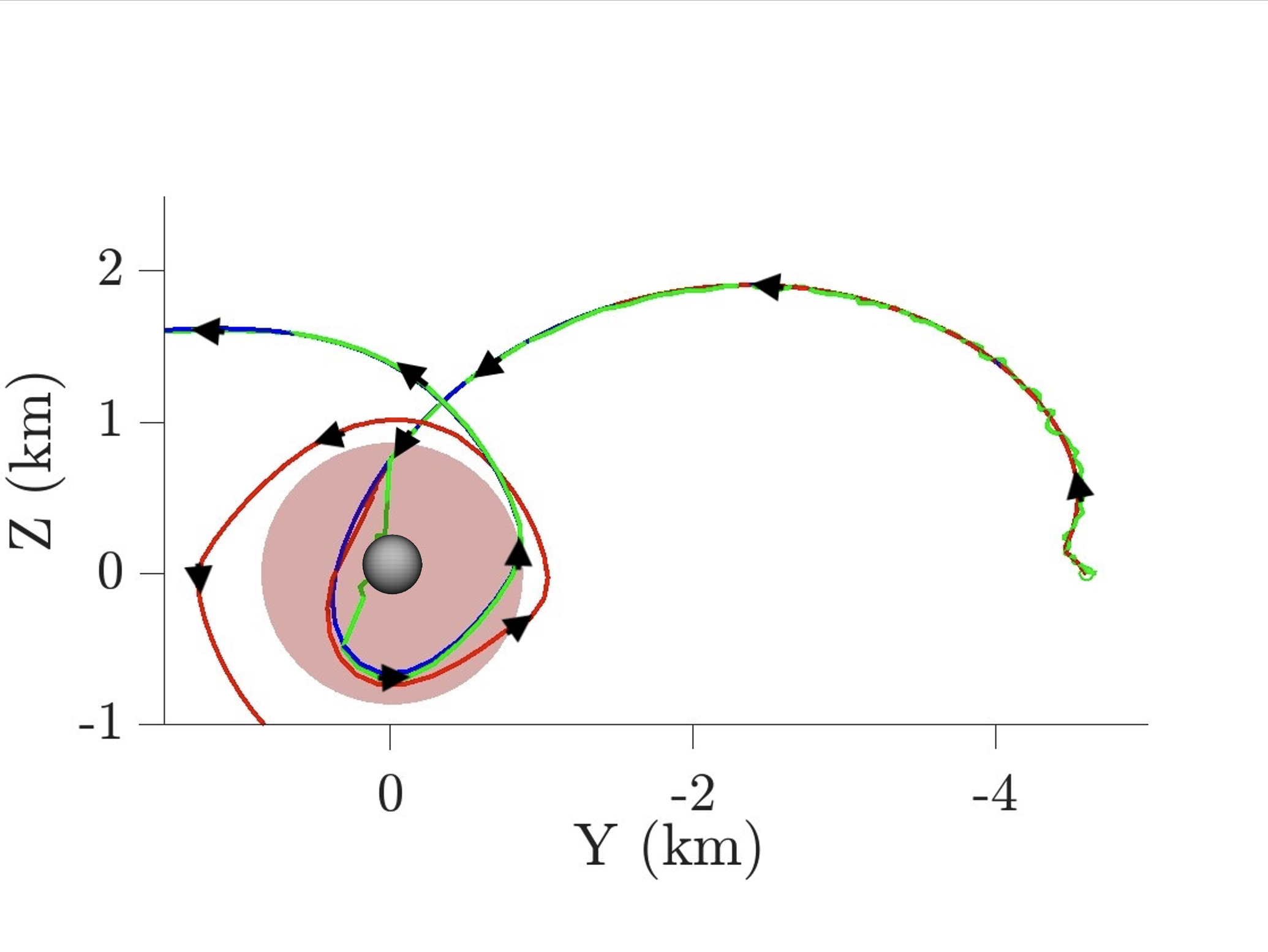}
\hfill
\includegraphics[width=0.45\linewidth,trim={1cm 0.1cm 0 0.1cm},clip]{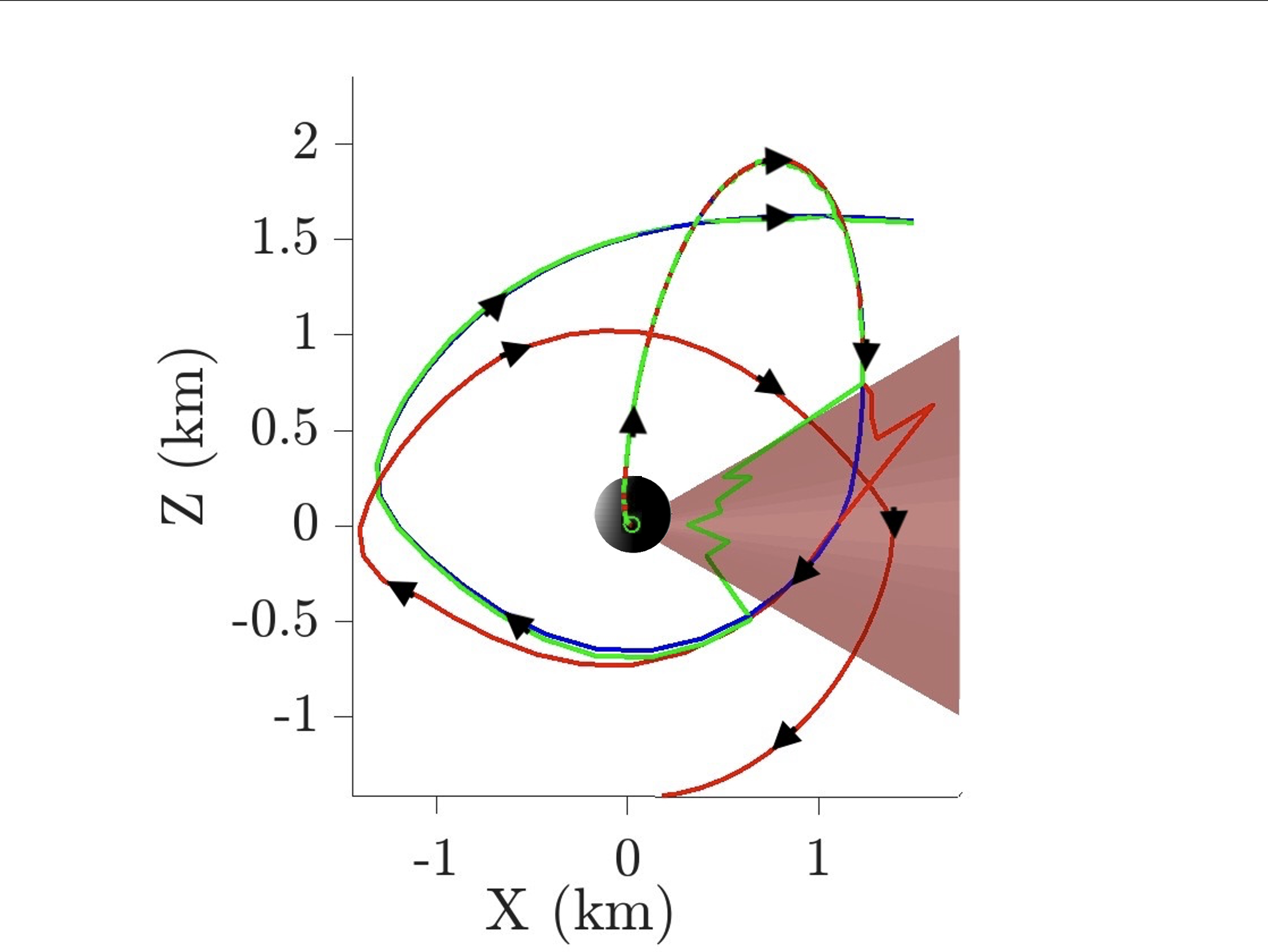}

\vspace{0.3em}
\makebox[0.45\linewidth][c]{\small(c) Trajectory (YZ view)}%
\hfill
\makebox[0.45\linewidth][c]{\small(d) Trajectory (XZ view)}%

\vspace{1em}

% --- Third row ---
\includegraphics[width=0.45\linewidth]{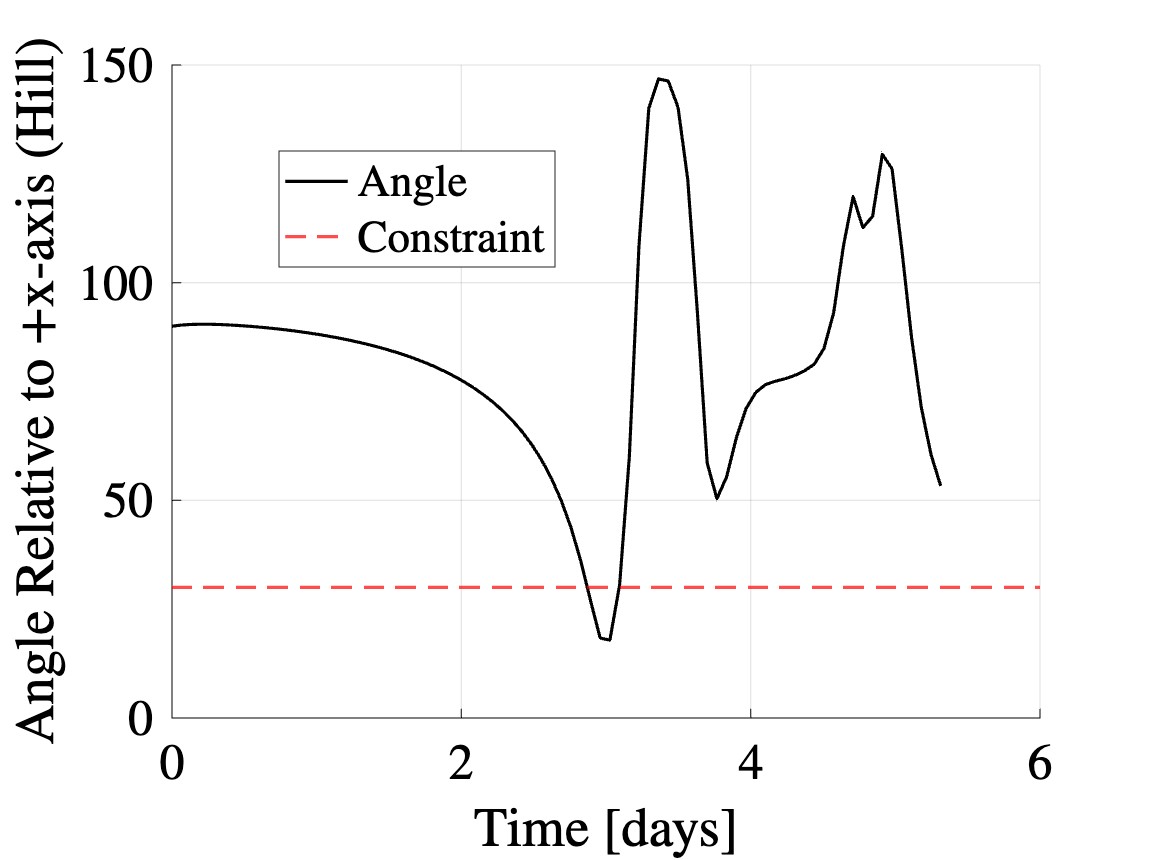}
\hfill
\includegraphics[width=0.45\linewidth]{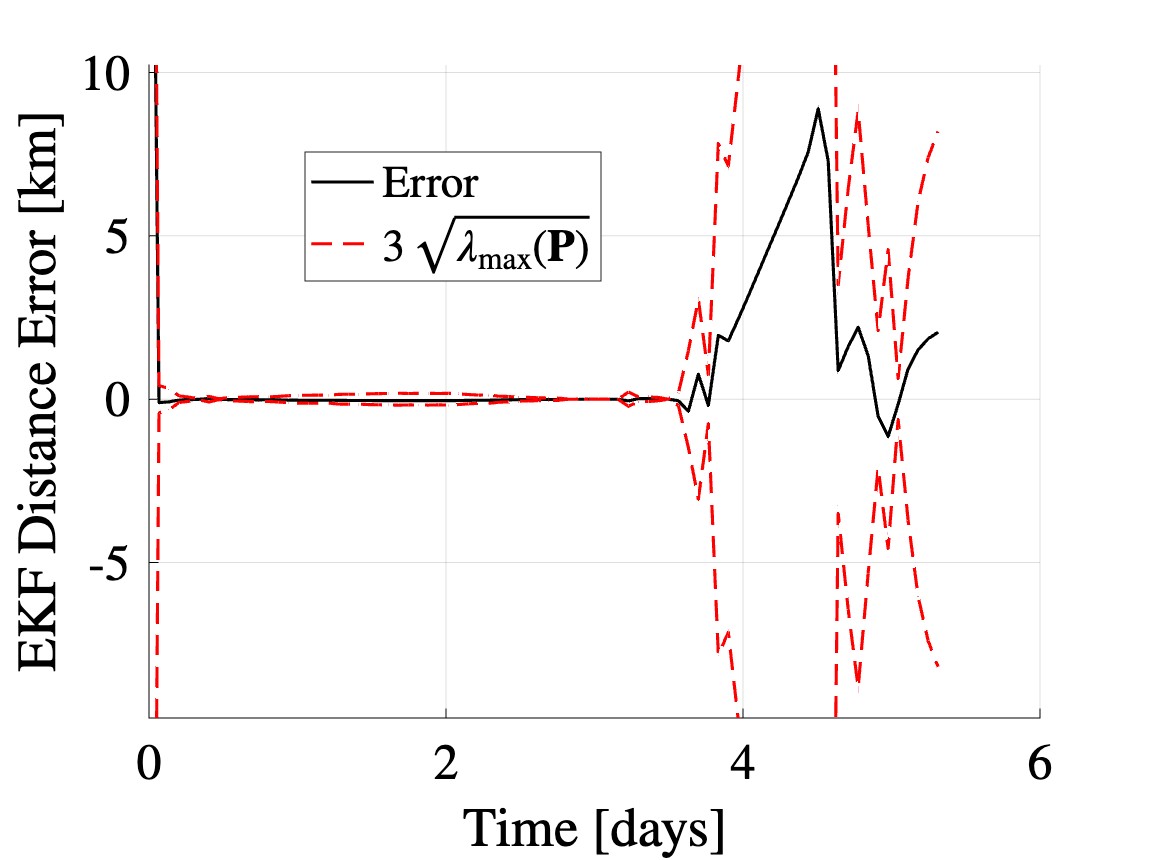}

\vspace{0.3em}
\makebox[0.45\linewidth][c]{\small(e) Angle to +x-axis (Hill) over time}%
\hfill
\makebox[0.45\linewidth][c]{\small(f) EKF distance error over time}%

\ocmt{Figure Updated with Smoother Trajectory}
\caption{\oldr{Approach and Circularization scenario \tr{without Observability-constrained
Penalty Lyapunov control}.} True (blue), OpNav measurement (green), EKF-estimated (red)
trajectories and keep-out cone (shaded red) shown in four views. Initial position (green
circle), final position (red cross) and orbital direction (black arrows) indicated.
\tr{+x-axis (Hill) relative angle and EKF distance error plots included to correlate
constraint violation and divergence.}}
\label{fig:np2}
\end{figure}

\begin{figure}[htbp!]
\centering

% --- First row ---
\includegraphics[width=0.45\linewidth,trim={0 0 0.1cm 0},clip]{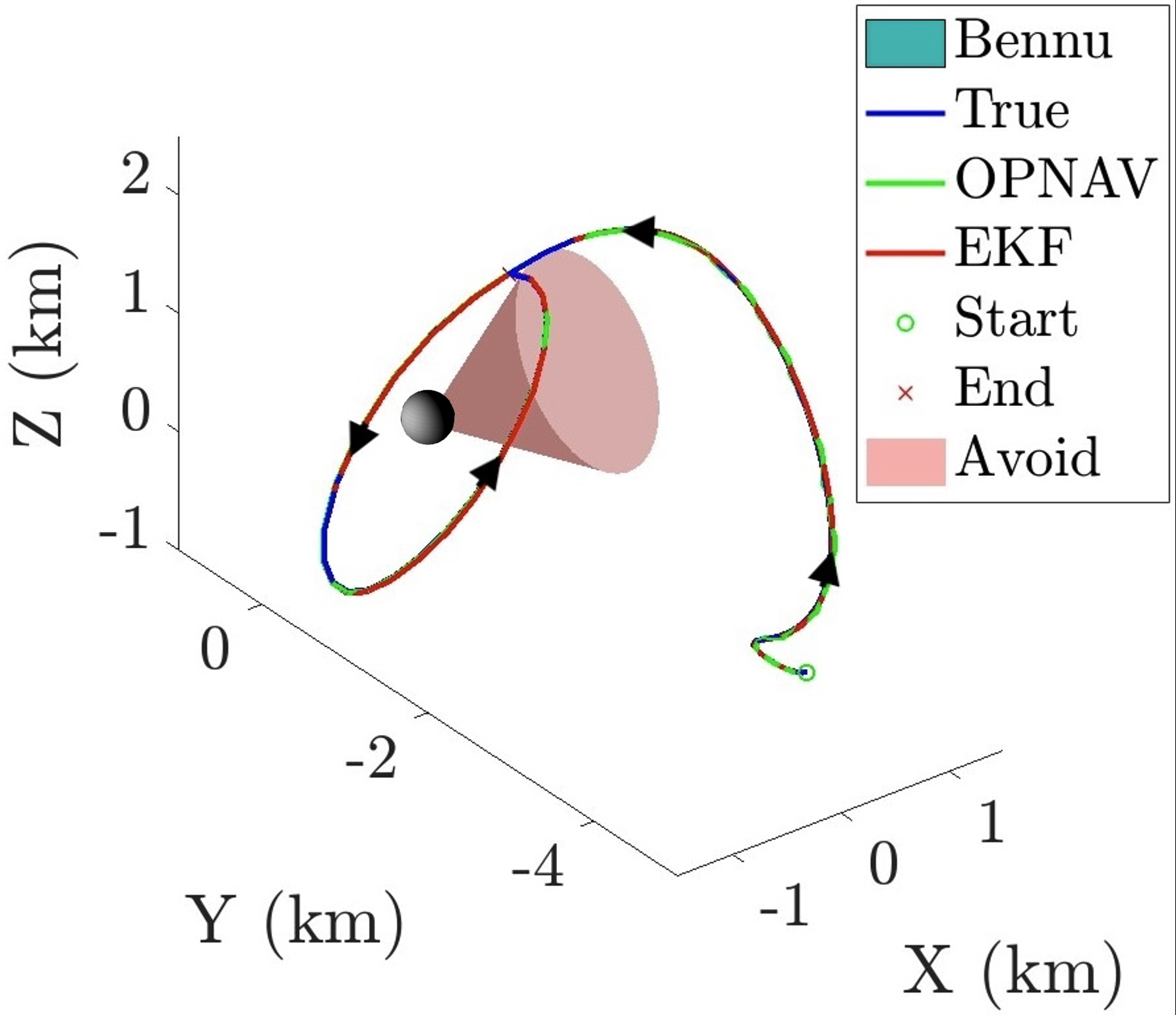}
\hfill
\includegraphics[width=0.45\linewidth]{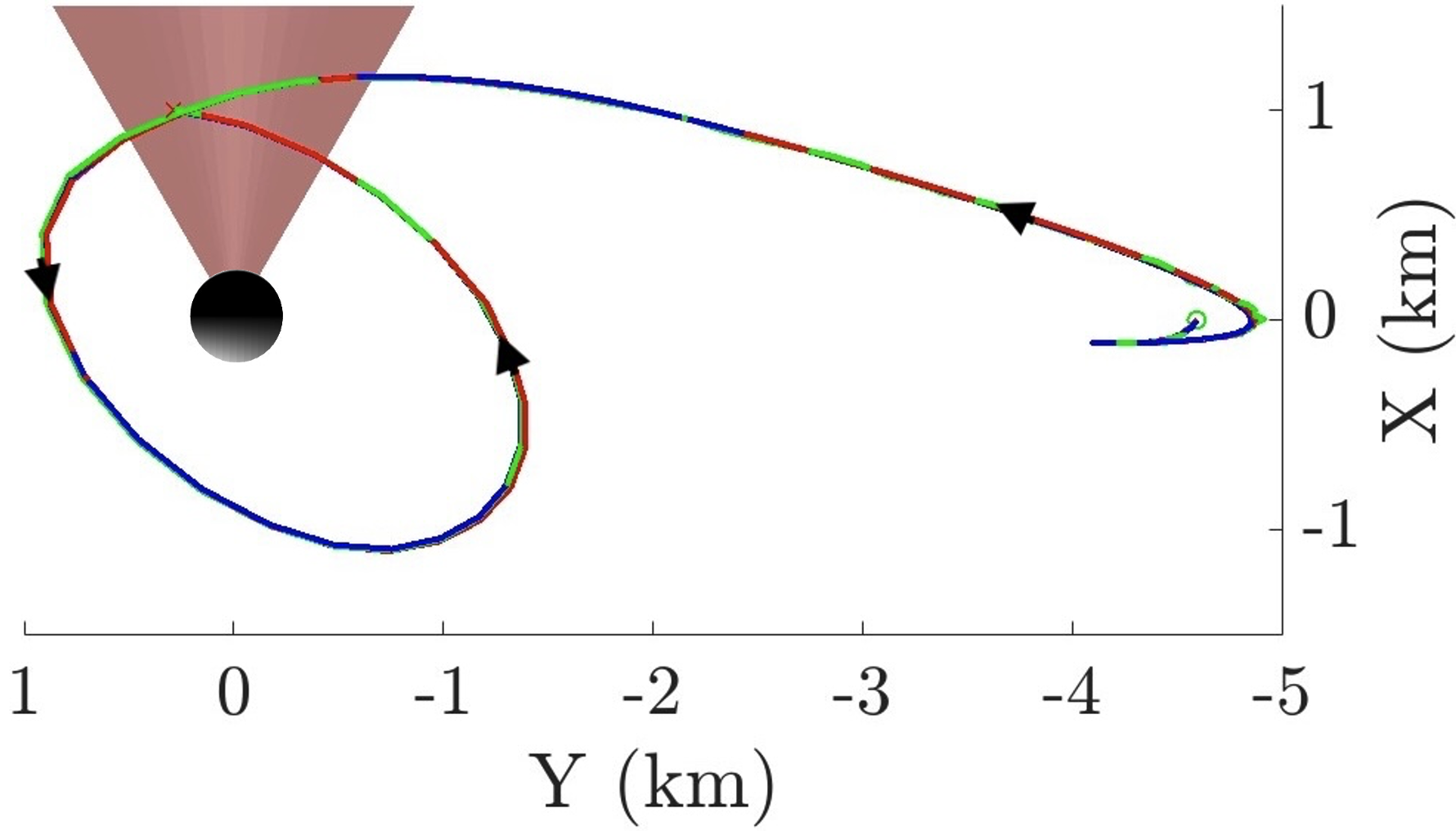}

\vspace{0.3em}
\makebox[0.45\linewidth][c]{\small(a) Trajectory (3D view)}%
\hfill
\makebox[0.45\linewidth][c]{\small(b) Trajectory (XY view)}%

\vspace{1em}

% --- Second row ---
\includegraphics[width=0.45\linewidth,trim={0cm 0.1cm 1cm 0.1cm},clip]{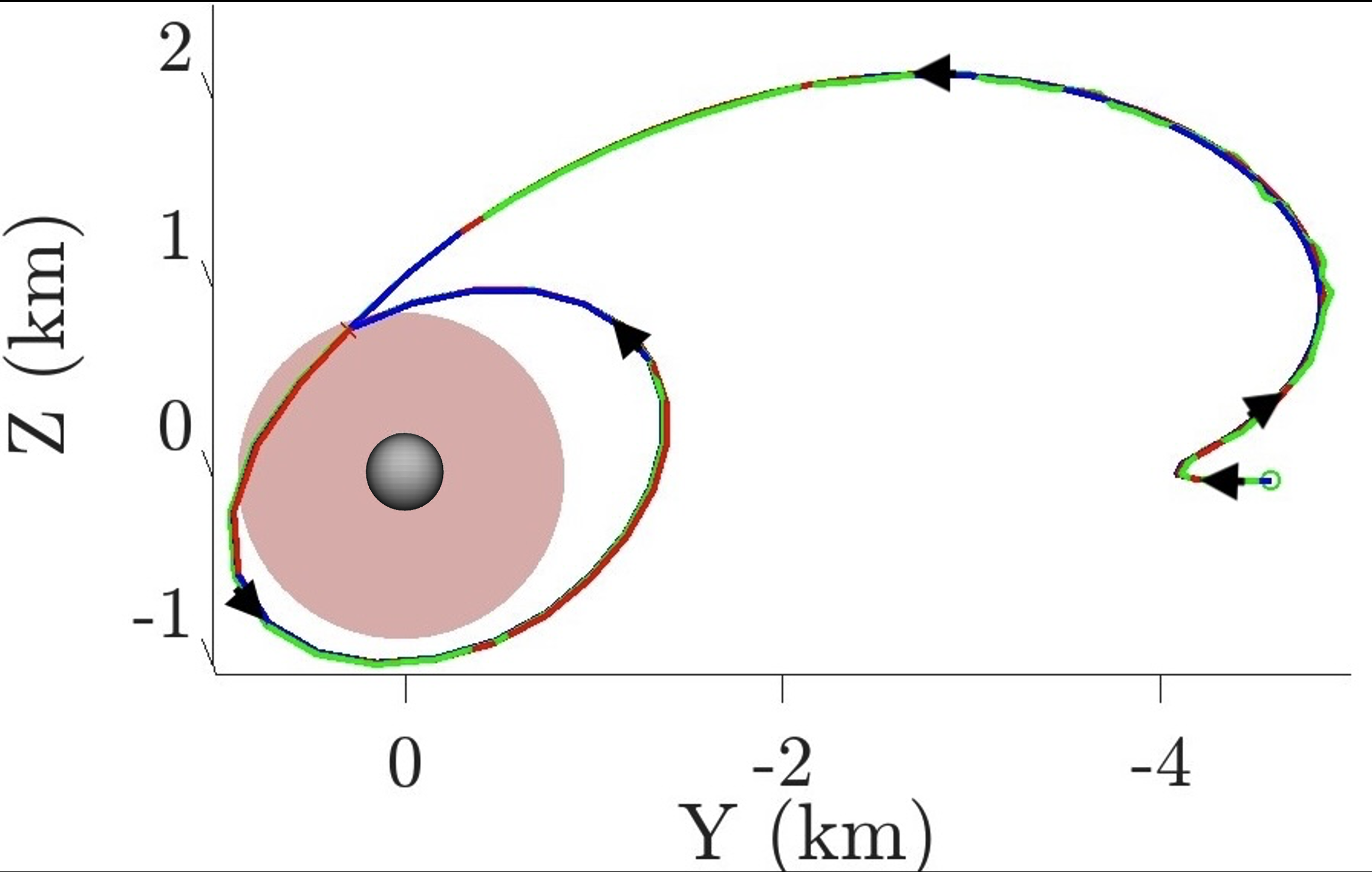}
\hfill
\makebox[0.45\linewidth][c]{%
    \includegraphics[width=0.35\linewidth]{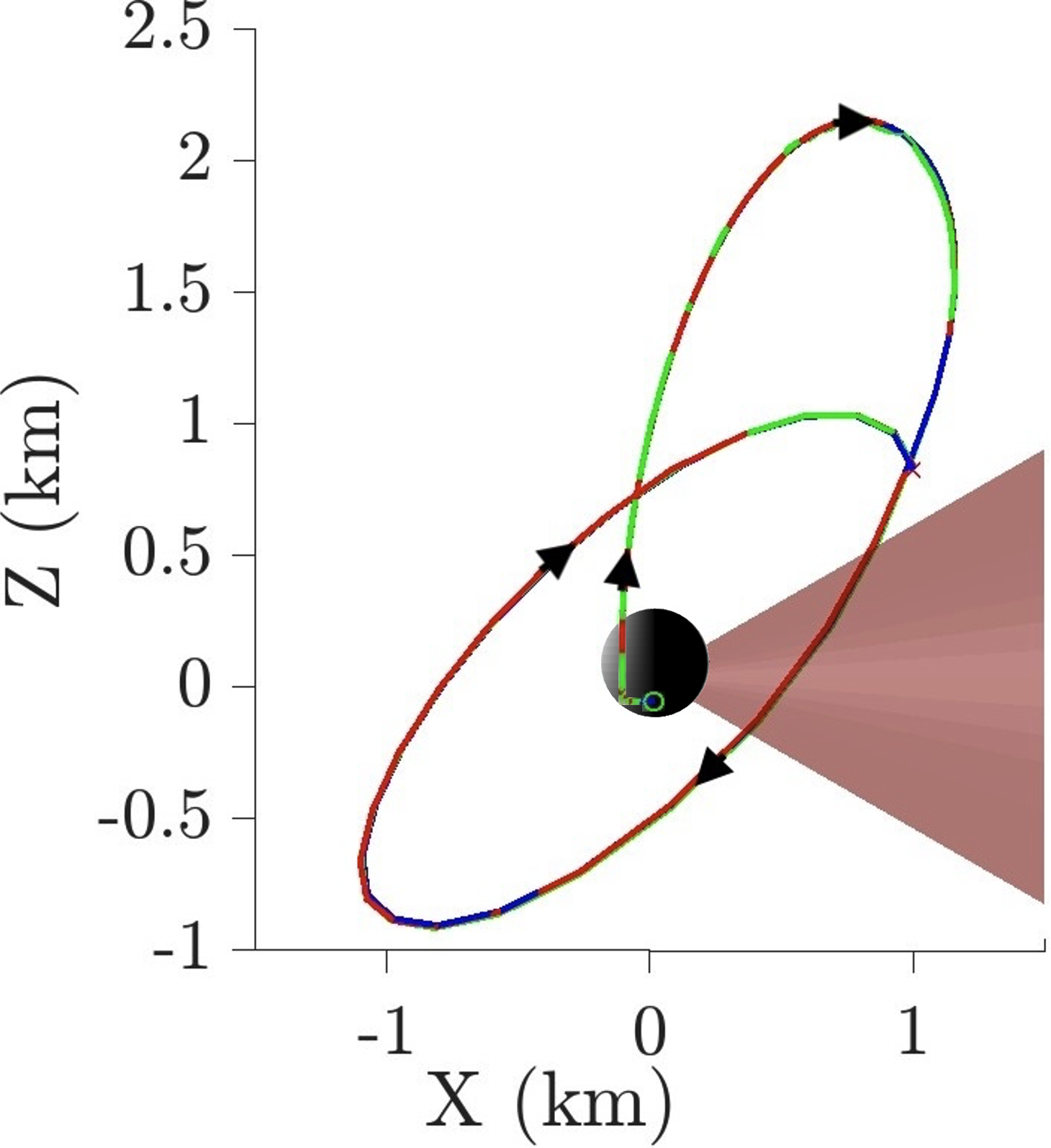}%
}

\vspace{0.3em}
\makebox[0.45\linewidth][c]{\small(c) Trajectory (YZ view)}%
\hfill
\makebox[0.45\linewidth][c]{\small(d) Trajectory (XZ view)}%

\vspace{1em}

% --- Third row ---
\includegraphics[width=0.45\linewidth]{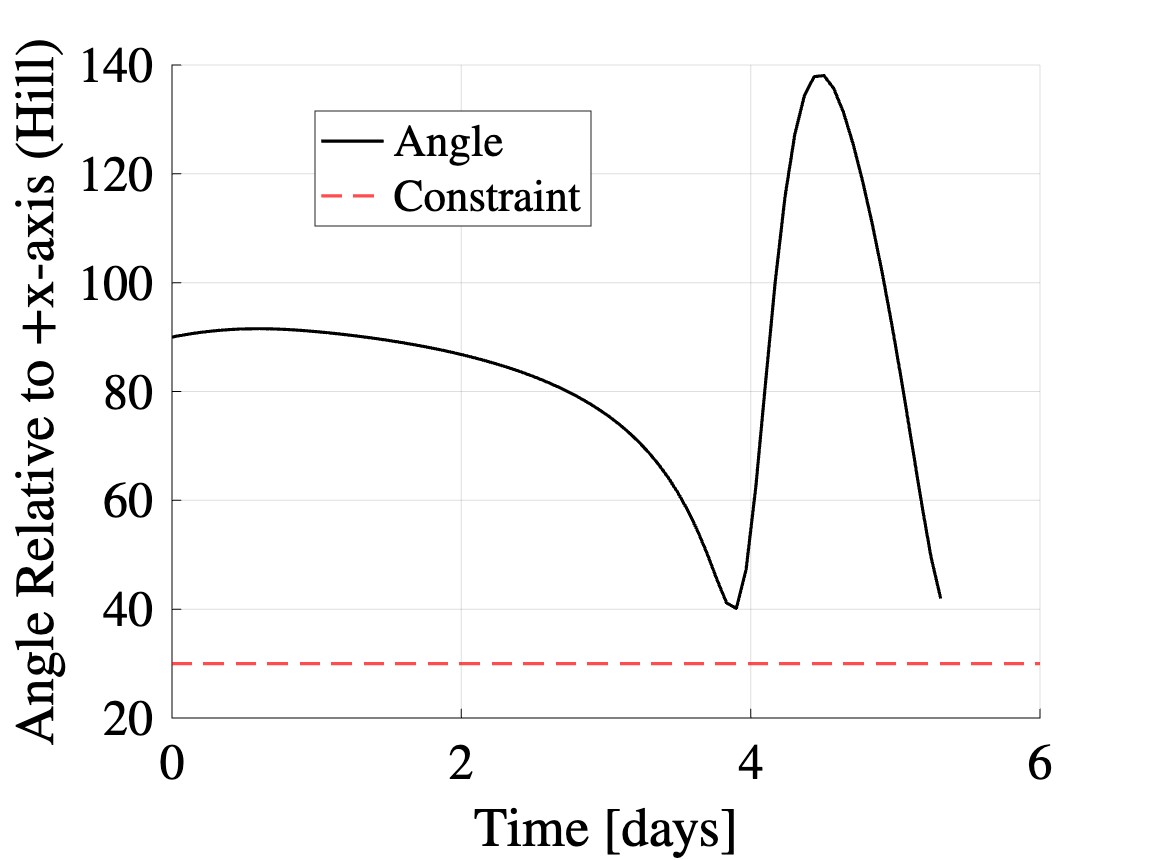}
\hfill
\includegraphics[width=0.45\linewidth]{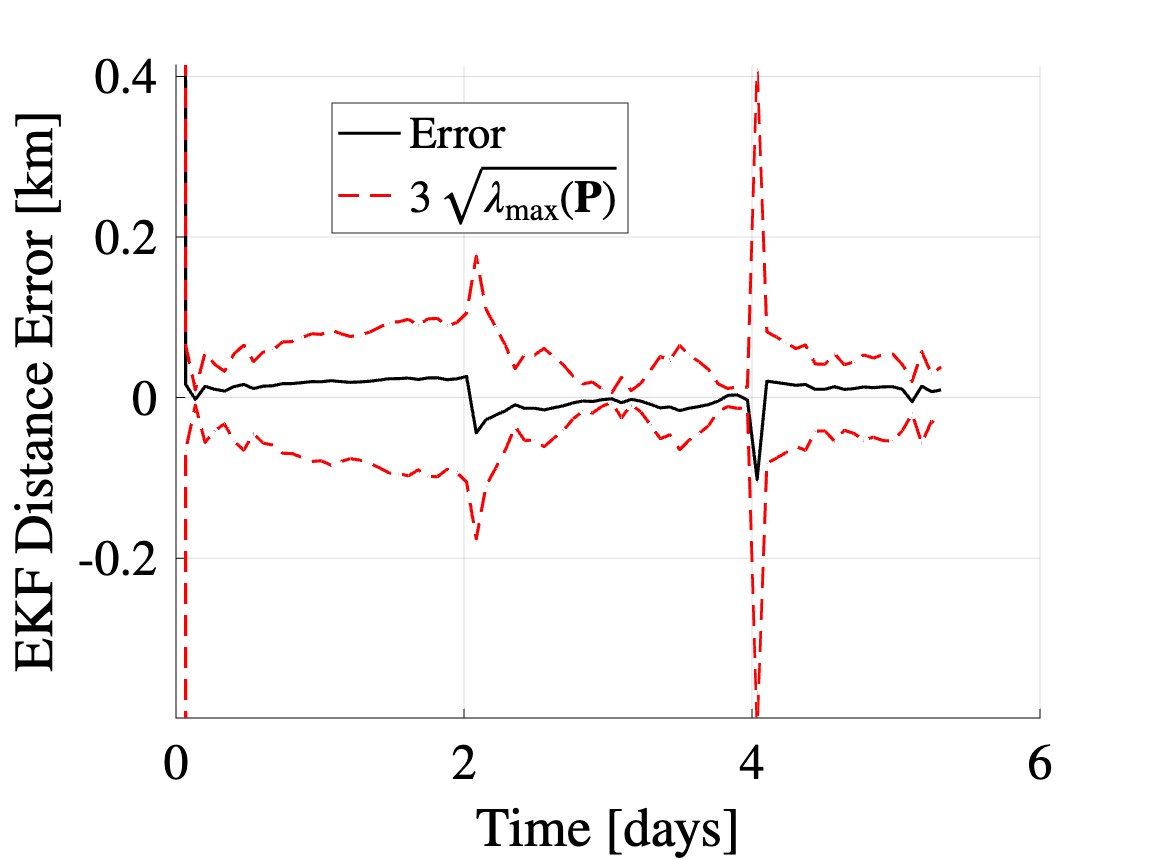}

\vspace{0.3em}
\makebox[0.45\linewidth][c]{\small(e) Angle to +x-axis (Hill) over time}%
\hfill
\makebox[0.45\linewidth][c]{\small(f) EKF distance error over time}%

\caption{\oldr{Approach and Circularization scenario \tr{with Observability-constrained
Penalty Lyapunov control}.} True (blue), OpNav measurement (green), EKF-estimated (red)
trajectories and keep-out cone (shaded red) with Observability-constrained Penalty
Lyapunov control shown in four views. Initial position (green circle), final position
(red cross) and orbital direction (black arrows) indicated. \tr{+x-axis (Hill) relative
angle and EKF distance error plots included to correlate constraint violation and
divergence.}}
\label{fig:wp2}
\end{figure}

\begin{figure}[htbp!]
\centering

% --- First image ---
\begin{minipage}[b]{0.3\textwidth}
    \centering
    \includegraphics[width=\linewidth,trim={0cm 0cm 22cm 0cm},clip]{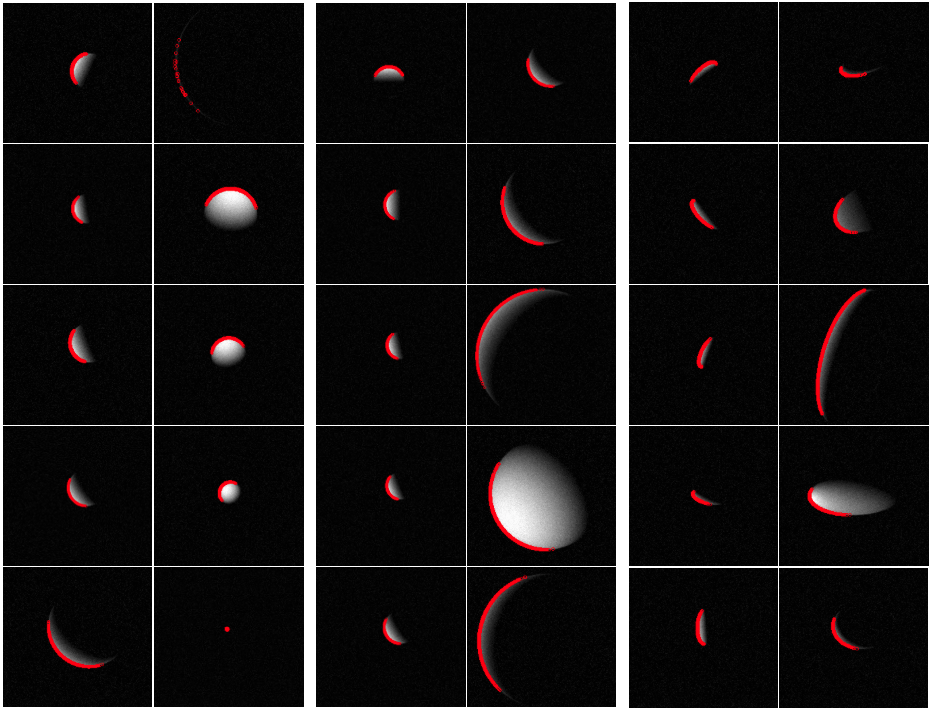}
    \vspace{0.3em}
    {\small (a) Without observability-constrained control}
\end{minipage}
\hfill
% --- Second image ---
\begin{minipage}[b]{0.3\textwidth}
    \centering
    \includegraphics[width=0.9\linewidth,trim={11.5cm 0cm 11.5cm 0cm},clip]{bcollage2.png}
    \vspace{0.3em}
    {\small (b) Constrained (Spherical primary)}
\end{minipage}
\hfill
% --- Third image ---
\begin{minipage}[b]{0.295\textwidth}
    \centering
    \includegraphics[width=\linewidth,trim={22cm 0cm 0cm 0cm},clip]{bcollage2.png}
    \vspace{0.3em}
    {\small (c) Constrained ([2.5 1 1] Ellipsoid primary)}
\end{minipage}

\caption{\oldr{Approach and Circularization scenario.} Sample image history with detected
horizon points from \cref{fig:np2} for case without observability-constrained control,
and \cref{fig:wp2} for cases with observability-constrained control using a spherical
primary and a [2.5\,1\,1] ellipsoid primary. Images are ordered top–bottom, then
left–right.}
\label{fig:gcol2}
\end{figure}

Beyond the orbit maintenance scenario, it is important to demonstrate the effectiveness of this controller in a greater variety of situations, including those starting further away from the \tr{small body}. Therefore, we test this algorithm in a specific test case where we approach the \tr{small body} from just within the effective range of horizon-based OpNav determined in \cref{distacc}. The controller aims to circularize around the \tr{small body} and executes maneuvers accordigly. The initial conditions in Cartesian terms are $^H\bm{r}=[0,0,-4.5964]^\top$ km and $^H\bm{v}=[-4.8927,0,2.4464]^\top$ mm/s. The same initial covariance is used as in \cref{oms}. \oldr{Using the tuning guidelines outlined in \cref{tuning},} the gain found to return the best performance for the controller in this particular test case is:
\begin{equation}K_\textrm{2}=\mathrm{diag}(10^{-3},10^{-3},10^{-3},10^{-3},10^{-7},10^{-7})\end{equation}
\\
The penalty function parameters chosen are identical to those from \cref{oms}, except the weight for the cone constraint is increased to $\omega_3$ = 100. This leads to better performance in avoiding the poor observability region at the larger scale of this scenario. 
\\\\
We repeat a comparison between the regular (\cref{fig:np2}) and observability-constrained (\cref{fig:wp2}) controllers. The figures depict the spacecraft's true trajectory, OpNav measurement, EKF position estimate and a visualization of the keep-out cone. The result matches the outcome from \cref{oms} with the observability-constrained controller avoiding the keep-out cone and successfully circularizing around the \tr{small body}. The regular controller's true and EKF-estimated trajectories diverge since it enters the keep-out cone and produces faulty measurements. An additional test with the observability-constrained controller was executed with a [2.5 1 1] ellipsoid using the rotation parameters of Bennu and an approximation from its known mean radius.
\\\\
Camera snapshots of each case with the detected horizon points highlighted in red can be seen in \cref{fig:gcol2}. The observations are mostly similar to those from \cref{oms}. The difference is that the lit limb is much smaller at the start of the simulation and grows as the spacecraft approaches the \tr{small body}. The measurement and controller perform well under such variable conditions.
\\\\
\oldr{For this trajectory, the average control acceleration was $2.6889\times10^{-7}\,\textrm{m/s}^{2}$, with a maximum instantaneous acceleration of $6.6203\times10^{-6}\,\textrm{m/s}^{2}$ and total $\Delta v=0.1234\,\textrm{m/s}$.} The maximum fuel consumption to execute the control used during this trajectory was calculated to be approximately \oldr{0.0025} kg over 5 days, which is a reasonable quantity within \oldr{Hayabusa2'}s fuel reserves. This calculation was done using the same method from \cref{oms}.

\newpage
\subsubsection{Monte Carlo Analysis}
\begin{figure}[htbp!]
\centering

% --- Left plot ---
\begin{minipage}[b]{0.25\textwidth}
    \centering
    % trim = {left bottom right top}
    \includegraphics[width=\linewidth,trim={0cm 0cm 36.75cm 0cm},clip]{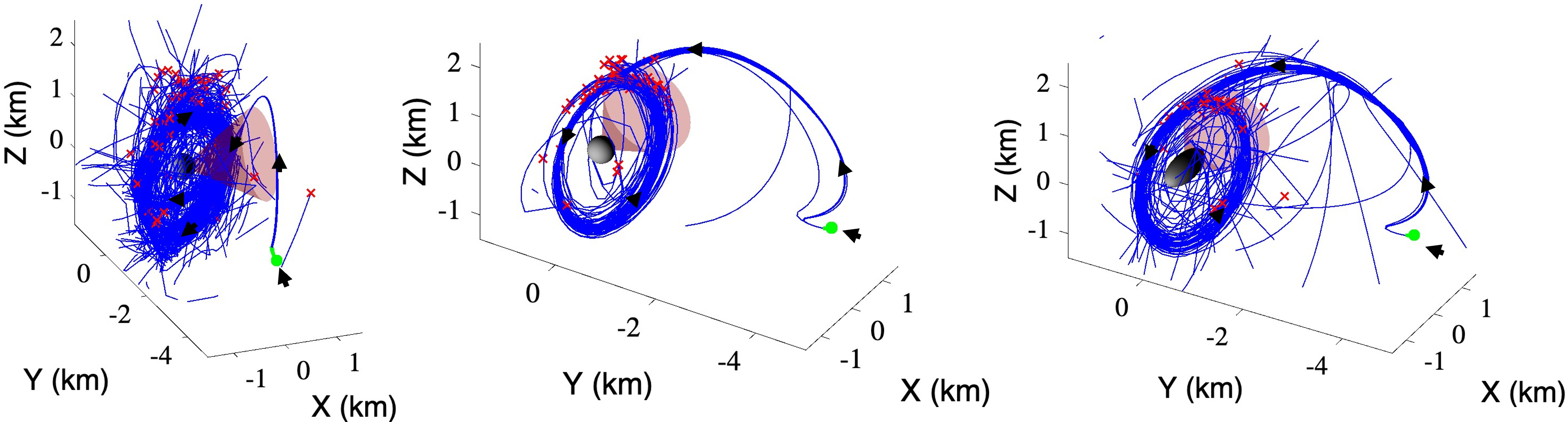}
    \vspace{0.3em}
    {\small (a) Unconstrained}
\end{minipage}
\hfill
% --- Middle plot ---
\begin{minipage}[b]{0.3\textwidth}
    \centering
    \includegraphics[width=\linewidth,trim={12.5cm 0cm 18.1cm 0cm},clip]{mccollage2.png}
    \vspace{0.3em}
    {\small (b) Constrained (Spherical primary)}
\end{minipage}
\hfill
% --- Right plot ---
\begin{minipage}[b]{0.3\textwidth}
    \centering
    \includegraphics[width=\linewidth,trim={31.1cm 0cm 0cm 0cm},clip]{mccollage2.png}
    \vspace{0.3em}
    {\small (c) Constrained ([2.5 1 1] Ellipsoid primary)}
\end{minipage}

\caption{\oldr{Approach and Circularization scenario.} Monte Carlo simulation of 200 true trajectories
using (a) basic Lyapunov controller, (b) observability-constrained Lyapunov controller
with spherical primary, and (c) with [2.5\,1\,1] ellipsoid primary. Initial position
(green circle), final position (red cross), and orbital direction (black arrows) are
indicated.}
\label{fig:mccollage2}
\end{figure}

\begin{figure}
    \centering
    \includegraphics[width=1\linewidth]{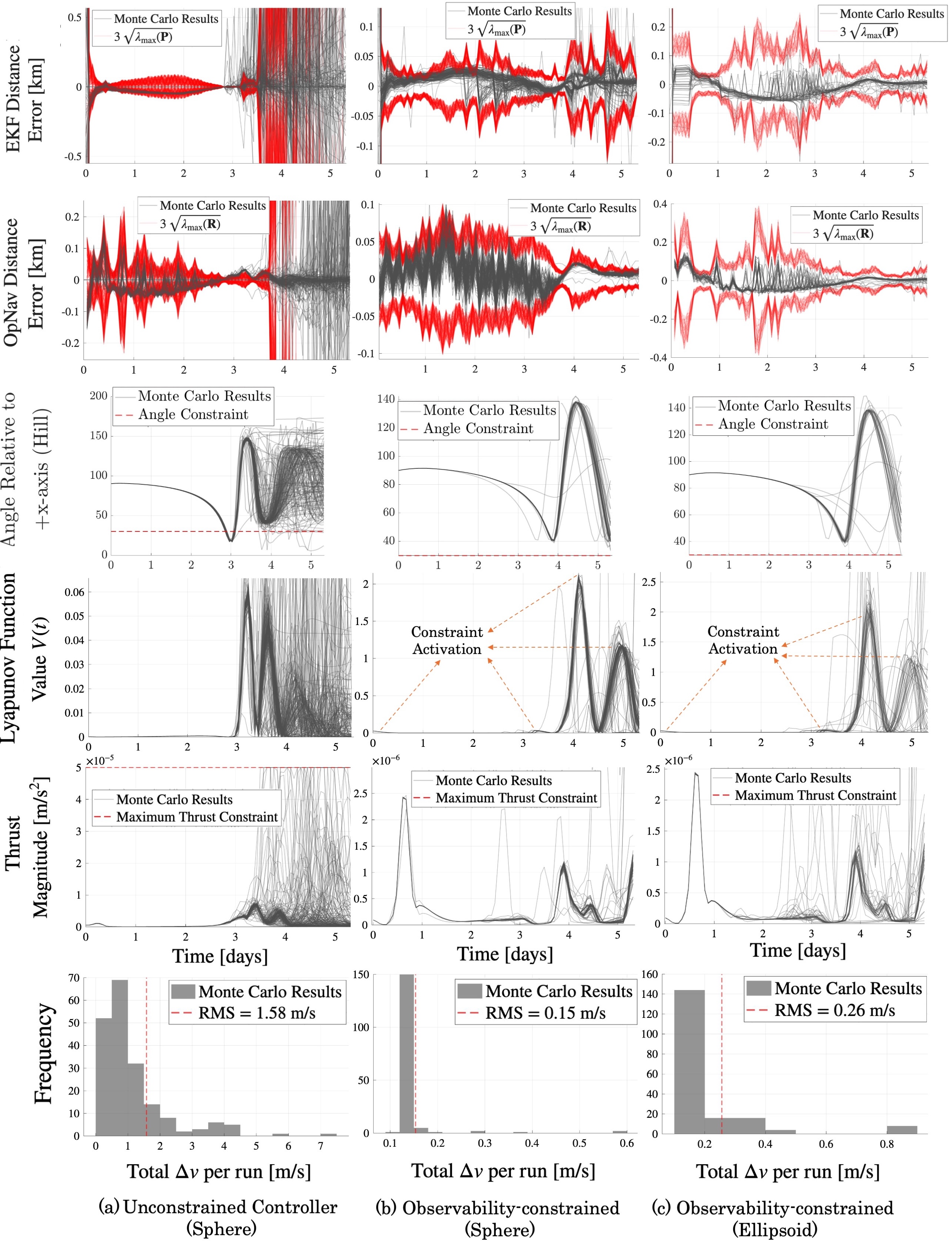}
    \caption{\oldr{Approach and Circularization scenario. Monte Carlo Simulation of EKF Position \tr{Norm Error} (Row 1), OpNav Measurement \tr{Norm Error} (Row 2) and Angle Constraint Satisfaction (Row 3) with 200 Trials using Basic Lyapunov Controller (Column 1), Observability-constrained Lyapunov Controller with Spherical (Column 2) and [2.5 1 1] Ellipsoid Primary (Column 3).}}
    \label{fig:set2}
\end{figure}

The monte carlo analysis is repeated for the approach and circularization scenario. The same settings from \cref{mc1} are used, with initial position standard deviation of $\sigma=30$ meters and 200 iterations. The regular controller succeeds in 68 out of 200 iterations for a success rate of 34\%. The EKF and true dynamics diverge often and lead to the spacecraft either crashing into the \tr{small body} or traveling out of the system. The observability-constrained Lyapunov controller shows improved performance with a success rate of 181 out of 200 iterations for a success rate of 90.5\%. Meanwhile, the same case with a [2.5 1 1] ellipsoid succeeds for 168 out of 200 iterations for a success rate of 84\%. This supports the consistent performance benefit of the observability-constrained controller across multiple scenarios. The controller is shown to be useful in approach scenario at larger distance, where the ellipsoid assumption better holds. While the performance degrades for the ellipsoid test in this case as well, it maintains better performance than the regular controller. All \oldr{approach and circularization} Monte Carlo results are shown in \cref{fig:mccollage2}.
\\\\
\oldr{Similarly as the previous scenario, the evolution of EKF position \tr{Norm Error}, OpNav measurement \tr{Norm Error} and angle constraint satisfaction over time across the unconstrained, constrained with sphere and constrained with ellipse test cases are displayed in \cref{fig:set2}. All of the trajectories in the unconstrained case violate the angle constraint, which corresponds to a significantly higher measurement error than the constrained cases with poorly characterized covariance. Thus, this leads to significant divergence in the unconstrained case. Both constrained cases offer significant improvements, with the results for the spherical body being better than the ellipsoid body due to lower error magnitude. \tr{However, the EKF covariance poorly bounds the errors in the basic controller case, corresponding to a deviation between the truth and EKF position estimate, which is concordant with our earlier definition of divergence.} Furthermore, the constraint\tr{s} \tr{are} correctly satisfied by the \tr{observability-maintaining} Lyapunov controller as the unconstrained scenario enters the dark side cone with an angle lower than $30^\circ$ while the constrained cases remain consistently above.} \tr{For the unconstrained case, the Lyapunov function value starts at a low value before increasing due to the natural progression of the trajectory, after which the controller attempts to reach the target orbit, in the process of which it encounters divergence since it does not consider the angle constraint. Meanwhile, the constrained cases start with a higher  initial value allowing the controller to apply initial input to avoid violating the angle constraint, which then leads to two peaks (relatively delayed compared to the unconstrained peaks) that ensure further angle constraint satisfaction. This allows the controller to maintain trajectories that maintain observability, thereby minimizing divergence. This is concordant with the continuous thrust profiles depicting behaviour with an initial control peak, following by final circularization and angle constraint satisfaction inputs. The unconstrained trajectory encounters the maximum thrust constraint due to divergence and therefore has a significantly higher $\Delta v$ usage, while the constrained trajectories have a reasonable thrust distribution.}
\pagebreak
\section{Discussion}\label{disc}
\tr{The numerical results demonstrate that incorporating observability path constraints into a Lyapunov controller can improve autonomous asteroid OpNav performance compared to an unconstrained baseline.} The developed controller \tr{maintains OpNav observability while considering the perturbation sensitivities of trajectories in low-gravity small-body environments}. When the spacecraft is positioned behind the asteroid's dark side, there is poor observability for the system due to an undetectable horizon, potentially leading to inaccurate state estimations. The observability-constrained controller concept allows the design of a control scheme that ensures a spacecraft follows a trajectory that avoids these faulty measurements. This method is better than simply discarding poor measurements, which may lead to an extended gap without measurements, causing the state estimate to diverge. 
\\\\
\tr{In both orbit-maintenance and approach–circularization scenarios, the constrained controller consistently maintained spacecraft trajectories within the observability and safety region, avoiding cone, range or collision violations observed in the unconstrained case. This stability was preserved across spherical and rotating ellipsoidal bodies, even under significant initial dispersions and geometry variations.}
\\\\
\tr{Across both testing scenarios, orbit maintenance and approach \& circularization, the constrained controller satisfied the keep-out cone and range constraints throughout successful runs, while the unconstrained Lyapunov baseline frequently violated the angle constraint and diverged. In the stationkeeping case, the constrained controller achieved a success rate of 98\% over 200 Monte Carlo trials with a spherical primary (vs.\ 4\% for the unconstrained baseline), and 80.5\% for a rotating ellipsoid. In the approach case starting near the range limit, the constrained controller achieved 90.5\% success for the sphere (vs.\ 34\% baseline) and 84\% for the ellipsoid. These outcomes align with the time histories: (i) the angle trajectories for the constrained runs remained above the $30^\circ$ half-angle threshold, (ii) the EKF position errors were predominantly bounded by the $3\sqrt{\lambda_{\max}(P_r)}$ envelopes, and (iii) the Lyapunov function exhibited characteristic peaks coincident with constraint activation, followed by monotone decrease as the trajectory re-entered the interior of the safe set. The controller maintained performance under modeling and operational variations such as rotating ellipsoids inducing periodic geometry variations, high initial state dispersion, and challenging target orbits that are near the observability constraint.}
\\\\
\oldr{It is important to note that this controller formulation is for continuous, low-thrust control. As determined in \cref{results}, the maximum instantaneous acceleration across both scenarios was $6.6203\times10^{-6}\,\textrm{m/s}^{2}$, which aligns with low-thrust capabilities \cite{jia-richards2023method}. \tr{The observability constraint also aids the controller in maintaining a thrust profile below the spacecraft's capabilities, while the unconstrained case leads to frequent thrust maximizations. Furthermore, the observability-constrained controller maintains consistently low $\Delta v$ levels within spacecraft capabilities.}}
\\\\
While effective under certain conditions, the implementation shows a divergence in scenarios with extreme noise, highly nonlinear dynamics, or erroneous initial states. This highlights either a possible need for further tuning and refinement of the EKF or a limitation due to the controller not accounting for navigation errors. However, it is effective within realistic bounds and expectations of these parameters that would be found in an actual space mission. The system demonstrates effective state estimation for both spherical and ellipsoidal bodies while the spacecraft is in stable orbits.
\\\\
The current approach to trajectory generation through Lyapunov control has shown significant improvements. However, testing it against a wider range of scenarios with varying tuning parameters is vital to determine a more intuitive approach to adapting the controller to any desired mission profile. Formulating an objective function that analytically quantifies observability within the Lyapunov controller rather than using a path constraint could lead to even more robust and stable performance. Another avenue for investigation is the OpNav measurement interval and how it may affect the controller's success rate. This may create a baseline for designing the spacecraft sensor to balance measurement cost and state estimate accuracy. Additionally, the gain must be very precisely tuned, as the asteroid environment requires precise control. It is very sensitive and, therefore, highly susceptible to minor gain or control profile changes. Potential improvements to the algorithm include refining controller gain settings for different mission profiles and exploring additional optical methods to enhance shape and state estimation simultaneously.
\\\\

\section{Conclusion}

This research successfully demonstrates a robust autonomous navigation and control method to maintain OpNav observability for spacecraft approaching or conducting operations near small bodies. A comprehensive simulation environment is developed that considers spacecraft dynamics around an asteroid, synthetic asteroid imaging and processing, horizon-based OpNav, EKF, and observability-constrained Lyapunov control. The novel contribution is a Lyapunov controller with path constraints to improve horizon-based OpNav measurement observability. Its effectiveness has been demonstrated by testing controller stability when combined with EKF-based state estimation within acceptable error bounds. 
\\\\
\tr{We conduct extensive empirical validation demonstrating onboard feasibility. In two representative proximity scenarios, the controller delivered substantial gains in success rate over an unconstrained Lyapunov baseline. It improved from 4\% to 98\% in stationkeeping and 34\% to 90.5\% in approach for a spherical primary while preserving constraint satisfaction throughout successful runs and producing low-thrust, low-$\Delta v$ profiles compatible with electric propulsion. Comparable performance for rotating ellipsoids (80.5\% and 84\% success) supports robustness to target geometry and operational variability. The analysis incorporates Monte Carlo statistics, Lyapunov function evolution, position error bounds, thrust histories and $\Delta v$ distribution histograms, providing a complete picture of stability, constraint adherence, and control effort across diverse mission conditions. The validation with simplified sphere and ellipse small-body models as a benchmark highlights the reliable test of its capabilities in this application. The system’s robustness in varying scenarios highlights its potential for broader applications in autonomous small body exploration. }

\newpage
\begin{appendices}
\section{\tr{Synthetic Image Generation Pipeline}}\label{app:sigp}
%\comment{This section is shifted to the Appendix, since it is not the novel focus, but useful for simulation repeatability.} 
The image generation is done using built-in \oldr{MATLAB} graphics tools. The procedure involved illuminating a white sphere \tr{or} ellipsoid against a black background using an infinitely far away light source that is aligned to originate from the negative $x$-axis in the Hill frame to represent a parallel ray light source like the Sun. The camera is then placed based on an input distance, azimuth, and elevation calculated from the current spacecraft position and an image is captured. Gaussian white noise is added to the image to add realistic complexity for the filter to process.  Edge detection is executed by identifying high gradient magnitudes between pixels on the image, which is commonly where the lit limb edge meets blank space. The generated images were compared with those produced by existing higher fidelity image generators in identical conditions and found to be sufficiently accurate when comparing the resulting OpNav measurement. \\

For the simulations in this paper, we generate the spheres and ellipsoids using the mean radius of any sample \tr{small body. }Below is a stepwise summary of the process utilized for the synthetic image generation:

\begin{enumerate}
    \item Generate a sphere or ellipsoid scaled according to the camera settings and distance or an experimentally determined scale factor.
    \item Rotate the object using the \tr{small body's axis, rotation period and simulation time elapsed}.  This information is used to calculate and apply the angle by which the object had rotated at each time step during simulation with the equation $\theta_\textrm{sb}=\frac{t}{t_\textrm{rp}}360^{\circ}$ where $\theta_\textrm{sb}$ is the angle by which the \tr{small body} has rotated, $t$ is the time elapsed and $t_\textrm{rp}$ is the \tr{small body}'s synodic rotation period.
    \item Create a figure with invisible, equal, tight axes and a black background. Plot the generated object as a surface with LineStyle set to none and a white colormap.
    \item Set the view angle, camera position, and roll according to the current spacecraft attitude.
    \item Delete existing lights and replace them with a light set at an infinite distance to simulate the Sun. Use Gouraud lighting and dull material for realistic surface and lighting conditions, with no ambient or specular reflection, a diffused reflection coefficient of 1, and a shininess coefficient of 10. These parameters were chosen as they produced the most visually accurate images compared to real examples.
    \item Set the figure size and axis limits according to the image size, apply the camera's FOV setting to the axes' camera view angle, and set aspect ratios to be equal\oldr{.}
    \item Store the image as a frame, then convert it to an image and resize it to the calculated image size in pixels. Convert the image to grayscale and output it for further processing.
    \item \oldr{Add white gaussian noise if desired. For this study, $\sigma=0.01$ of normalized noise was applied to the images.} 
\end{enumerate}

\begin{figure}[htbp!]
    \centering
    \includegraphics[width=1\linewidth,trim={0 0 9cm 0},clip]{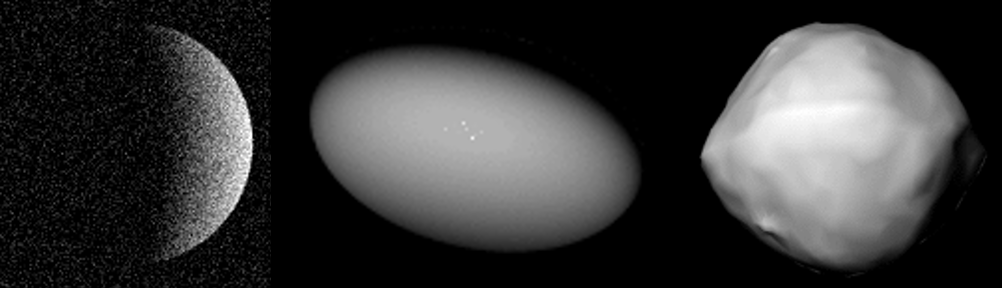}
    \caption{Example synthetic images of a half-illuminated sphere with noise (left), \tr{and} ellipsoid \tr{(right)}}
    \label{fig:illum}
\end{figure}

\begin{figure}[htbp!]
    \centering
    \includegraphics[width=1\linewidth]{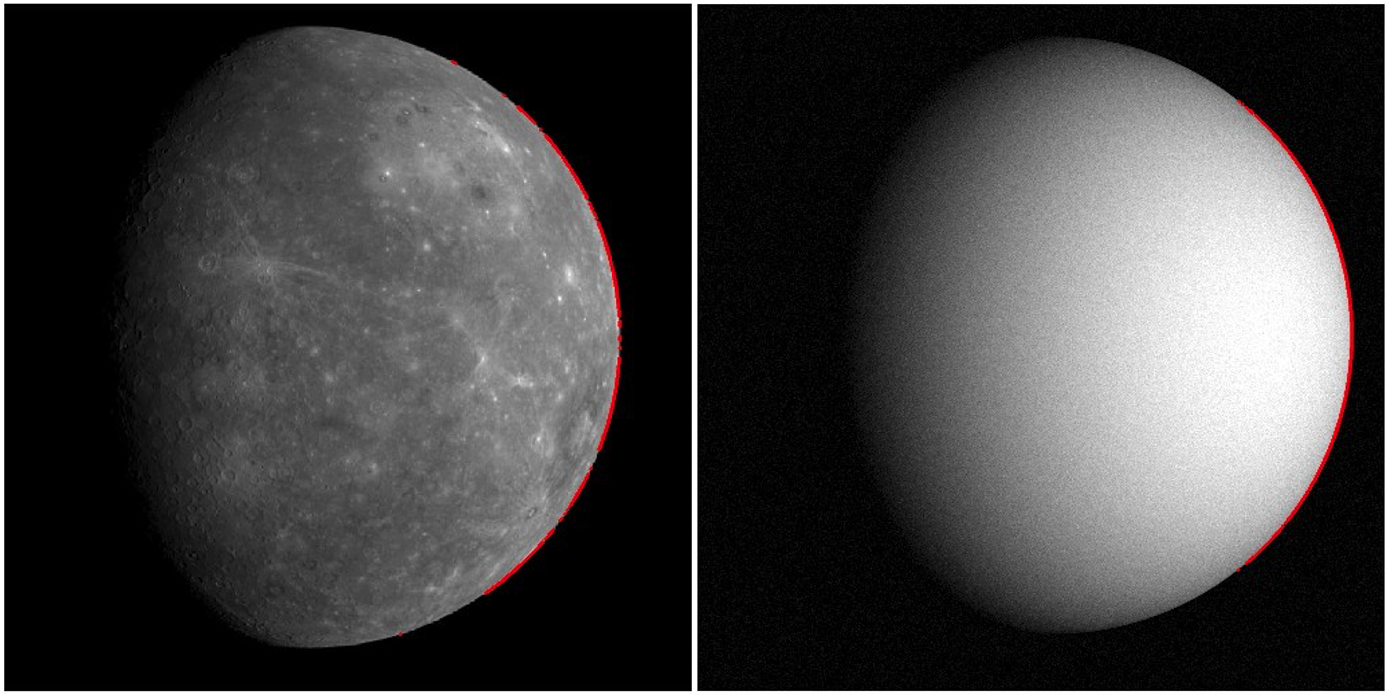}
    \oldr{\caption{Sample image comparison between PIA10172 from MESSENGER \cite{mess} of Mercury and a synthetic generated image at a distance of $\approx27000\;\mathrm{km}$, FOV of $10.5^\circ$ and resolution of $1024\times1024$, with detected horizon points overlaid.}\label{fig:whzcmp}}
\end{figure}

The synthetic image generation process is also expressed in \cref{alg:asteroid_image_generation}:

\begin{algorithm}[htbp!]
\caption{Image Generation Procedure}
\label{alg:asteroid_image_generation}
\begin{algorithmic}[1]
    \Procedure{$img$=ImageGenerator}{$x\_y\_z, up, dist, FOV, imgSize, rad, ellip\_shape, R\_ellip$}
        \Statex \textbf{Input:} 
        \Statex \hspace{\algorithmicindent}$x\_y\_z$: Camera direction vector
        \Statex \hspace{\algorithmicindent}$up$: Camera "up" pointing direction vector
        \Statex \hspace{\algorithmicindent}$dist$: Distance from camera to object center
        \Statex \hspace{\algorithmicindent}$FOV$: Camera field of view
        \Statex \hspace{\algorithmicindent}$imageSize$: Desired square image dimension in pixels
        \Statex \hspace{\algorithmicindent}$rad$: Characteristic radius of the \tr{small body}
        \Statex \hspace{\algorithmicindent}$ellip\_shape$: Scaling factors $(a,b,c)$ matrix defining ellipsoid shape
        \Statex \hspace{\algorithmicindent}$R\_ellip$: Rotation angle about the chosen rotation axis
        \Statex \hspace{\algorithmicindent}$R\_axis$: Rotation axis vector in body frame

        \Statex \textbf{Output:}
        \Statex \hspace{\algorithmicindent}A grayscale image of the half-illuminated ellipsoid approximating the \tr{small body}.

        \vspace{0.5em}

        \State Generate object (sphere, ellipsoid or mesh) $[X,Y,Z]$ coordinates using center=$(0,0,0)$ and shape=$ellip\_shape$
        \State Rotate Object by axis=$R\_axis$ and angle=$R\_ellip$
        \State Create a figure with a black background and invisible, equal, and tight axes
        \State Plot object with invisible edges and white color

        \State Set viewDirection=$x\_y\_z$
        \State Set cameraPositionNormalize=$dist \cdot x\_y\_z$
        \State Set cameraUpDirection=$up$

        \State Set light at infinite distance with direction=$[-1 0 0]$ and Gouraud type
        \State Set dull material with 0 ambient and specular, full diffuse reflectivity and shininess coefficient of 10

        \State Set figure size = $imgSize$ and axis limits = [$-imgSize/2$,$imgSize/2$]
        \State Apply camera field of view = $FOV$

        \State Convert image to grayscale
        \State Apply noise to the image
        \State Return the image
    \EndProcedure
\end{algorithmic}
\end{algorithm}
\newpage
A few examples of generated images are shown in \cref{fig:illum}
\\
A gradient-based approach detects the horizon points from the generated synthetic images. First, calculate the gradient of the image matrix and find the gradient magnitude at each pixel. Next, find the maximum gradient magnitude across the entire image. Then, multiply this value by 0.7 to define a minimum gradient threshold and find the pixel coordinates of all points with a gradient magnitude above it. These are the desired horizon points. The value of 0.7 was chosen for the best performance after some experimental testing. To avoid detecting random noisy pixels, the absolute minimum gradient threshold was set to 0.1, which was determined experimentally by finding an appropriate threshold above the maximum gradient in dark-side lighting conditions but below the maximum gradient of regular lighting conditions. The edge detection process is expressed in \cref{alg:edge_detect}: 
\begin{algorithm}[htbp!]
\caption{Gradient Edge Detection Procedure}
\label{alg:edge_detect}
\begin{algorithmic}[2]
    \Procedure{$points$=GradientEdgeDetect}{$img, threshold$}
        \Statex \textbf{Input:} 
        \Statex \hspace{\algorithmicindent}$img$: A grayscale image of the half-illuminated ellipsoid approximating the \tr{small body}.
        \Statex \hspace{\algorithmicindent}$threshold$: The gradient threshold used to determine edge points.

        \Statex \textbf{Output:}
        \Statex \hspace{\algorithmicindent}$points$: A $3\times n$ matrix containing the coordinates of the lit edge pixels detected with each pixel stored in the form $[x,y,1]^\top$

        \vspace{0.5em}

        \State Convert $img$ to numerical double format ($img\_dbl$) for gradient calculation
        \State Calculate the 2D $gradient$ of $img\_dbl$
        \State Compute the $gradient\_magnitude$ from the $gradient$
        \State Calculate $minimum\_gradient=threshold\times MAX(gradient\_magnitude)$
        \State Add all points with $gradient\_magnitude>minimum\_gradient$ to $points$
        \State Return $points$
    \EndProcedure
\end{algorithmic}
\end{algorithm}
\oldr{To validate the image generation and edge detection methods, sample images of Mercury from the MESSENGER mission and corresponding distance and camera settings were used to compare results, with the OpNav measurement difference when applied on either image remaining within $1\%$ of the absolute distance. \cref{fig:whzcmp} shows the similarity between the aforementioned synthetic methods and a real image.}
\subsection{\tr{Edge Detection Threshold Analysis}}
\begin{figure}[htbp!]
    \centering
    \includegraphics[width=0.5\linewidth]{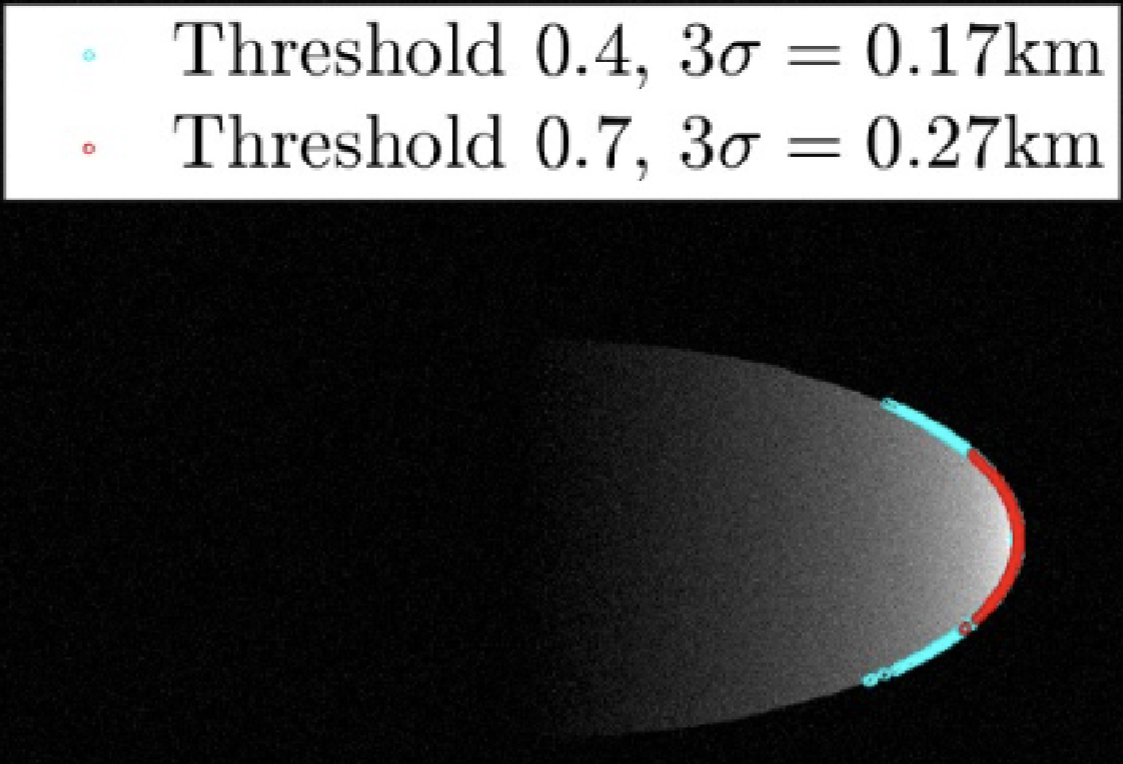}
    \caption{Example of detected limbs for edge detection thresholds of 0.4 (longer cyan arc) and 0.7 (shorter red arc) with 3$\sigma$ values of the error in position measurement norm}
    \label{fig:thres}
\end{figure}

This implementation uses a pixel gradient method to detect the \tr{small body}'s lit limb edge. The maximum gradient on the image is found. All pixels with a gradient within a certain threshold of the maximum gradient are stored as the detected edge points for the OpNav measurement. The selection of this gradient, a value between 0 and 1, is critical to the measurement performance. Thus, we analyzed the performance for varying thresholds. The difference between the limb detected by a high threshold and the ideal threshold is shown in \cref{fig:thres}. Lower thresholds were found to lead to a higher lit limb angle and number of detected points, coupled with a lower analytical covariance. Using a lower threhshold may also increase the risk of misdetecting background noise or excess points (such as those along the terminator line) leading to inconsistencies between the analytical and measurement covariance. The opposite effect holds for higher thresholds. Its error did not exceed the measurement covariance as lower thresholds may have due to detecting an excess number of points or including background noise. The ideal balance was found to be a value of 0.4 as it had consistently improved performance from higher thresholds, particularly for ellipsoid cases. At this threshold, sufficient points are included to characterize the \tr{small body} silhouette, while avoiding inconsistencies in analytical and measured covariances or detecting erroneous points.
\newpage
\section{\tr{OpNav Validation with Elliptic Bodies}}
\tr{In this section, we validate our synthetic image generation, edge detection and OpNav implementation for consistency across spherical and elliptical bodies to support correct implementation and provide repeatability reference.}
%\comment{These two subsections are shifted to appendix since they are not relevant to observability constraints, but useful for repeatability. Can be removed if needed.}
\subsection{\tr{Elliptic Shape Performance}}
\begin{figure}[htbp!]
    \centering
    \includegraphics[width=0.7\linewidth]{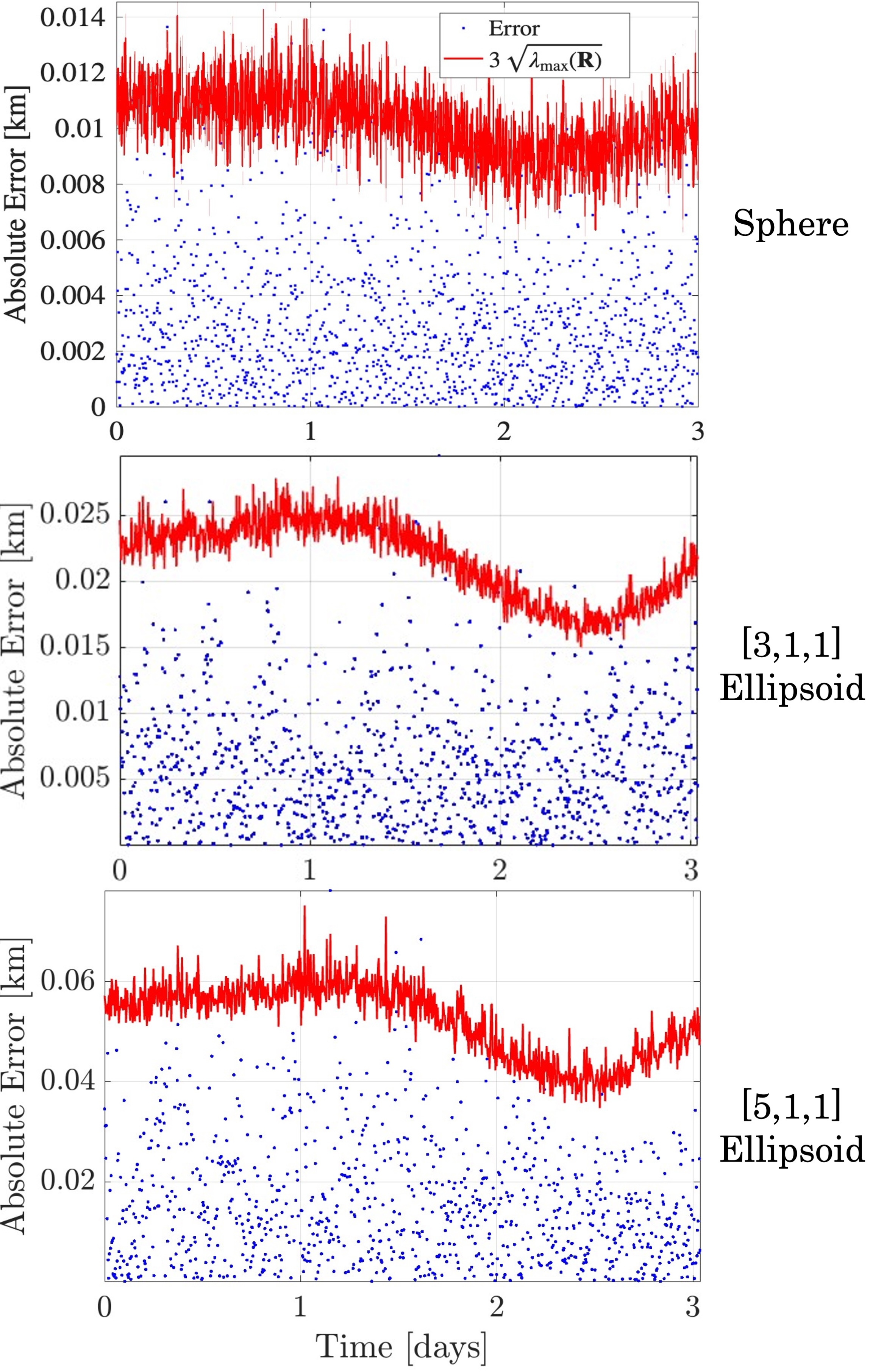}
    \caption{OpNav Measurement Error Plots for Sphere (top), [3,1,1] (center) and [5,1,1] (bottom) Ellipsoid with Error Magnitude (blue points) and Analytical Covariance (red line) over time}
    \label{fig:ellip}
\end{figure}

We now test the OpNav performance against increasingly elliptical bodies. The shape ratios are represented using the notation [a,b,c] where a, b and c represent the principal axis dimensions of the triaxial ellipsoid body and the principal axis frame is defined such that $a\geq b\geq c$ according to convention \cite{OpNav}. To isolate the effect of body ellipticity, rotation is ignored for this test. As can be seen in \cref{fig:ellip}, the OpNav error and covariance increase with highly elliptic bodies. The error \oldr{remains within $\approx3\%$ of the absolute distance and} is correctly bounded by the covariance for all cases \tr{including} an ellipsoid with a shape ratio of [5,1,1], supporting its usability for state estimation with ellipsoid bodies. For this test case, we simulate \tr{the spacecraft camera} moving along one revolution of an FTO \tr{around the small body} with a radius of $2.0429$km \tr{and orbit period of 3 days}. 

\subsection{\tr{Rotating Ellipsoid Performance}}
\begin{figure}[htbp!]
    \centering
    \includegraphics[width=0.75\linewidth]{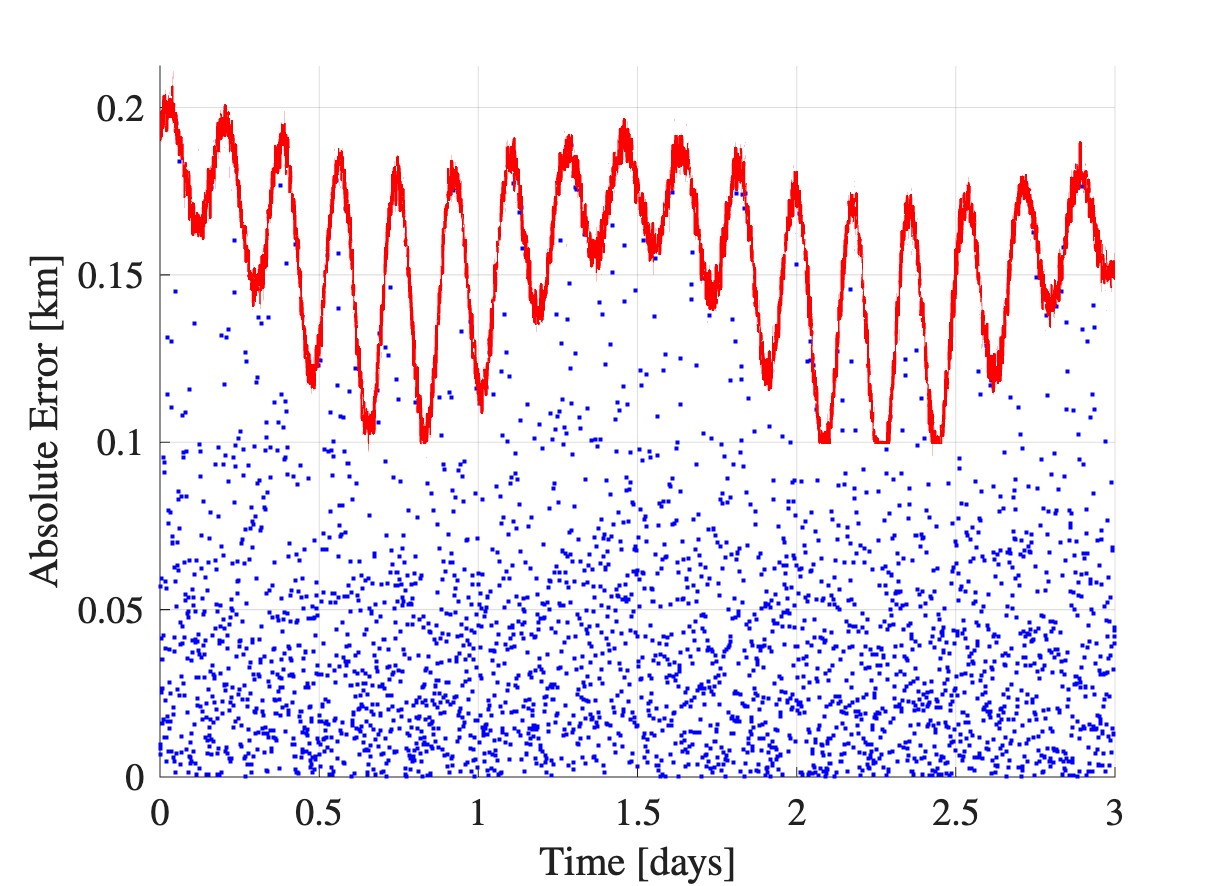}
    \caption{OpNav measurement error (blue points) and analytical covariance (red line) over time for a rotating [2.5,1,1] ellipsoid}
    \label{fig:limb}
\end{figure}

The elliptic shape scenario is repeated with rotation to determine its effect on OpNav performance. A [2.5,1,1] ellipsoid with the rotation period and axis of Bennu are used as described in \cref{app:sigp}. \tr{The axis is at approximately 180$^{\circ}$ \cite{lauretta_unexpected_2019} from the z-axis in the Hill frame with a synodic rotation period of 4.296057 hours \cite{lauretta_osiris-rex_2015}.} For this test case, we simulate along one revolution of an FTO with a radius of $2.0429$km. The result \oldr{in \cref{fig:limb}} suggests that rotation significantly \oldr{varies} the accuracy of OpNav due to various alignments leading to lit limb geometries that are better or worse for accuracy. \oldr{High errors occur more frequently at higher ellipse ratios not due to limitations of the CRA, but rather due to camera limitations, as the limb reaches the limit of what can be accurately characterized with the chosen camera settings, particularly the FOV and resolution, causing relatively poorer curvature resolution or the limb exceeding the image frame. The error and covariance vary periodically with the rotation which corresponds with the higher frequency behaviour. This occurs as the limb oscillates between eccentric and circular when viewing at a perpendicular angle. The lower frequency behaviour is an effect of viewing the orbiting body from different elevations relative to the rotation axis. When straight-on to the pole, the limb curvature and error remain consistent, which corresponds to the dips in error while viewing at a perpendicular angle results in the general peaks and greatest oscillation.} The measurement error was found to be correctly bounded in \oldr{either rotating or non-rotating case but with increased magnitude for the worst error, reaching $\approx10\%$ or higher levels than} the non-rotating [5,1,1] ellipsoid test case\oldr{. However, since the error is still correctly bounded by the covariance, the measurement can be used in rotating cases and accounted for through filtering methods by using the covariance knowledge.}

\section{Path Constraint Partial Derivatives}\label{app:pcpd}
This section extends the derivation in \cref{pcf} by computing the partial derivatives of the path constraints. The path constraint partial derivatives in terms of Milankovitch orbital elements are as follows:
\begin{align}
\frac{dg_1(^H\bm{x}_\textrm{slow})}{d^H\bm{x}_\textrm{slow}}&=\begin{bmatrix}
-\frac{(2 h_1 + h_2^2 + h_3^2)}{\mu} \frac{1}{1-e} \\
-\frac{(2 h_2 + h_1^2 + h_3^2)}{\mu} \frac{1}{1-e} \\
-\frac{(2 h_3 + h_2^2 + h_1^2)}{\mu} \frac{1}{1-e} \\
-\frac{e_1 h^2}{\mu e (e-1)^2} \\
-\frac{e_2 h^2}{\mu e (e-1)^2} \\
-\frac{e_3 h^2}{\mu e (e-1)^2}
\end{bmatrix}\\
\frac{dg_2(^H\bm{x}_\textrm{slow})}{d^H\bm{x}_\textrm{slow}}&=\begin{bmatrix}
-\frac{(2 h_1 + h_2^2 + h_3^2)}{\mu} \frac{1}{1-e} \\
-\frac{(2 h_2 + h_1^2 + h_3^2)}{\mu} \frac{1}{1-e} \\
-\frac{(2 h_3 + h_2^2 + h_1^2)}{\mu} \frac{1}{1-e} \\
-\frac{e_1 h^2}{\mu e (e-1)^2} \\
-\frac{e_2 h^2}{\mu e (e-1)^2} \\
-\frac{e_3 h^2}{\mu e (e-1)^2}
\end{bmatrix}\\
\frac{dg_3(^H\bm{x}_\textrm{slow})}{d^H\bm{x}_\textrm{slow}}&=[\frac{h_2^2 + h_3^2}{h^{1.5}},-\frac{h_1 h_2}{h^{1.5}},-\frac{h_1 h_3}{h^{1.5}},0_{3\times1}]
\end{align}
$h_2$ and $h_3$ are the components of angular momentum in the y-axis and z-axis direction respectively, in the Hill frame.
\section{Controller Gain Tuning Guidelines}\label{tuning}
Since the gains determined for the controller used in this research are for specific test cases with Bennu\tr{'s properties}, it is imperative to develop guidelines to derive it for different scenarios to make it useful for diverse missions. After investigating the relations between the gains and the controller performance, the following observations and guidelines are made:
\begin{enumerate}
    \item The base gain magnitude used is $10^{-3}$ where the gains in the angular momentum vector and eccentricity vector are tweaked to a different ratio across varying axes depending on in which direction the control is intended to be imparted. This value seems to offer a critical balance and match with a few orders of magnitude below the angular momentum and eccentricity values used in this research, allowing precise changes. The eccentricity gains never exceed the angular momentum, as this is found to create unstable results.
    \item The gain matrix used in the first scenario is used for stationkeeping. Here, the angular momentum in the x direction in the Hill frame has a higher gain, and the eccentricity in the x and y directions in the Hill frame has lower gains. The intuition is that this allows the controller to readily tweak the spacecraft's position and counter the effect of SRP. This distinction in gain may not be required for a different application where the dynamics are more balanced.
    \item The gain used in the second scenario is intended to focus on circularization. Here, the eccentricity in the y and z direction in the Hill frame is set to a lower gain, which allows the controller to rapidly adjust the eccentricity in the x direction in the Hill frame to circularize the orbit while maintaining the spacecraft in the desired orbital plane. Depending on the nature of the mission and target orbit, different axes can have modified gains to execute transfers within a desired plane.
\end{enumerate}
This Lyapunov controller formulation may be adapted for different use cases using these observations. Results demonstrating the effectiveness of these gains can be seen in \cref{results}.
\end{appendices}
\section*{Declarations}
\subsection*{Conflict of Interest}
On behalf of all authors, the corresponding author states that there is no conflict of interest.
\subsection*{Funding}
On behalf of all authors, the corresponding author states that there was no source of funding for this work.
\subsection*{Data Availability}
Data sets generated during this study during simulations and that support its findings are available from the corresponding author on reasonable request. 
\subsection*{Author Contributions}
Both authors jointly contributed to the conceptualization and development of the ideas, theoretical framework and derivations. Aditya Arjun Anibha led the implementation, conducted the simulations, and drafted the manuscript. Dr. Kenshiro Oguri supervised the analysis, provided critical feedback, and contributed to the revision and refinement of the manuscript.
\bibliographystyle{sn-mathphys-num}
\bibliography{bib}
\end{document}